%% file: colm2024_conference.tex
\documentclass[dvipsnames]{article} %

\PassOptionsToPackage{hidelinks}{hyperref}
\PassOptionsToPackage{dvipsnames}{xcolor}
\usepackage{colm2024_conference}
\makeatletter
\def\addcontentsline#1#2#3{%
  \addtocontents{#1}{%
    \protect\contentsline{#2}{#3}{\thepage}{\@currentHref}%
  }%
}
\makeatother
\input{config.tex}

\title{RxBrain: Embodied Cognition Foundation Model with \\ Joint Language-Visual Reasoning and Imagination}

\author{
\bf Tencent Robotics X Team $\times$ Futian Laboratory $\times$ Tencent Hy Team
}

\begin{document}

\maketitle

\pagestyle{rxheader}
\thispagestyle{rxheader}

\begin{abstract}

Embodied cognition requires agents to connect high-level task reasoning with the physical states to be achieved. We introduce \textbf{Hy-Embodied-RxBrain}, an embodied cognition foundation model with joint language-visual reasoning and imagination. Unlike vision-language models that emphasize scene understanding and textual decision making, or generative world models that mainly predict future visual states, RxBrain represents embodied plans in a single planning sequence where language and visual imagination play complementary roles. Language provides the abstract structure of a plan, including task decomposition, planning primitives, constraints, temporal order, and decision logic, while visual imagination grounds this structure through world state prediction and joint subgoal planning, associating each planning step with intermediate and final physical states. RxBrain adopts a unified multimodal Mixture-of-Transformers architecture that supports language, image, and video understanding and generation within one model. To train this capability, we build an automatic pipeline that converts embodied videos into joint text-visual planning supervision by decomposing videos into planning steps and aligning them with visual state transitions. We further introduce RxBrain-Bench to evaluate whether models can represent embodied plans through joint textual and visual components rather than separate understanding or generation. Experiments show that RxBrain maintains embodied understanding and generation abilities, and produces plans with coupled textual reasoning, world state prediction, and joint subgoal planning. We also extend RxBrain to continuous robot action generation, where it shows promising real-robot performance without large-scale action-data pretraining. These results provide an initial step toward foundation models for embodied cognition.

\end{abstract}

\begin{center}
  \vspace{1em}
  \includegraphics[width=0.98\linewidth]{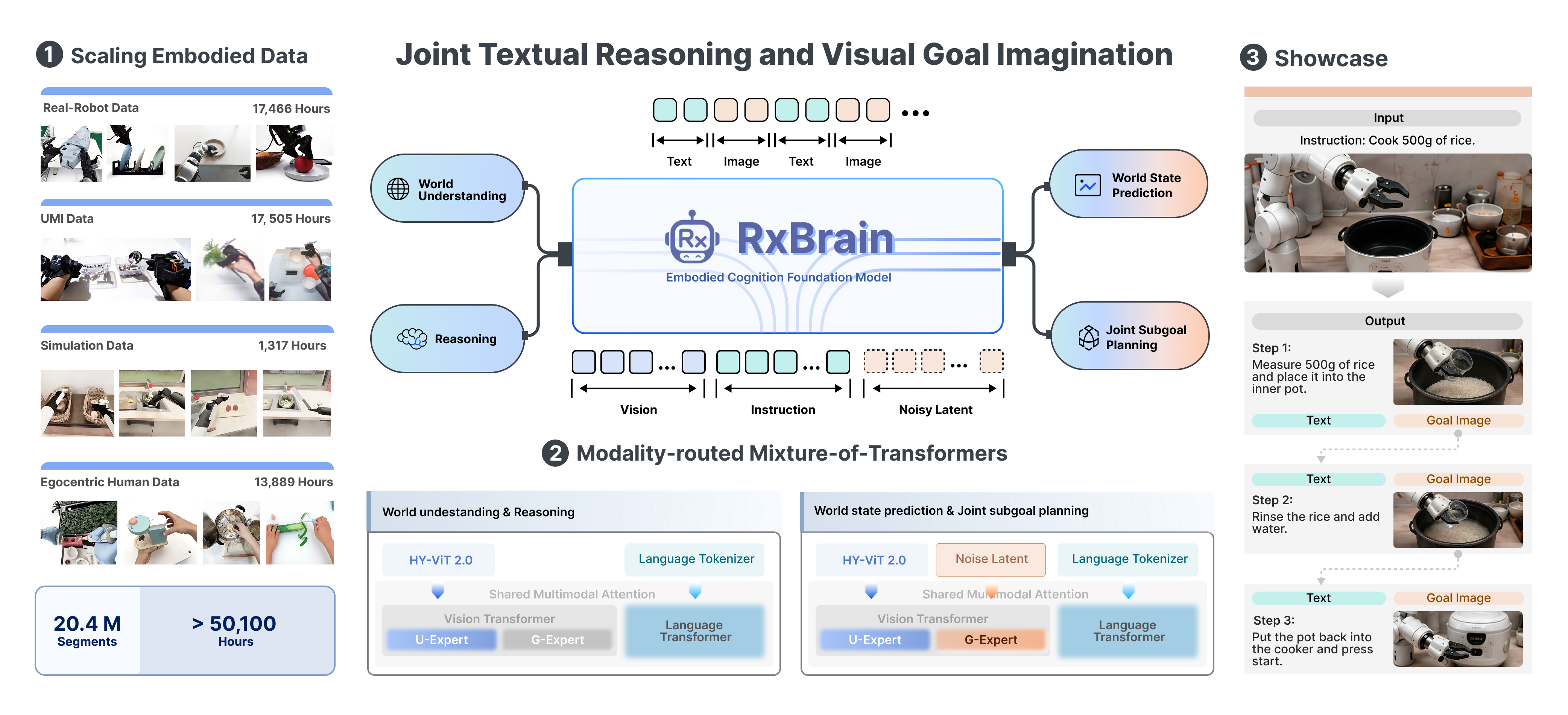}
\end{center}

\vfill

\newpage

\tableofcontents
\clearpage

\input{content/intro}

\input{content/model_arch}

\input{content/data}

\input{content/training_fix}
\input{content/self_bench}

\input{content/evaluation}
\input{content/action}

\input{content/related}

\input{content/conclusion}

\clearpage
\appendix
\input{content/appendix.tex}

\bibliography{biblio}
\bibliographystyle{colm2024_conference}

\end{document}

%% file: config.tex
\usepackage{booktabs}
\usepackage{graphicx}
\usepackage{enumitem}
\usepackage{wrapfig}
\usepackage{algorithm}
\usepackage{algpseudocode}
\usepackage{array}
\usepackage{booktabs}
\usepackage[table]{xcolor}
\usepackage{xspace}
\usepackage{acronym}
\usepackage{float}
\usepackage{microtype}
\usepackage{amsmath}
\usepackage{amssymb}
\usepackage{cleveref}
\usepackage{colortbl}
\usepackage[utf8]{inputenc}
\definecolor{lightgray}{rgb}{0.9,0.9,0.9}
\usepackage{caption}
\usepackage{subcaption}
\usepackage{multicol}
\usepackage{xcolor}

\usepackage{setspace}
\usepackage{url}
\usepackage{multirow}
\usepackage{colortbl}
\usepackage{tabularx}
\usepackage{blindtext}
\usepackage{pgfplots}
\pgfplotsset{compat=1.18}
\usepackage{tikz}
\usetikzlibrary{er,positioning,bayesnet}
\usepackage{makecell}
\usepackage{siunitx}
\usepackage{nicefrac}
\usepackage{tocloft}

\usepackage{listings}
\usepackage[raster,skins]{tcolorbox}
\usepackage{xltabular}
\usepackage{adjustbox}
\usepackage{xurl}
\usepackage{rotating}
\usepackage{placeins}
\usepackage[normalem]{ulem}
\usepackage{pdflscape}

\useunder{\uline}{\ul}{}

\definecolor{todopink}{RGB}{220,20,120}
\definecolor{first}{rgb}{0.75, 0.92, 0.75}      %
\definecolor{second}{rgb}{0.82, 0.93, 0.60}      %
\definecolor{third}{rgb}{1.00, 0.97, 0.65}      %
\definecolor{annobg}{rgb}{0.90,0.95,1.00}  %

\crefname{algorithm}{Alg.}{Algs.}
\Crefname{algocf}{Algorithm}{Algorithms}
\crefname{section}{Sec.}{Secs.}
\Crefname{section}{Section}{Sections}
\crefname{table}{Tab.}{Tabs.}
\Crefname{table}{Table}{Tables}
\crefname{figure}{Fig.}{Figs.}
\Crefname{figure}{Figure}{Figures}
\crefname{equation}{Eq.}{Eqs.}
\Crefname{equation}{Equation}{Equations}
\crefname{appendix}{Appx.}{Appxs.}
\Crefname{appendix}{Appendix}{Appendices}

\makeatletter
\DeclareRobustCommand\onedot{\futurelet\@let@token\@onedot}
\def\@onedot{\ifx\@let@token.\else.\null\fi\xspace}

\makeatother

\acrodef{ai}[AI]{Artificial Intelligence}
\acrodef{llm}[LLM]{Large Language Model}
\acrodef{vlm}[VLM]{Visual Language Model}
\acrodef{mllm}[MLLM]{Multimodal Large Language Model}

\input{math_commands.tex}

\newcommand{\model}{RxBrain}
\newlength{\ilcw}
\newcolumntype{K}{>{\raggedright\arraybackslash}p{\ilcw}}
\newcommand{\ilcell}[2]{\includegraphics[width=\linewidth]{#1}\par\vspace{1.5pt}{\scriptsize #2\par}}
\newcommand{\mfcell}[1]{\includegraphics[width=\linewidth]{#1}}

\newcommand*\justify{%
  \fontdimen2\font=0.4em%
  \fontdimen3\font=0.2em%
  \fontdimen4\font=0.1em%
  \fontdimen7\font=0.1em%
  \hyphenchar\font=`\-%
}

\renewcommand{\texttt}[1]{%
  \begingroup
  \ttfamily
  \begingroup\lccode`~=`/\lowercase{\endgroup\def~}{/\discretionary{}{}{}}%
  \begingroup\lccode`~=`[\lowercase{\endgroup\def~}{[\discretionary{}{}{}}%
  \begingroup\lccode`~=`.\lowercase{\endgroup\def~}{.\discretionary{}{}{}}%
  \catcode`/=\active\catcode`[=\active\catcode`.=\active
  \justify\scantokens{#1\noexpand}%
  \endgroup
}

\usepackage{fancyhdr}

\fancypagestyle{rxheader}{
  \fancyhf{}
  \fancyhead[L]{\includegraphics[height=11pt]{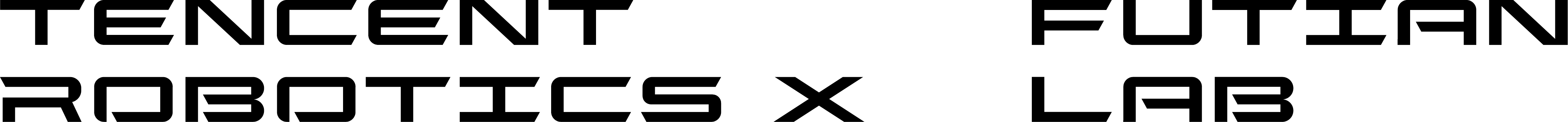}}
  \fancyhead[R]{\includegraphics[height=13pt]{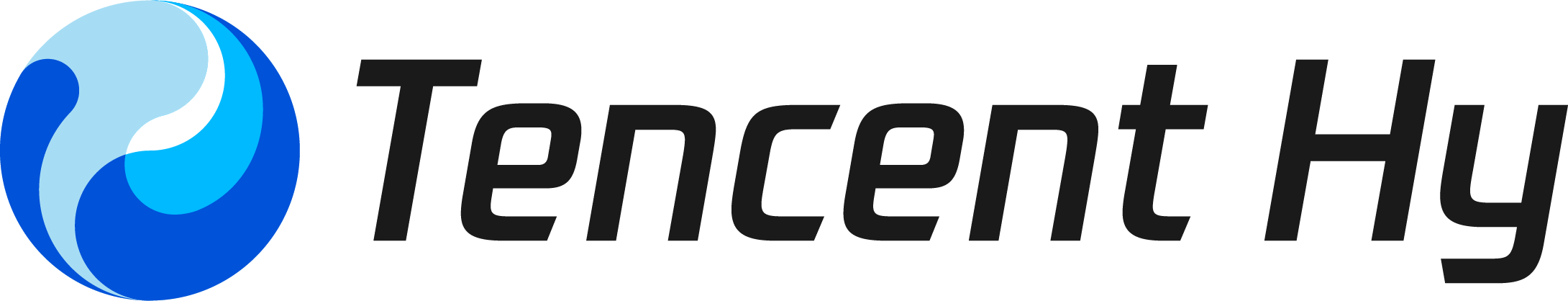}}
  \fancyfoot[C]\thepage
  \renewcommand{\headrulewidth}{0.4pt}
  \renewcommand{\headrule}{%
    {\color[HTML]{4D4D4D}\hrule width\headwidth height\headrulewidth}%
    \vskip-\headrulewidth
  }
}

\fancypagestyle{plain}{
  \fancyhf{}
  \fancyhead[L]{\includegraphics[height=11pt]{./logo/rx-ft-logo.pdf}}
  \fancyhead[R]{\includegraphics[height=13pt]{./logo/hy-logo.png}}
  \fancyfoot[C]\thepage
  \renewcommand{\headrulewidth}{0.4pt}
  \renewcommand{\headrule}{%
    {\color[HTML]{4D4D4D}\hrule width\headwidth height\headrulewidth}%
    \vskip-\headrulewidth
  }
}

%% file: math_commands.tex
\usepackage{amsmath,amsfonts}
\providecommand{\bm}[1]{\boldsymbol{#1}}

\newcommand{\newterm}[1]{{\bf #1}}

\def\Figref#1{Figure~\ref{#1}}

\def\Secref#1{Section~\ref{#1}}

\def\eqref#1{equation~\ref{#1}}
\def\Eqref#1{Equation~\ref{#1}}

\def\Algref#1{Algorithm~\ref{#1}}

\def\1{\bm{1}}

\DeclareMathAlphabet{\mathsfit}{\encodingdefault}{\sfdefault}{m}{sl}
\SetMathAlphabet{\mathsfit}{bold}{\encodingdefault}{\sfdefault}{bx}{n}

\def\gG{{\mathcal{G}}}

\def\gK{{\mathcal{K}}}

%% file: content/intro.tex
\section{Introduction}
\label{sec:intro}

Embodied cognition requires an agent to reason not only about what to do, but also about what the world should look like after its actions. Textual reasoning provides a natural interface for abstract task understanding, causal inference, constraint satisfaction, and long-horizon decomposition, enabling an embodied agent to formulate plans beyond immediate perception. Meanwhile, visual goal imagination offers a complementary form of reasoning by representing desired physical states, spatial configurations, and object interactions in the modality where embodied actions ultimately take effect. However, in complex tasks such as planning, neither textual reasoning nor visual imagination alone can fully specify an executable intention: language may describe high-level procedures while omitting critical spatial or state details, whereas visual goals may depict desired outcomes without exposing the underlying causal structure or intermediate decisions required to achieve them. This motivates a joint formulation of embodied cognition, where textual reasoning and visual goal imagination are integrated to represent, reason about, and plan toward goals in the physical world.

Recent advances in embodied AI have largely developed along two complementary directions. Vision-language models (VLMs)~\citep{ji2025robobrain,robobrain2025v2,dang2025rynnec,dang2026rynnbrain,lee2025molmoact,fang2026molmoact2} provide a general interface for connecting perception, language, and action: they can interpret visual observations, understand instructions, ground objects and spatial relations, decompose tasks, and support high-level decision-making through textual reasoning. However, their cognitive process is typically expressed in language, leaving the desired physical outcome only implicitly specified and often lacking an explicit imagination of the visual goal state. In parallel, world models and generative models~\citep{guo2025ctrl,team2026lingbotworld,lingbot-va2026,ye2026dreamzero} focus on predicting future observations, simulating state transitions, or synthesizing goal-conditioned visual outcomes, making them particularly suitable for visual planning. Yet these models are usually less capable of explicit causal reasoning, abstract task understanding, and constraint-aware decision-making. As a result, existing approaches often separate the two sides of embodied cognition: VLMs reason over what should be done, while world models imagine what might happen, but neither fully integrates textual reasoning with visual goal imagination.

More recently, embodied foundation models have begun to move beyond the separation between VLMs and world models. A representative example is Cosmos 3~\citep{agarwal2026cosmos}, which connects understanding, generation, simulation, and action within an omnimodal framework for physical AI. Such models mark an important step toward unified embodied intelligence. However, despite integrating reasoning and generation within a unified framework, they still tend to treat them as separate capabilities rather than organizing them into a coupled cognitive process. This suggests that joint embodied reasoning requires more than architectural unification: the model must be endowed with the ability to coordinate textual reasoning and visual imagination toward the same planning objective. $\pi_{0.7}$~\citep{intelligence2026pi07steerablegeneralistrobotic} explores combining a vision-language model with an image generation model to generate joint language-visual subgoals, which are used as conditioning signals for VLA policies to guide action generation. In the broader multimodal domain, models such as BAGEL~\citep{bagel} further show that text and images can be generated in a mixed or interleaved form, where language can describe, refine, or guide visual generation. This form of jointness is often centered on representing the same semantic content across modalities. Embodied planning requires a more complementary relation between language and visual, where language specifies actions, constraints, and decisions, while visual imagination specifies the goal and world states those actions should realize.

\begin{figure}[h]
  \centering
  \includegraphics[width=1.0\linewidth]{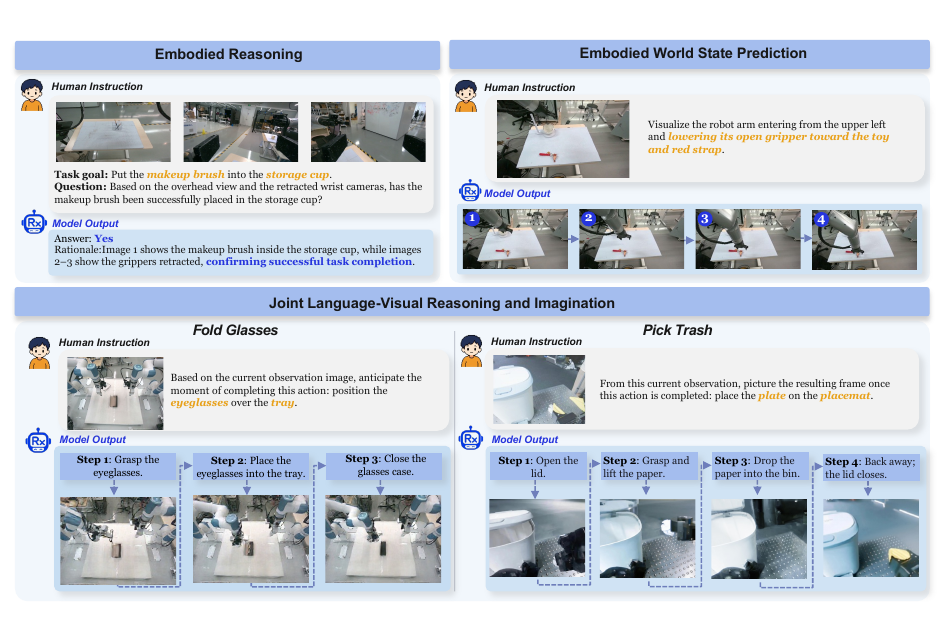}
  \caption{\textbf{Capability overview of \model.} \model~supports multiple embodied tasks within a unified model, including embodied visual question answering, multi-frame visual generation, and joint textual reasoning with visual goal imagination. Given human instructions and visual observations, the model can reason about the current scene, imagine future or goal states, and generate joint language and visual planning step by step.}
  \label{fig:intro}
\end{figure}

To address this challenge, we introduce \model, an embodied cognition foundation model with joint language-visual reasoning and imagination for embodied planning. Rather than treating language reasoning and visual generation as separate capabilities, \model~organizes them within the same planning sequence: language expresses task decomposition, planning primitives, constraints, temporal order, and decision logic, while visual imagination grounds the plan through world state prediction and joint subgoal planning, depicting the intermediate and final physical states that each planning stage is intended to realize, as shown in \Figref{fig:intro}. This allows \model~to represent an embodied plan through complementary textual and visual components, specifying both how a task should progress and what world states should be achieved along the way. Built upon HY-Embodied-0.5\citep{team2026hy}, \model~adopts a unified multimodal architecture based on Mixture-of-Transformers~\citep{liang2024mixture}, supporting understanding and generation across language, images, and videos within a single model. Its modality-specific text and vision transformers enable language understanding/generation and visual understanding/generation to be learned in a shared framework. In the vision transformer, visual understanding tokens and generation tokens are processed by separate FFNs while sharing QKV projections, allowing perception and imagination to interact within the same visual tower. On the generation path, \model~removes the need for additional VAE-encoded visual input tokens while keeping visual generation in the VAE latent space, simplifying the multimodal input pipeline and encouraging visual understanding and generation to share a more aligned semantic representation.

Training such a model requires supervision in which language-visual reasoning and imagination are coupled around the same embodied task. However, data that jointly specifies what actions should be taken and what physical states those actions should produce is scarce and expensive to annotate manually. To address this, we propose an automatic data construction pipeline that converts embodied videos into joint text-visual planning supervision. Given an embodied video, the pipeline decomposes it into action-centric segments, associates each action with its corresponding visual state transition, filters out ambiguous or unsuitable sequences, and organizes the remaining data into multi-granularity samples ranging from step-level subgoals to full-task plans. This supervision teaches \model~to align textual action descriptions with the visual state transitions they produce, enabling each planning step to be represented by both what action should be taken and what physical change it should achieve. We further introduce RxBrain-Bench, a benchmark for embodied cognition foundation models. Unlike existing evaluations that separately measure embodied understanding, or world states prediction, RxBrain-Bench evaluates whether a model can jointly use language and visual imagination to understand tasks, predict goal states, and construct coherent embodied plans across diverse scenarios.

We evaluate \model~on a broad set of embodied and multimodal benchmarks to assess its general image generation embodied understanding and planning capabilities. Beyond these standard evaluations, we use RxBrain-Bench to examine whether a model can represent embodied plans through complementary language and visual components. On RxBrain-Bench, \model~produces planning outputs in which language specifies actions, constraints, and decisions, while images depict the goal and intermediate world states associated with them. We further extend \model~to continuous robot action generation, where it shows encouraging real-robot performance without relying on large-scale action-data pretraining. These results provide an initial step toward embodied cognition models that reason through coupled language and visual planning representations.

%% file: content/model_arch.tex
\section{Model}
\label{sec:arch}

\model\ is a unified embodied foundation model designed to support three core capabilities: embodied understanding and reasoning, world state prediction and joint language-visual embodied subgoal planning. To enable these capabilities, RxBrain is built on a modality-aware Mixture-of-Transformers (MoT) architecture. Tokens from different modalities are routed to specialized Transformer pathways and communicate through cross-modal attention, allowing language, understanding visual tokens and generated visual tokens to be modeled within a unified framework. For visual modeling, visual-understanding tokens and visual-generation tokens share a common vision Transformer, including shared attention projection layers, while modality-aware expert routing provides specialized computation for perception and generation. In addition, RxBrain adopts a hybrid attention design: causal attention is used for autoregressive text generation, whereas frame-wise bidirectional attention is applied to visual tokens for both visual understanding and flow-matching-based generation.

\begin{figure}[t]
  \centering
  \vspace{-0.5cm}
  \includegraphics[width=0.9\linewidth]{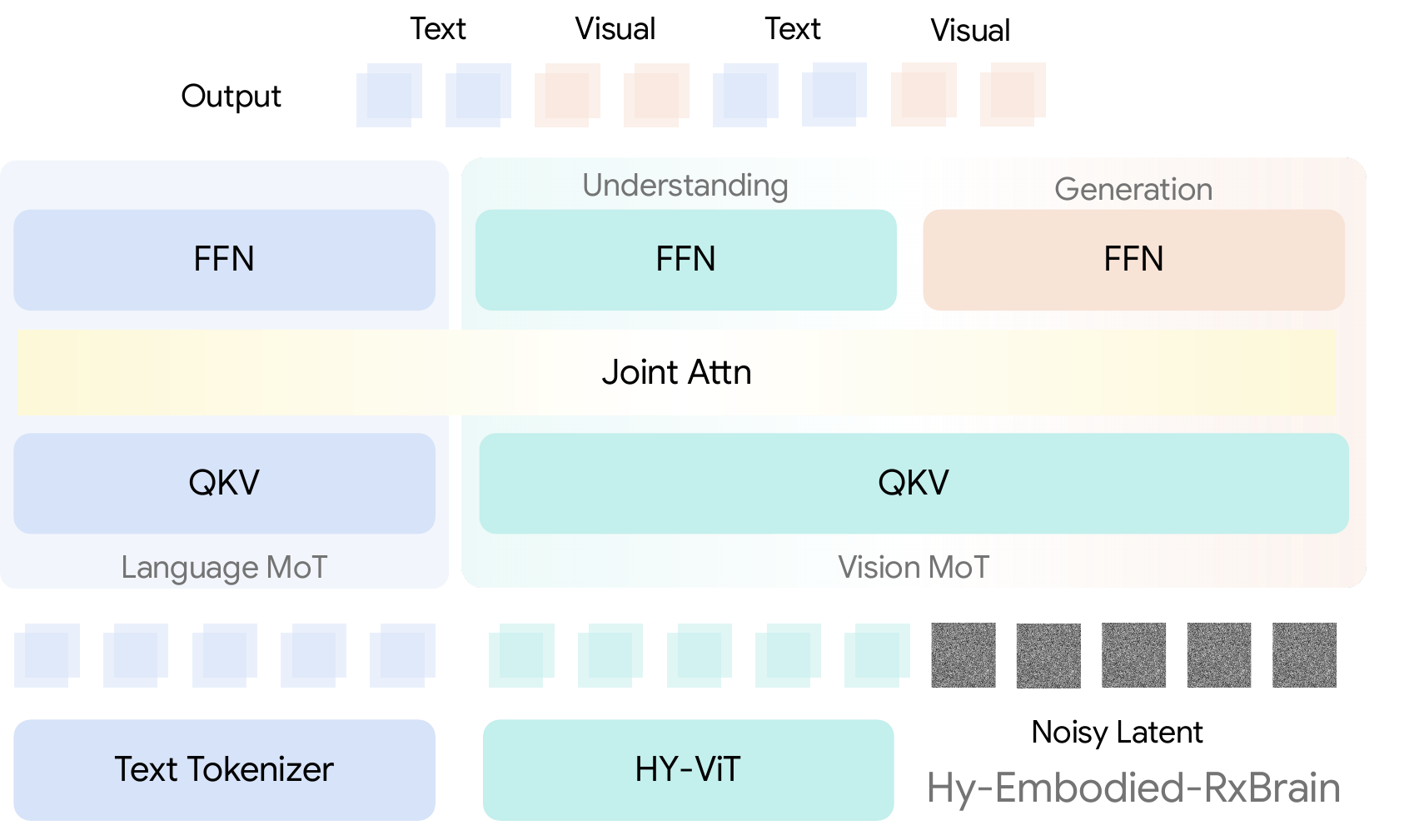}
  \caption{\textbf{Architecture overview of \model.} The model adopts a modality-aware MoT architecture to jointly process visual input tokens, text tokens, and generated vision tokens. Visual tokens are handled by a shared vision backbone with full attention, where understanding and generation tokens share attention projections but are routed to separate FFN experts. Text tokens are processed by the Language Transformer with global causal attention for autoregressive generation. Generated vision tokens are decoded through a VAE decoder. Different token colors indicate visual inputs, text inputs/outputs, and generated visual outputs.}
  \label{fig:mot_arc}
\end{figure}

\subsection{Model Architecture}
\paragraph{Visual representations.}
A key challenge in building a unified model is how to align visual representations for both understanding and generation within a single architecture. RxBrain inherits the Vision Encoder design from HY-Embodied-0.5, where reconstruction ability is introduced during its pre-training~\citep{team2026hy}. Specifically, the Vision Encoder is supervised in the VAE latent space, allowing it to preserve fine-grained image details while maintaining strong semantic feature extraction capability. Inspired by this design, RxBrain adopts HY-ViT 2.0 as the single visual encoder to extract both semantic features and VAE-aligned fine-grained features from visual inputs. To improve image generation quality, we employ a pretrained VAE model from FLUX \citep{flux} as the encoder and decoder for visual generation tokens.

\paragraph{Modality-Aware Mixture-of-Transformers.}

RxBrain adopts a modality-aware Mixture-of-Transformers (MoT) design. Input tokens
are first distinguished by their modality and routed to the corresponding
Transformer branch, while a global self-attention enables cross-modal
information exchange. Structurally, the backbone exposes two modality-specialized
attention branches (language and vision) together with three modality-specific
feed-forward experts for text, visual understanding, and visual generation.

The text pathway and the visual-understanding pathway form the two
modality-specialized branches for language and vision processing. For visual
modeling, the input vision tokens (used for understanding) and the output vision
tokens (used for generation) are processed by the same vision Transformer
and share the same attention projection layers; instead of introducing a
fully separate Transformer for generation, RxBrain adds a dedicated
\emph{visual-generation FFN expert} whose intermediate dimension is widened from
$6{,}144$ to $12{,}288$. A modality-conditioned routing then dispatches
visual-understanding and visual-generation tokens to their respective experts, so that the two visual functions share
visual attention while preserving modality-specific computation for visual input
understanding and visual output generation.

In terms of capacity, the base RxBrain model contains \textbf{6.2\,B} parameters
in total. The text and visual-understanding branches are of roughly equal size
($\sim$\,1.5\,B each), the widened visual-generation expert contributes about
$2.4$\,B, and the remaining parameters lie in the shared and peripheral
modules---the SigLIP vision tower ($\sim$\,0.45\,B), the tied input/output token
embedding ($\sim$\,0.25\,B), and the lightweight image flow-matching head
($\sim$\,7\,M); image latents are produced by an external, frozen VAE tokenizer
(83.8\,M) outside the backbone.

\subsection{Multimodal Generation Formulation}
The formulation is designed to support three generation modes for embodied tasks: autoregressive text generation, future multi frames visual prediction, and interleaved text-vision generation for embodied planning.

\begin{figure}[h]
  \centering
  \includegraphics[width=1.0\linewidth]{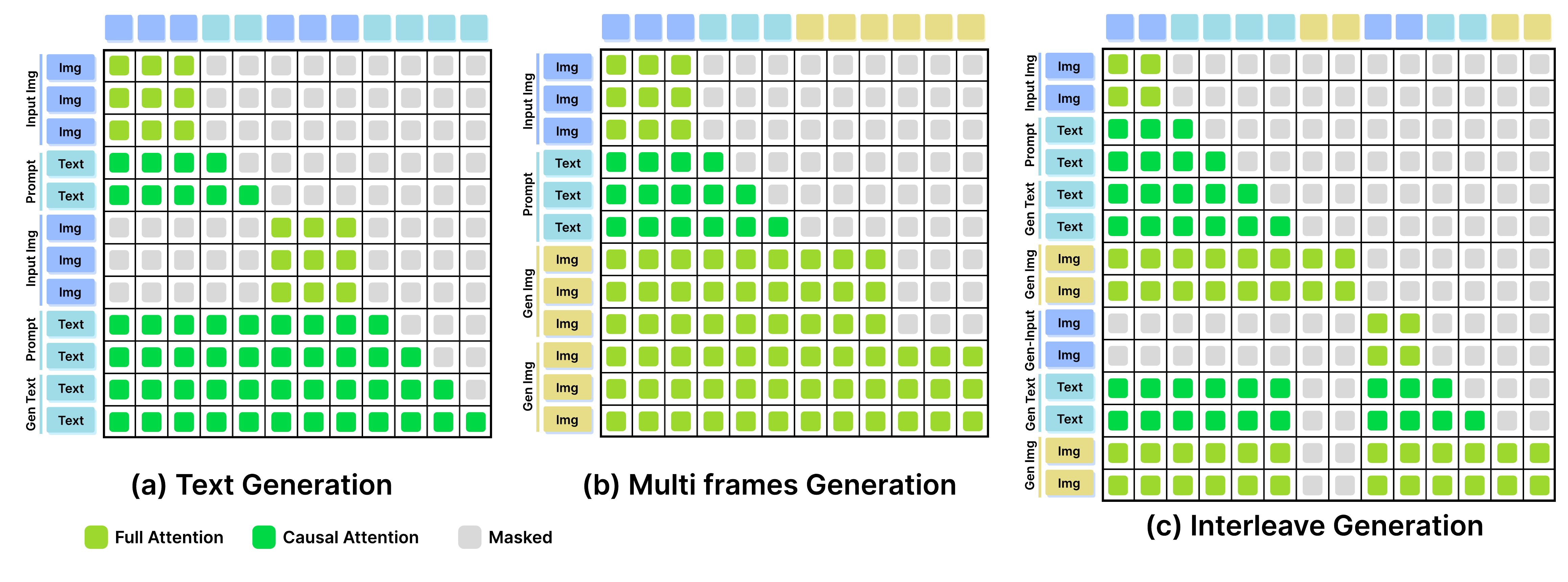}
  \caption{
  \textbf{Hybrid attention patterns in \model.}
  RxBrain adopts hybrid attention masks for unified multimodal understanding and generation. For text generation (a), input images are encoded with bidirectional visual attention, while text tokens follow causal attention for autoregressive decoding. For multi frames generation (b), input images and text prompts follow the same pattern, while generated frames use bidirectional attention within each frame and causal attention across frames. For interleaved generation (c), text is generated autoregressively before image generation; generated images are then re-encoded as visual input for next steps.
  }
  \label{fig:attention_mask}
\end{figure}

\paragraph{Visual Prefix Representation.}
\model\ processes visual tokens with a dedicated intra-image bidirectional attention mechanism in addition to the standard causal attention computation. Specifically, after computing causal attention over the full multimodal sequence, the model performs an additional bidirectional self-attention operation among the visual tokens belonging to each individual image. The resulting intra-image visual representations are then used to replace the corresponding outputs of the causal attention branch at the visual token positions. This design enables each image to achieve global information exchange among its own visual tokens while preserving the causal structure across the multimodal sequence and the independence between different visual observations.

\paragraph{Text Generation.}
For language outputs, RxBrain follows a standard autoregressive generation formulation. Given the multimodal prefix, including language instructions and visual observations, text tokens are generated sequentially with causal attention and optimized through next-token prediction. This enables the model to answer questions about embodied observations and produce language-based reasoning conditioned on both textual and visual context.

\paragraph{Multi Frames Generation.}
\model\ formulates world state prediction tasks as the generation of four future image frames. Given historical observations and task context as the prefix, the model predicts the future frames in the VAE latent space using a flow-matching objective. During generation, the four frames are denoised in parallel for efficiency, while the internal attention mask preserves the causal structure of the visual sequence. As a result, each generated frame can attend to its own frame tokens and all preceding prefix tokens, including historical observations, text context, and earlier generated frames. This design allows \model\ to perform parallel visual denoising while maintaining causal temporal dependencies for future multi-frame prediction.
At the interface level, \model\ represents the four future frames as consecutive latent spans within a single \texttt{<Video>}\,$\cdots$\,\texttt{</Video>} wrapper.

\paragraph{Interleave Generation.}
For joint subgoal planning, \model\ adopts an interleave generation scheme that
jointly produces text reasoning and visual state prediction within a autoregressive process with
the trajectory decoded left-to-right under causal attention:
\begin{center}
\texttt{<assistant>}\;
$r_1$\;\texttt{<Image>}\,$\mathbf{z}_{1,1:N}$\,\texttt{</Image>}\;
$r_2$\;\texttt{<Image>}\,$\mathbf{z}_{2,1:N}$\,\texttt{</Image>}\;
$\cdots$\;\texttt{<eos>}.
\end{center}
Here, $r_i$ denotes the reasoning text at planning step $i$, and
$\mathbf{z}_{i,1:N}$ denotes the flow-matched latent representation of the
imagined frame. The reasoning tokens are generated through standard
autoregressive next-token prediction, while the \texttt{<Image>} token acts as a
learned modality transition token that allows the model to decide when visual
imagination should be performed.

After generating \texttt{<Image>}, the model synthesizes the corresponding latent
span $\mathbf{z}_{i,1:N}$ through flow matching in the VAE latent space,
conditioned on the preceding multimodal context within the full attention. Once the frame is synthesized, it is re-encoded by ViT visual encoder and appended to the context, after which the
\texttt{</Image>} token is automatically inserted and autoregressive decoding
continues with the next reasoning step.

%% file: content/data.tex
\section{Joint Planning Data Construction}
\label{sec:data}

Learning joint textual-visual planning requires supervision that captures both task semantics and the physical state changes produced during execution. Standard text-to-image pairs teach open-domain visual synthesis~\citep{sd3,flux,ldm}, but they do not capture how embodied tasks unfold over time, including object motion, hand or gripper interactions, and intermediate visual states. We therefore use embodied task videos collected from robots, wrist-mounted human demonstrations, simulation, and egocentric human activity. Processing these videos produces joint text-visual planning samples that align textual planning steps with observed state transitions and provide supervision for world state prediction and subgoal planning.

\begin{figure}[t]
  \centering
  \includegraphics[width=0.95\linewidth]{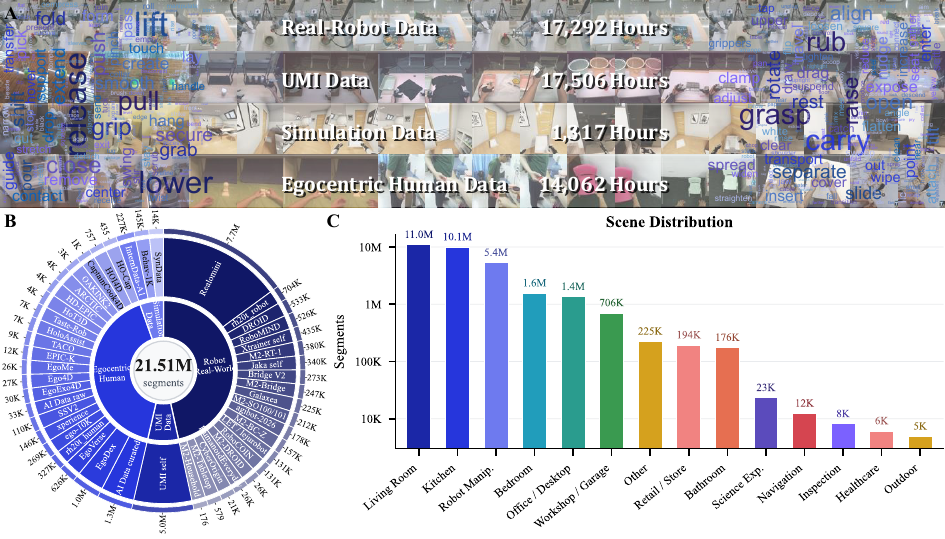}
 \caption{\textbf{Data construction overview and statistics.}
This figure summarizes the verified post-training corpus by scale, source composition, and scene coverage. Panel A reports total video hours across four data categories: Real-Robot Data, UMI Data, Simulation Data, and Egocentric Human Data. Panel B shows the source composition, with source categories in the inner ring and individual datasets in the outer ring. Panel C ranks the fourteen top-level scene categories by segment count.}
  \label{fig:data-statistics}
\end{figure}

Our corpus comprises four source categories, namely Real-Robot Data, UMI Data, Simulation Data, and Egocentric Human Data. Real-Robot Data and UMI Data include self-collected sources, while detailed descriptions of all sources are provided in \Secref{app:dataset-descriptions}. \Figref{fig:data-statistics} summarizes the verified corpus, which spans $46$ processed source splits and $50{,}177.2$ hours and contains $21{,}506{,}919$ trainable segments retained from approximately $28.61$\,M candidates at a reported $75.18\%$ pass rate. Category-level duration, source and dataset composition, and scene coverage are reported in \Figref{fig:data-statistics}A, \Figref{fig:data-statistics}B, and \Figref{fig:data-statistics}C, respectively, with finer scene taxonomies provided in \Secref{app:taxonomy-sunburst}.

To turn raw videos into usable supervision, we process all sources with a three-stage pipeline: Temporal Segment Annotation, Quality Verification, and Segment Structuring. This pipeline produces four post-training tasks: continuous prediction, planning, high-level planning, and final-state imagining. These tasks provide supervision at different levels of granularity, from local state transitions to higher-level planning structures.

\subsection{Temporal Segment Annotation}
\label{sec:data-segmentation}

\begin{figure}[t]
  \centering
  \includegraphics[width=\textwidth]{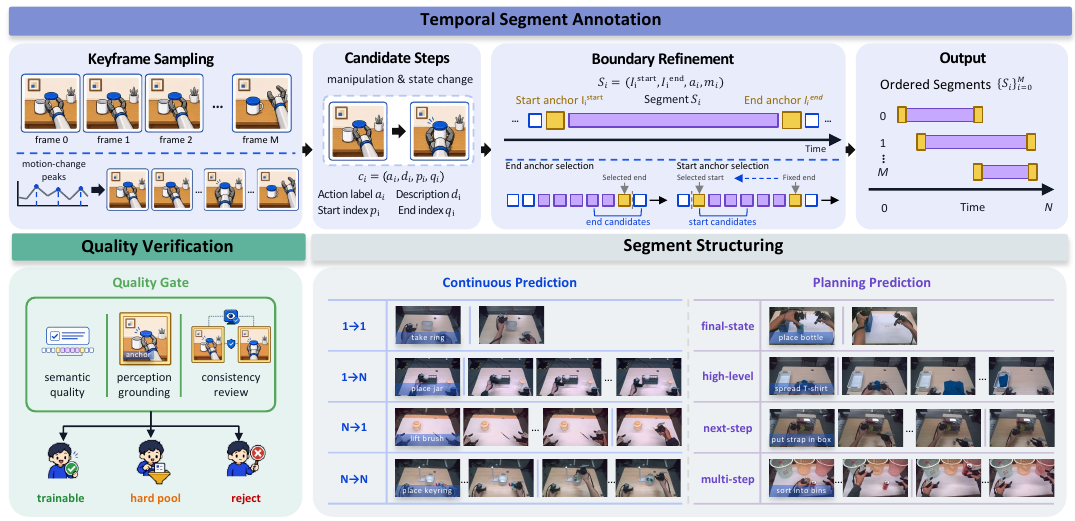}
  \caption{\textbf{Data construction pipeline.}
Raw embodied task videos are converted into joint text-visual planning samples through three stages: Temporal Segment Annotation, Quality Verification, and Segment Structuring. The final samples are organized into L0--L3 tasks, covering intra-step world state prediction, step-level planning, subgoal planning, and final-state imagination.}
  \label{fig:data-pipeline}
\end{figure}

Temporal Segment Annotation converts long, untrimmed embodied task videos into localized planning steps, as shown in \Figref{fig:data-pipeline}. Each localized step contains a short step name, a detailed textual description, and two visual anchors: a start frame and an end frame that make the physical state change visible. We denote the multimodal large language model used for semantic decisions as $\Phi$. The overall process first samples visually informative keyframes, then asks $\Phi$ to propose candidate planning steps, and finally refines the start and end boundaries of each step. A raw video is an ordered sequence $V=(f_1,\dots,f_N)$ recorded at $\nu$ frames per second, with timestamp $\tau_j=j/\nu$. We define an \newterm{atomic planning step} as one localized step in a task whose effect can be observed through a clear visual state change. A localized segment is represented as \begin{equation} S_i=\bigl(I_i^{\mathrm{start}},\,I_i^{\mathrm{end}},\,a_i,\,d_i,\,m_i\bigr), \label{eq:segment} \end{equation} where $I_i^{\mathrm{start}}$ and $I_i^{\mathrm{end}}$ are the start and end frames, $a_i$ is a short step name, $d_i$ is a detailed description of the step, and $m_i$ records source, viewpoint, index, and temporal metadata. The annotation of a video is a temporally ordered sequence of segments $\{S_i\}_{i=1}^{M}$. Neighboring segments may overlap in time, which allows the same video interval to be reused when multiple planning steps share visual context, while their indices still follow the chronological order of the task. \Eqref{eq:segment} is the unit consumed by Quality Verification and Segment Structuring. 

\paragraph{Keyframe sampling.} Long videos often contain many static intervals, so we allocate the MLLM context to frames where visible changes are more likely to occur. Instead of uniformly sampling the whole video, we first decode the clip at a reduced frame rate and score adjacent frames by visual difference. After smoothing and pruning nearby peaks, we keep a compact set of keyframes that covers both major visual changes and long temporal gaps. The first and last frames are also included, and the final anchor set is capped to fit the global keyframe budget $\gG$. This source-agnostic sampler is applied to all data sources with the same empirical constants. 

\paragraph{Candidate step proposal.} Given the ordered global keyframes $\gG$, $\Phi$ first judges whether the clip is suitable for annotation. Clips that are too dark, heavily cluttered, dominated by tiny objects, or otherwise difficult to interpret are marked as unannotatable and produce no steps. For annotatable clips, $\Phi$ returns a short clip summary, a high-level goal, the camera viewpoint, and an ordered list of candidate planning steps. Each candidate step $c_i=(a_i,d_i,p_i,q_i)$ contains a short step name $a_i$, a detailed description $d_i$, and coarse start and end indices $(p_i,q_i)$ in $\gG$. The prompt asks $\Phi$ to follow the atomic planning step definition and to keep the coarse candidates in chronological order. 

\paragraph{Boundary refinement.} The candidate ranges provide a task-level prior, but their sparse keyframes may not capture the exact start or completion of a state change. We therefore refine each coarse segment locally. Around the coarse end, we sample a small set of candidate frames and ask $\Phi$ to select the first frame where the completed state is visible. Around the coarse start, we sample another local candidate set and ask $\Phi$ to select the start frame, using the chosen end frame as reference. This bidirectional refinement gives each planning step a clear pair of visual anchors while allowing adjacent steps to overlap when they share temporal context. The final start and end frames are rendered at native resolution for downstream training. Additional implementation details are provided in \Secref{app:segment-method-details}. 

\paragraph{Output.} The annotation stage outputs a temporally ordered sequence of localized planning segments $\{S_i\}_{i=1}^{M}$ for each source video. Each segment contains start and end visual anchors, a short step name, a detailed description, and metadata. Quality Verification (\Secref{sec:data-quality}) filters these segments, and Segment Structuring (\Secref{sec:data-organization}) organizes the retained segments into model-ready training samples.

\subsection{Quality Verification}
\label{sec:data-quality}

Quality Verification uses an MLLM-based verifier to decide whether each localized segment $S_i$ from \Eqref{eq:segment} is suitable for training. A valid segment should contain a clear visual state transition between the start anchor $I_i^{\mathrm{start}}$ and the end anchor $I_i^{\mathrm{end}}$, and its textual description should correctly describe the observed planning step. This verification ensures that the retained segments can supervise world state prediction and joint subgoal planning, rather than introducing noisy text-visual pairs.

The verifier checks three aspects in one MLLM call: semantic quality, visual grounding, and consistency between the text and the visual transition. Each rejected segment is associated with a specific review item, making the filtering decision interpretable instead of relying only on an opaque score. The complete review protocol and item-level rejection criteria are provided in \Secref{app:data-review-items} and Table~\ref{tab:data-review}.

For semantic quality, each segment is scored on a six-level scale from $5$ (clean and usable) to $0$ (missing or unreadable). Segments with scores $4$--$5$ are kept only if they also pass all consistency checks, while segments with scores $0$--$3$ are rejected. We additionally apply a simple near-duplicate filter that removes pairs whose start and end anchors are more than $99.5\%$ identical, since such pairs usually do not contain a meaningful state change.

When the visual transition is valid but the textual description $d_i$ is inaccurate or incomplete, the verifier corrects $d_i$ instead of discarding the segment. We also allow certain apparent object disappearances when they are consistent with the planning step, such as pick-up, move-away, or withdraw behaviors. After verification, $75.18\%$ of candidate segments are retained for training.

\subsection{Segment Structuring}
\label{sec:data-organization}

Segment Structuring organizes verified segments into joint text-visual planning samples for \model{}, following the packed-sequence format in \Secref{sec:arch}. The goal is to construct training samples at different temporal scales, from the state change inside a single planning step to the final state of a complete task. Each sample contains a task wrapper, a goal condition, textual planning information, and visual context or target frames. We use $t_i$ to denote the text condition associated with segment $i$, which can be a short step name $a_i$, a detailed description $d_i$, or an LLM-generated subgoal, depending on the task.

\begin{figure}[t]
  \centering
  \includegraphics[width=\textwidth]{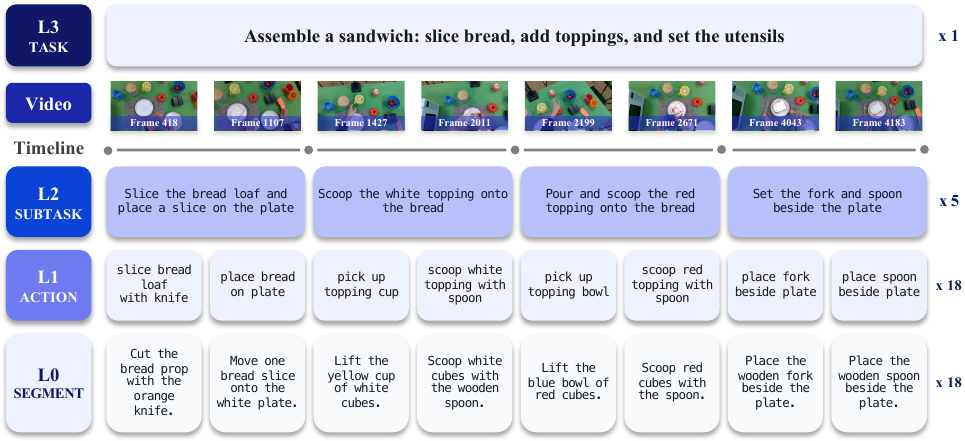}
  \caption{\textbf{Hierarchical data construction pipeline.}
  Verified planning segments are organized into four levels of supervision.
  L0 learns the visual state change inside a single planning step, while L1
  links consecutive steps for step-level joint planning. L2 groups neighboring
  steps into subgoals, and L3 connects the initial observation with the final
  task state.}
  \label{fig:data-structuring-pipeline}
\end{figure}

We organize the samples into four post-training tasks, corresponding to the L0--L3 levels in \Figref{fig:data-structuring-pipeline}. L0 and L1 are both built from the verified atomic planning segments. L0 uses each segment independently to learn intra-step visual state changes, while L1 links consecutive segments to learn step-level planning. Above this shared segment layer, L2 groups adjacent steps into higher-level subgoals, and L3 summarizes the whole video into a final task goal. This hierarchy allows \model{} to learn world state prediction and joint subgoal planning at multiple levels of granularity.

\paragraph{L0 segment.}
\label{sec:data-org-continuous}

L0 samples model the visual change inside a single atomic planning step. The basic form takes the start frame $I_i^{\mathrm{start}}$ and a text condition $t_i$ as input, and predicts the end frame $I_i^{\mathrm{end}}$:
$$
[I_i^{\mathrm{start}}, t_i] \rightarrow I_i^{\mathrm{end}}.
$$
We also create variants with richer visual context or denser targets. Some samples prepend pre-step history frames $I_i^{\mathrm{pre}}$, while others include intermediate frames $I_i^{\mathrm{mid}}$ as additional targets. These variants teach the model not only the final outcome of a planning step, but also the intermediate visual dynamics leading to that outcome.

\paragraph{L1 step-level planning.}
\label{sec:data-org-planning}

L1 samples connect consecutive atomic planning steps. Given visual context and a goal condition, the model predicts a sequence of future planning steps and their corresponding visual outcomes. Each predicted step is represented by a text description and an end frame:
$$
[C_{1:m}, g_{\mathrm{seg}}] \rightarrow [T_1, I_1^{\mathrm{end}}, \dots, T_k, I_k^{\mathrm{end}}],
$$
where $C_{1:m}$ denotes visual context frames, $g_{\mathrm{seg}}$ denotes the goal condition for the segment window, and $T_i$ denotes the predicted step text. This task teaches the model to jointly produce textual planning steps and visual state transitions.

We construct three layouts for L1. The first is goal-to-future planning, where the model starts from the initial visual context and predicts multiple future steps. The second is history-to-next-step planning, where the model observes completed steps and predicts the next step and its resulting state. The third is history-to-future planning, where the model conditions on several completed steps and predicts a sequence of subsequent steps. These layouts expose the model to both forward planning from a goal and continuation planning from partial progress.

\paragraph{L2 subgoal planning.}
\label{sec:data-org-hlp}

L2 samples operate at the subgoal level. Starting from the sequence of verified planning steps, an LLM groups adjacent steps into a smaller number of high-level subgoals. Each subgoal is paired with a boundary frame that shows the state after the subgoal has been completed. The construction mirrors L1, but replaces step-level texts and end frames with subgoal descriptions and subgoal boundary frames:
$$
[B_{1:m}, g_{\mathrm{seg}}] \rightarrow [H_1, B_1, \dots, H_k, B_k],
$$
where $B_{1:m}$ denotes visual context at the subgoal level and $H_j$ denotes a high-level subgoal description. This task encourages \model{} to plan over coarser task stages rather than only local step transitions.

\paragraph{L3 final-state imagination.}
\label{sec:data-org-fsi}

L3 samples model the final outcome of the whole task. Given the initial frame $I_0^{\mathrm{start}}$ and a concise task goal $g_{\mathrm{task}}$, the model predicts the final frame $I_T^{\mathrm{end}}$:
$$
[I_0^{\mathrm{start}}, g_{\mathrm{task}}] \rightarrow I_T^{\mathrm{end}}.
$$
The task goal is generated by an LLM from the full sequence of subgoals and summarizes the intended final objective of the video. Each video contributes at most one L3 sample. This task teaches the model to imagine long-range task outcomes without requiring every intermediate step to be specified.

%% file: content/training_fix.tex
\section{Training}
\label{sec:training}

\subsection{Training Objective}
\label{sec:training-objective}

RxBrain is optimized with a unified multimodal objective that jointly trains
language understanding, visual generation, and interleaved embodied generation.
Different generation modes share the same MoT backbone while adopting
modality-specific supervision objectives.

\paragraph{Text Generation Objective.}
For language generation, RxBrain follows the standard autoregressive formulation.
Given a multimodal context $c$ and a target text sequence
$y=(y_1,\dots,y_T)$, we optimize the next-token prediction objective:

\begin{equation}
\mathcal{L}_{\mathrm{text}}
=
-\sum_{t=1}^{T}
\log p_{\theta}(y_t|y_{<t},c).
\end{equation}

The text branch adopts causal attention, where each token attends only to
previous text tokens and available multimodal context.

\paragraph{Image Generation Objective.}
RxBrain generates visual tokens in the VAE latent space using a flow-matching objective. Given a clean VAE-encoded image latent $x_0$ and
Gaussian noise $\epsilon\sim\mathcal{N}(0,I)$, the intermediate latent is
constructed with a linear flow path:

\begin{equation}
x_t=(1-t)x_0+t\epsilon ,
\end{equation}

where $t\in[0,1]$ is sampled from a logit-normal distribution. The generation
branch predicts the velocity field with the flow-matching objective:

\begin{equation}
\mathcal{L}_{\mathrm{flow}}
=
\mathbb{E}_{t,x_0,\epsilon}
\left[
\left\|
v_{\theta}(x_t,t,c)-(\epsilon-x_0)
\right\|_2^2
\right].
\end{equation}

The learned velocity field is used for image synthesis by solving the
corresponding ODE during inference. This objective is applied to both
future-frame prediction and image generation tasks.

\paragraph{Interleave Generation Objective.}
RxBrain is trained to generate interleaved language and visual states for
embodied reasoning. Each training sample is represented as an ordered sequence
of text segments and visual states:

\begin{equation}
S=\{(x_1,I_1),(x_2,I_2),\dots,(x_N,I_N)\},
\end{equation}

where $x_i$ denotes the reasoning text segment and $I_i$ denotes the
corresponding visual state. At each step $k$, RxBrain predicts the next text
and visual state conditioned on the preceding multimodal context:

\begin{equation}
h_k=\{x_{<k},\tilde{I}_{<k}\},
\end{equation}

where $\tilde{I}_{<k}$ denotes the visual representations of previous states.
The text generation follows an autoregressive objective:

\begin{equation}
\mathcal{L}_{\mathrm{text}}
=
-\sum_k \log p_{\theta}(x_k|h_k).
\end{equation}

For visual generation, the model additionally conditions on the current
reasoning text $x_k$:

\begin{equation}
\mathcal{L}_{\mathrm{vision}}
=
\mathcal{L}_{\mathrm{flow}}
(I_k|h_k,x_k),
\end{equation}

where $\mathcal{L}_{\mathrm{flow}}$ is the flow-matching objective
described above.

To reduce the discrepancy between training and inference contexts, we employ
an \textbf{inference-aware visual state construction strategy}. During
training, the previous visual states are obtained from ground-truth images
through the same reconstruction and encoding pipeline used during inference:

\begin{equation}
\tilde{I}_i =
E_{\mathrm{vis}}
\left(
D_{\mathrm{VAE}}
\left(
E_{\mathrm{VAE}}(I_i)
\right)
\right),
\end{equation}

where $D_{\mathrm{VAE}}$ and $E_{\mathrm{VAE}}$ denote the decoder and encoder of VAE, $E_{\mathrm{vis}}$ denotes the ViT encoder. During inference, the generated latent state $\hat{z}_i$ is first decoded into
an image and then transformed into visual representation:

\begin{equation}
\tilde{I}_i =
E_{\mathrm{vis}}
\left(
D_{\mathrm{VAE}}(\hat{z}_i)
\right).
\end{equation}

By aligning the visual state construction process between training and
inference, the model learns to reason over visual representations consistent
with those encountered during generation, while reducing the impact of error accumulation from interleave rollout.

The overall interleaved generation objective is:

\begin{equation}
\mathcal{L}_{\mathrm{interleave}}
=
\sum_k
\left(
\lambda_t
\mathcal{L}_{\mathrm{text}}(x_k|h_k)
+
\lambda_v
\mathcal{L}_{\mathrm{vision}}(I_k|h_k,x_k)
\right),
\end{equation}
where $\lambda_t$ and $\lambda_v$ denote the weighting coefficients for the text generation loss and the vision generation loss respectively.

\subsection{Training Stage}
\label{sec:training-stage}

\model{} is trained with a two-stage curriculum that progressively transforms
the model from a general multimodal foundation model into an unified embodied model with joint language-visual reasoning and imagination abilities. The first stage focuses on establishing a unified
vision-language generation capability, where the model learns to understand,
reason over, and generate visual content while preserving the underlying
language capability. Starting from this multimodal foundation, the second stage
adapts \model{} to embodied scenarios by introducing world modeling and
long-horizon planning supervision. This progressive training strategy enables
\model{} to acquire general visual knowledge before specializing in embodied
tasks.

\paragraph{Stage 1: Joint Vision-Language Pretraining.}
The first stage jointly trains all components of \model{} using large-scale
image-text pairs and multimodal instruction-following data. The objective of
this stage is to establish a unified representation space that connects
language understanding, visual perception, and visual generation. Image-text
alignment and instruction data provide semantic grounding and visual reasoning
capabilities, while visual generation supervision based on the flow-matching
objective equips the model with the ability to synthesize visual states. By
jointly optimizing understanding and generation objectives, this stage
preserves the general language capability of the base model while extending it
with multimodal reasoning and generation abilities.

\paragraph{Stage 2: Embodied Supervised Fine-tuning.}
The second stage adapts \model{} to embodied scenarios through supervised multimodal generation learning. The
training mixture combines general multimodal understanding data with embodied
data, including embodied instruction-following examples, multi-frame prediction
samples, and interleaved planning trajectories. The embodied instruction data
further enhances the model's ability to understand embodied environments and
generate language responses for reasoning and planning tasks, while the
multi-frame prediction objective improves the model's ability to generate future states. 

For interleaved planning data, each multimodal trajectory is transformed into
multiple text-image generation samples during training, where each generated
visual state is paired with its corresponding language context. This training
strategy enables \model{} to learn interleaved generation over sequential
language and visual states while maintaining compatibility with the inference process.

During this stage, \model{} is optimized with a combination of language
generation, visual generation, and embodied supervision objectives. More
details about training hyperparameters, data mixture, and optimization settings
are provided in Appendix~\ref{app:training-config}.

%% file: content/self_bench.tex
\section{RxBrain-Bench}
\label{sec:xbrain-bench}

\subsection{Overview}

Unified multimodal models for embodied AI are increasingly expected to do more than answer questions about an observed scene or generate a future video from a fixed instruction. To act as embodied agents, interleaving the two modalities helps a lot: generate the next action in language, imagine the visual state produced by that action, use the generated state to decide what to do next, and repeat until the goal is achieved. While existing benchmarks provide valuable tests of visual understanding, embodied reasoning, and video generation, but these abilities are typically evaluated in isolation. To our knowledge, no existing benchmark directly and comprehensively evaluates the full interleaved generation loop, 
evaluating this \emph{interleaved text--visual generation} capability is the primary motivation for \textbf{RxBrain-Bench}.

For embodied reasoning, RxBrain-Bench covers a broad set of evaluation dimensions, including action and trajectory understanding, initial state judgment, visual grounding and localization, complex task planning, simulation-based planning, task completion judgment, error recognition and recovery, task state estimation, spatial relation understanding, and a small portion of multi-view and temporal reasoning. These dimensions systematically evaluate a model's perception, understanding, reasoning, planning, and state judgment abilities in embodied manipulation scenarios.

Based on these dimensions, the benchmark is organized into three tracks: RxBrain-Bench-EVQA, RxBrain-Bench-WorldPred, and RxBrain-Bench-JointPlan.

\subsection{Tracks}

\paragraph{RxBrain-Bench-EVQA.}The RxBrain-Bench-EVQA track evaluates whether models can understand embodied scenes and reason about task-relevant decisions from visual observations. The scenarios are selected from domains with potential commercial value, including industrial, retail, and service environments. Different from general-purpose VQA benchmarks, this track focuses on questions grounded in real-world deployment, where models must reason not only about objects and spatial relations, but also about task efficiency, operational safety, goal satisfaction, and physical constraints. Concretely, it covers several categories of embodied reasoning: high-level planning, which asks models to decompose a goal into executable steps; simulation-based planning, which evaluates multi-round planning by requiring models to update subsequent decisions according to simulated state transitions and intermediate outcomes; fine-grained step reasoning over grasp points, motion trajectories, action selection, temporal order, and multi-view observations; error recovery, which tests whether models can identify abnormal states and choose appropriate responses; and task-completion judgment, which requires models to determine whether the current scene state satisfies the task goal. To make the evaluation discriminative, we design distractors around realistic physical conflicts, production safety rules, and trade-offs between safety and efficiency.

\paragraph{RxBrain-Bench-WorldPred.} The RxBrain-Bench-WorldPred track evaluates visual goal imagination in embodied scenarios. Given a current observation and a natural-language action instruction, a model is required to generate a short future video depicting the physical state transition caused by the action. This tests whether the model can imagine not only what action is being executed, but also how the scene should change as a result. Unlike existing embodied video generation benchmarks~\citep{zhou2025paibench,deng2026rbench} that emphasize long-horizon synthesis, our track focuses on short-horizon future prediction. This setting better matches the closed-loop nature of embodied control, where an agent repeatedly predicts the near future, executes an action, and re-grounds its next decision on new observations. Short-horizon prediction also reduces error accumulation in long rollouts and provides the visual look-ahead needed for step-by-step planning.

\paragraph{RxBrain-Bench-JointPlan.} The RxBrain-Bench-JointPlan track evaluates the core capability targeted by RxBrain-Bench: representing an embodied plan through joint textual reasoning and visual goal imagination. Unlike Embodied VQA, which focuses on understanding and reasoning over the current scene, or Embodied Video Generation, which focuses on predicting the visual consequence of a given action, this track requires a model to use text and images as complementary parts of the same plan. Given a goal and an initial observation, the model must describe the next action in language, generate the corresponding visual state after that action, judge progress toward the goal, and continue this process until the task is completed. In this setting, language specifies the action logic, constraints, and decisions, while visual imagination specifies the intermediate and final world states associated with them. This track therefore evaluates whether a model can move beyond separate understanding or generation tasks and jointly represent what should be done and what physical state should be achieved.

\subsection{Evaluation Metrics}
\label{sec:metrics}

We evaluate the three benchmark tracks using task-specific metrics.
Embodied VQA is evaluated by exact-match accuracy. The two generative tracks
(Embodied Video Generation and Multimodal Embodied Planning) are evaluated
primarily by an \textbf{MLLM-as-judge} using task-specific weighted rubrics,
together with auxiliary fidelity metrics computed against the ground-truth
frames. We use a fixed GPT-5.5 model as the evaluator and require structured JSON outputs for all judgments~\citep{singh2025openai}. Each
criterion is rated on a five-point scale, and the reported score is averaged
over $k=2$ independent evaluations.

\paragraph{EVQA.}
All questions are multiple-choice with a single correct answer among physically
grounded distractors. We report exact-match \textbf{accuracy}, both overall and
macro-averaged across the four question categories (high-level planning,
fine-grained step reasoning, error recovery, and task-completion judgment).

\paragraph{WorldPred.}
Given the current observation and an action instruction, the model predicts
future frames. We evaluate each generated trajectory using five criteria:
\emph{Observation Continuity (OC)}, \emph{Action Correctness (AC)},
\emph{Goal Completion (GC)}, \emph{Temporal Plausibility (TP)}, and
\emph{Ground-Truth Trajectory Match (TM)}.
The overall score is the weighted sum
\[
S_{\mathrm{gen}}
=
0.15\,\mathrm{OC}
+
0.30\,\mathrm{AC}
+
0.20\,\mathrm{GC}
+
0.20\,\mathrm{TP}
+
0.15\,\mathrm{TM}.
\]
Action Correctness and Temporal Plausibility receive the largest weights, as
they measure whether the predicted future both follows the commanded action and
remains physically coherent over time. We additionally report PSNR and DINO
similarity as auxiliary frame-level fidelity metrics.

\paragraph{JointPlan.}
Planning is evaluated in a fully autoregressive (free-running) setting, where
the generated reasoning text and imagined visual state at each step are fed
back as inputs for subsequent prediction. This protocol evaluates true
closed-loop planning and naturally captures error accumulation over long
horizons.

Each planning trajectory is assessed using six criteria:
\emph{Observation Understanding (OU)},
\emph{Subtask Planning (SP)},
\emph{Goal-Image Correctness (GI)},
\emph{Subtask--Goal Consistency (SG)},
\emph{Chain Completion (CC)}, and
\emph{Image Similarity (IS)}.
The weighted planning score is defined as
\[
S_{\mathrm{plan}}
=
0.10\,\mathrm{OU}
+
0.25\,\mathrm{SP}
+
0.25\,\mathrm{GI}
+
0.20\,\mathrm{SG}
+
0.10\,\mathrm{CC}
+
0.10\,\mathrm{IS}.
\]
The first five criteria are rated by the MLLM judge, while Image Similarity is
computed as the DINO cosine similarity between each generated visual state and
its corresponding ground-truth state. The three step-level criteria (SP, GI,
and SG) are averaged across the entire planning trajectory. We report both the
overall planning score and the per-criterion breakdown, enabling separate
analysis of failures in textual reasoning and visual prediction.
The full generation-side evaluation protocol is provided in
\Secref{app:genbench-protocol}. It specifies the exact task inputs and held-out
set sizes for both generative tracks, the GPT-5.5 VLM-as-judge configuration
(multi-image prompting, structured JSON scoring, and $\text{votes}{=}2$
averaging), the complete criterion definitions and weights of the two rubrics
(Table~\ref{tab:il-rubric} for JointPlan and Table~\ref{tab:mf-rubric} for
WorldPred), the local perceptual pass that supplies the DINO image-similarity
term, and how each baseline is composed into a comparable interleaved planner or
video generator for a like-for-like comparison.

%% file: content/evaluation.tex
\section{Evaluation}
\label{sec:evaluation}
We evaluate \model{} along four axes: (i) its core text-to-image generation
ability; (ii) its embodied and spatial \emph{understanding} across a broad suite of
benchmarks; (iii) its world state prediction ability based on multi frames generation and (iv) its joint subgoal planning ability. For the first two
axes we report standard benchmark and RxBrain-EVQA comparisons against representative open and
released models. For world state prediction and joint subgoal planning we report our evaluation results on RxBrain-WorldPred and RxBrain-JointPlan benchmarks.

\subsection{Evaluation Tasks}
\label{sec:exp-tasks}
We organize the evaluation along the four axes introduced above, and describe
the benchmarks, baselines, and metrics used for each before presenting results
in \Secref{sec:exp-results}.

\paragraph{Text-to-image generation.}
We first assess \model{}'s core text-to-image generation ability. We use
GenEval~\citep{ghosh2023genevalobjectfocusedframeworkevaluating}, a standard
benchmark of prompt-faithful, \emph{open-domain}
text-to-image synthesis rather than any embodied skill, to verify that the
unified model preserves general-purpose, world-knowledge image generation. We
compare against the generation-specialist baseline Bagel~\citep{bagel} and the
other generation-capable model Cosmos-3-Nano~\citep{agarwal2026cosmos}.

\paragraph{Embodied and spatial understanding.}
Second, we evaluate embodied and spatial \emph{understanding} across a broad
suite of public benchmarks together with our self-built RxBrain-Bench-EVQA. The
public benchmarks are grouped into three families: \emph{embodied and spatial
reasoning} (ERQA~\citep{team2025gemini},
EmbSpatial~\citep{du2024embspatialbenchbenchmarkingspatialunderstanding},
CV-Bench~\citep{tong2024cambrian1}, SAT~\citep{ray2024sat},
DA-2k~\citep{depth_anything_v2}); \emph{robot affordance,
trajectory and placement} (Share-Robot-Affordance and
Share-Robot-Trajectory~\citep{ji2025robobrain},
RefSpatial~\citep{zhou2025roborefer}, Where2Place~\citep{yuan2024robopoint}, and
the \emph{context}/\emph{compatibility}/\emph{configuration} splits of
RoboSpatial-Home~\citep{song2025robospatial}); and \emph{multi-view and 3D
understanding} (MindCube~\citep{yin2025mindcube},
MMSI-Bench~\citep{yang2025mmsi}, ViewSpatial~\citep{li2025viewspatial},
3DRSBench~\citep{ma20243dsrbench}, SITE-Bench-Image~\citep{wang2025site},
VSI-bench~\citep{yang2024vsi}). \model{} is compared against representative open
and released vision--language and embodied models; higher is better for all
metrics.

To probe physical-world understanding beyond generic image QA, we additionally
construct RxBrain-Bench-EVQA, a self-built multiple-choice VQA benchmark
grounded in \emph{real} robot-manipulation episodes. Each question is built from
frames sampled from task-execution videos and posed as a multiple-choice
problem---four options, or two for binary yes/no judgments---with hand-written
distractors, and answered by a single option letter. Questions are grounded in
one to nine images (median three, mean $4.5$)---covering single observations,
multiple synchronized camera views, and temporally ordered frames---so that
answering requires genuine visual and physical reasoning rather than language
priors. The benchmark comprises \textbf{1{,}123} multiple-choice questions
spanning ten categories, which we group into four families: (i) \emph{perception
and state}---State Estimation and State Understanding; (ii) \emph{spatial and
grounding}---Spatial Reasoning and Pointing (localizing a queried target); (iii)
\emph{action, trajectory and temporal}---Action Reasoning, Trajectory Reasoning,
and Temporal Reasoning (recovering the chronological order of shuffled frames);
and (iv) \emph{outcome and planning}---Success Detection, Error Recovery,
Complex Planning, and Simulation Planning. Simulation Planning is a separate
multi-round task-planning split with \textbf{258} samples, evaluated by a
weighted MLLM judge. Together, the benchmark contains \textbf{1{,}381} evaluated
items. Items come from a model-assisted, verified pool together with a
human-authored cognition set, and every item is checked to have a unique,
image-grounded correct answer; the temporal-ordering split is additionally
curated with a strong external VLM. We report top-1 accuracy for the
multiple-choice subtasks and the weighted judge score for Simulation Planning.

\paragraph{World state prediction.}
Third, we evaluate world state prediction on RxBrain-Bench-WorldPred: given the
current observation and an action instruction, the model predicts a
short-horizon ($4$-frame) future. Since no released \emph{unified} model
supports instruction-conditioned short-horizon video generation in our
closed-loop setting, we construct strong baselines from released components---the
general-purpose video model \emph{Wan2.2-TI2V-5B}~\citep{wan2025} and the
embodied world model \emph{Cosmos3-Nano}~\citep{agarwal2026cosmos}. All
evaluation trajectories are held out from training, and
we report the weighted MLLM-judge score $S_{\mathrm{gen}}$ (Section~\ref{sec:metrics})
as the primary metric.

\paragraph{Joint subgoal planning.}
Finally, we evaluate joint subgoal planning on RxBrain-Bench-JointPlan, which
requires interleaved text-and-image planning under a fully autoregressive
(free-running) closed-loop rollout. Since no released unified model supports
this setting either, we again build strong baselines from released components: a
modular \emph{Qwen-Agent} (Qwen3-VL-2B~\citep{bai2025qwen3vl} as the reasoner
with Qwen-Image-Edit as the image generator), an agent built upon the omni model
\emph{Cosmos3-Nano}~\citep{agarwal2026cosmos}, and the unified
\emph{BAGEL-7B-MoT}~\citep{bagel} run as a planning agent. All
evaluation trajectories are held out from training, and we report the weighted
MLLM-judge score $S_{\mathrm{plan}}$ (Section~\ref{sec:metrics}) as the primary metric.

\subsection{Evaluation Results}
\label{sec:exp-results}
\begin{table}[h]
    \centering
    \caption{\textbf{Main comparison on embodied reasoning and image generation benchmarks.}
    \model{} is compared against representative open and released vision--language
    and embodied models across 19 benchmarks, grouped into four families:
    text-to-image generation, embodied and spatial reasoning, robot affordance/
    trajectory/placement, and multi-view/3D understanding. GenEval measures general
    open-domain text-to-image generation to demonstrate model's world knowledge generation ability. Larger models,
    family-specific variants, sparsely evaluated models, diagnostic splits, and
    benchmarks without reported \model{} results are included in the full comparison
    in the appendix. Benchmarks are listed as rows and models as columns; higher is
    better for all metrics and ``--'' marks entries that are not applicable or not
    reported. RoboSpatial-Home is split into its \emph{context}/\emph{compatibility}/
    \emph{configuration} sub-rows. Best result per row is in \textbf{bold}, and
    second-best result is underlined.}
    \label{tab:embodied-bench}
    \setlength{\tabcolsep}{5pt}
    \renewcommand{\arraystretch}{1.0}
    \begin{adjustbox}{max width=\linewidth}
    \small
    \begin{tabular}{l*{6}{c}}
      \toprule
      
        & {Qwen3.5}
        & {Qwen3.5}
        & {Bagel}
        & {RoboBrain2.5}
        & {Cosmos-3-Nano}
        & {\textbf{\model}} \\
    Benchmark & 4B & 2B & 14B(A7B) & 4B  & 16B(A8B) & 6B \\
      \midrule
      \rowcolor{annobg}
      \multicolumn{7}{l}{\textbf{Text-to-image generation}} \\
      GenEval                      & -- & -- & \underline{82} & -- & 71.68 & \textbf{82.4} \\
      \addlinespace[4pt]
      \rowcolor{annobg}
      \multicolumn{7}{l}{\textbf{Embodied \& spatial reasoning}} \\
      ERQA                         & \underline{45.8} & 39.8 & 39.5 & 35.8 & \textbf{46} & 42.8 \\
      EmbSpatial                   & 74.2 & 64.3 & 67.0 & 80.2 & \underline{82.1} & \textbf{82.3} \\
      CV-Bench                     & 85.9 & 78.98 & 77.4 & 87.4 & \underline{87.9} & \textbf{88.59} \\
      SAT                          & \textbf{78.33} & 63.67 & 74.3 & 66 & 73.3 & \underline{74.33} \\
      DA-2k                        & 68.5 & 63.9 & 62.9 & \underline{82.8} & 78.9 & \textbf{83.4} \\
      \addlinespace[4pt]
      \rowcolor{annobg}
      \multicolumn{7}{l}{\textbf{Robot affordance, trajectory \& placement}} \\
      Share-Robot-Affordance       & 24.75 & 18.09 & 3.2 & \textbf{50.31} & \underline{27.97} & 11.74 \\
      Share-Robot-Trajectory       & 29.41 & 28.51 & 11.78 & 27.74 & \underline{39.02} & \textbf{51.13} \\
      RefSpatial                   & 34.27 & 24.22 & 8.62 & \textbf{62.5} & \underline{59.5} & 34.08 \\
      Where2Place                  & 44.28 & 15.09 & 8.83 & \underline{52.47} & \textbf{60.64} & 39.75 \\
      RoboSpatial-Home~(context)   & 7.22 & 6.74 & 6.15 & \textbf{56.56} & \underline{31.22} & 28.28 \\
      RoboSpatial-Home~(compatibility)  & 50.5 & 32.38 & 42.86 & 36.19 & \textbf{71.43} & \underline{67.62} \\
      RoboSpatial-Home~(configuration)  & 82.1 & 71.54 & 73.98 & 83.74 & \underline{86.18} & \textbf{86.28} \\
      \addlinespace[4pt]
      \rowcolor{annobg}
      \multicolumn{7}{l}{\textbf{Multi-view \& 3D understanding}} \\
      MindCube                     & 38.5 & 31.9 & 37.6 & \underline{43.7} & 36.7 & \textbf{47.5} \\
      MMSI-Bench                   & 31 & 26.6 & 29.8 & 28.3 & \underline{34.6} & \textbf{35.8} \\
      ViewSpatial                  & 42.3 & 36.6 & 38.6 & 41.5 & \textbf{54.1} & \underline{47.3} \\
      3DRSBench                    & 37.2 & 34.2 & 45.8 & 52.1 & \underline{52.6} & \textbf{54.1} \\
      SITE-Bench-Image             & 50.7 & 47.6 & 51.4 & 52.1 & \underline{54.8} & \textbf{54.9} \\
      VSI-bench                    & 40.57 & 38.29 & 34.45 & 40.96 & \textbf{52.97} & \underline{43.53} \\
      \bottomrule
    \end{tabular}
    \end{adjustbox}
  \end{table}

  \paragraph{Text-to-image generation.}
  Table~\ref{tab:embodied-bench} reports the main comparison across the four
  benchmark families of \Secref{sec:exp-tasks}.
  \model{} first preserves strong
  general image generation: GenEval places \model{} ($82.4$) on par with the
  generation-specialist baseline Bagel ($82$) and well ahead of the other
  generation-capable model Cosmos-3-Nano ($71.68$). Crucially, this generative
  competence is preserved \emph{alongside}, not traded against, broad embodied
  understanding, on which most generation-capable models do not operate at all.

  \paragraph{Embodied and spatial understanding.}
  On \emph{embodied and spatial reasoning}, \model{} leads CV-Bench ($88.59$),
  EmbSpatial ($82.3$), and DA-2k ($83.4$), and is second-best on SAT ($74.33$).
  Its clearest advantage is in \emph{multi-view and 3D understanding}, where it
  attains the best score on 3DRSBench ($54.1$), MMSI-Bench ($35.8$), MindCube
  ($47.5$), and SITE-Bench-Image ($54.9$), and second-best on ViewSpatial ($47.3$)
  and VSI-bench ($43.53$). On \emph{robot affordance, trajectory, and placement} it
  gives the best trajectory prediction (Share-Robot-Trajectory, $51.13$) and leads
  RoboSpatial-Home configuration ($86.28$); fine-grained pointing and affordance
  (RefSpatial, Where2Place, Share-Robot-Affordance) remain the main gap, where the
  pointing-specialist RoboBrain2.5~\citep{tan2026robobrain25depthsight} is stronger. Overall, this combination---general
  prompt-faithful generation together with broad embodied and spatial competence in
  a single model---is the central capability \model{} is designed to deliver, rather
  than trading one axis of ability for the other.
\paragraph{RxBrain-Bench-EVQA.}
Table~\ref{tab:evqa-breakdown} reports the per-subtask results on RxBrain-Bench-EVQA.
RxBrain achieves an overall score of \textbf{72.7} across the
1,123 multiple-choice questions and 258 simulation-planning samples.
Its clearest strengths lie in outcome-aware reasoning and multi-round planning:
the model obtains the best results on Success Detection (\textbf{85.1}) and
Simulation Planning (\textbf{51.6}), and achieves the second-best result on
State Estimation (\underline{69.8}). In particular, the Simulation Planning
result indicates that Hy-Embodied-RxBrain can update its decisions over
multiple rounds according to intermediate simulated states and execution
outcomes, rather than producing only a one-shot task decomposition.
Overall performance remains below Qwen3.5-4B (78.1) and Cosmos3-Nano (76.7),
with the main gaps appearing in fine-grained spatial grounding,
trajectory and temporal reasoning, and error recovery. These results suggest
that Hy-Embodied-RxBrain is particularly effective at judging task outcomes
and maintaining goal-directed plans, while more precise perception and
physical-state reasoning remain important directions for improvement.

\begin{table}[htbp]
    \centering
    \caption{\textbf{RxBrain-Bench-EVQA: per-subtask accuracy breakdown (\%).}
    Ten multiple-choice subtasks covering 1{,}114 of 1{,}123 questions; three rare
    categories ($n \leq 4$) are omitted from the per-row listing due to insufficient
    sample size. Best result per row is \textbf{bold}; second-best is underlined.
    $^\dagger$\model{} Temporal Reasoning evaluated on 100 of 102 questions (two items
    filtered). $^\ddagger$Simulation Planning is a separate multi-round task-planning
    split ($n{=}258$) scored by a weighted MLLM judge (rescaled to \%). The
    \textbf{Overall} row is the sample-weighted average over all evaluated items
    (1{,}123 multiple-choice questions plus the 258 simulation-planning samples,
    $n{=}1{,}381$).}
    \label{tab:evqa-breakdown}
    \setlength{\tabcolsep}{4pt}
    \renewcommand{\arraystretch}{1.1}
    \begin{adjustbox}{max width=\linewidth}
    \small
    \begin{tabular}{lrcccccc}
      \toprule
      & & {Qwen3.5} & {Qwen3.5} & {Bagel} & {RoboBrain2.5} & {Cosmos-3-Nano} & {\textbf{\model{}}} \\
      Subtask & $n$ & 2B & 4B & 14B(A7B) & 4B & 16B(A8B) & 6B \\
      \midrule
      \rowcolor{annobg}
      \multicolumn{8}{l}{\textbf{Perception \& state}} \\
      State Understanding  & 171 & 75.4 & \textbf{95.3} & 81.3 & 87.7 & \underline{90.6} & 86.0 \\
      State Estimation     & 86  & 64.0 & \textbf{84.9} & 58.1 & 65.1 & 65.1 & \underline{69.8} \\
      \addlinespace[4pt]
      \rowcolor{annobg}
      \multicolumn{8}{l}{\textbf{Spatial \& grounding}} \\
      Spatial Reasoning    & 88  & 44.3 & \textbf{89.8} & 80.7 & 85.2 & \underline{86.4} & 73.9 \\
      Pointing             & 133 & 62.4 & \underline{72.2} & 69.9 & 64.7 & \textbf{79.7} & 69.9 \\
      \addlinespace[4pt]
      \rowcolor{annobg}
      \multicolumn{8}{l}{\textbf{Action, trajectory \& temporal}} \\
      Action Reasoning     & 96  & 49.0 & \textbf{89.6} & 75.0 & 69.8 & \underline{88.5} & 84.4 \\
      Trajectory Reasoning & 119 & 42.9 & \underline{78.2} & 66.4 & 72.3 & \textbf{84.0} & 69.7 \\
      Temporal Reasoning$^\dagger$ & 102 & 51.1 & \textbf{81.1} & 57.8 & 68.3 & \underline{79.3} & 69.1 \\
      \addlinespace[4pt]
      \rowcolor{annobg}
      \multicolumn{8}{l}{\textbf{Outcome \& planning}} \\
      Complex Planning     & 119 & 58.8 & \textbf{92.4} & 79.8 & 89.1 & \underline{91.6} & 88.2 \\
      Success Detection    & 101 & 47.5 & 76.2 & 69.3 & 60.4 & \underline{80.2} & \textbf{85.1} \\
      Error Recovery       & 99  & 65.7 & \underline{96.0} & \textbf{97.0} & \textbf{97.0} & \textbf{97.0} & 72.7 \\
      Simulation Planning$^\ddagger$ & 258 & 26.8 & \underline{45.3} & 27.4 & 41.2 & 42.0 & \textbf{51.6} \\
      \midrule
      \textbf{Overall}     & 1{,}381 & 51.6 & \textbf{78.1} & 65.1 & 69.7 & \underline{76.7} & 72.7 \\
      \bottomrule
    \end{tabular}
    \end{adjustbox}
\end{table}

\paragraph{World state prediction and joint subgoal planning.}
Tables~\ref{tab:planning-bench} and~\ref{tab:videogen-bench} summarize
\model{} on the two \emph{generative} tracks of \model{}-Bench, including both
the per-subset/per-criterion breakdowns and comparisons with the baselines
described in \Secref{sec:exp-tasks}. We report the weighted MLLM-judge scores
$S_{\mathrm{plan}}$ and $S_{\mathrm{gen}}$ (\S\ref{sec:metrics}) as the primary
metrics.

\begin{table}[h]
    \centering
    \caption{\textbf{RxBrain-Bench-WorldPred: method comparison.}
    Held-out, short-horizon ($4$ frames). $S_{\mathrm{gen}}$ is the weighted MLLM-judge score;
    the five columns after it are its judge criteria. Baselines: \emph{Wan2.2-TI2V-5B} = a
    general-purpose I2V diffusion model; \emph{Cosmos3-Nano} = an embodied omni world model (I2V).
    Both baselines and \model{} are scored on the same held-out balanced subset with identical
    $4$-frame alignment; $n$ differs due to judge content-filtering.}
    \label{tab:videogen-bench}
    \setlength{\tabcolsep}{5pt}\renewcommand{\arraystretch}{1.1}
    \begin{adjustbox}{max width=\linewidth}\small
    \begin{tabular}{l c ccccc c}
      \toprule
      Method & $S_{\mathrm{gen}}$ & ObsCont & ActCorr & GoalComp & TempPlaus & GTMatch & $n$ \\
      \midrule
      \textbf{\model{}}   & \textbf{0.62} & 0.67 & \textbf{0.65} & \textbf{0.62} & \textbf{0.53} & \textbf{0.59} & 420 \\
      \midrule
      Cosmos3-Nano                  & 0.591 & \textbf{0.76} & 0.62 & 0.58 & 0.48 & 0.53 & 420 \\
      Wan2.2-TI2V-5B                & 0.429 & 0.75 & 0.42 & 0.37 & 0.32 & 0.34 & 420 \\
      \bottomrule
    \end{tabular}
    \end{adjustbox}
  \end{table}
On RxBrain-Bench-WorldPred, \model{} achieves
$S_{\mathrm{gen}}=0.62$. Observation continuity ($0.67$) and action
correctness ($0.65$) receive the highest scores, while temporal and physical
plausibility ($0.53$) remains the weakest aspect, suggesting that motion
consistency is the main limitation of the current model. Performance ranges
from $0.73$ on \emph{umi} to $0.35$ on \emph{arctic}, with two-frame
conditioning providing the best overall results. Compared with the baselines
(Table~\ref{tab:videogen-bench}), \model{} substantially outperforms the
general-purpose video model Wan2.2-TI2V-5B ($0.429$) and edges out
Cosmos3-Nano ($0.591$), despite the latter being a specialized
embodied world model pretrained on substantially larger-scale video data.
Overall, these results indicate that a single unified model can match dedicated
world models on future prediction while significantly outperforming modular
pipelines on interleaved multimodal planning. Across both tasks, language
reasoning consistently exceeds visual generation quality, with goal-image
correctness and temporal plausibility remaining the two primary bottlenecks.
\Cref{fig:qual-videogen-umi} illustrates this behavior on two held-out actions:
conditioned on the initial observation and the action instruction, \model{}
predicts four future frames that preserve scene layout and reproduce the
intended manipulation, closely tracking the ground-truth rollout, while the
general-purpose Wan2.2-TI2V baseline introduces spurious objects and
appearance changes and Cosmos3-Nano advances the motion more slowly.
\begin{figure}[h]
  \centering
  \setlength{\ilcw}{104pt}

  {\footnotesize\bfseries \emph{UMI} --- handheld gripper: approach the white cup on the table and grasp it.\par}
  \vspace{2pt}

  \noindent\makebox[\textwidth]{%
  \footnotesize
  \setlength{\ilcw}{98pt}
  \setlength{\tabcolsep}{1.5pt}\renewcommand{\arraystretch}{1.0}
  \begin{tabular}{@{}>{\bfseries\scriptsize}p{40pt}@{\hspace{2pt}}KKKK@{}}
   & \multicolumn{1}{@{}c@{}}{\scriptsize\bfseries Frame 1}
   & \multicolumn{1}{c}{\scriptsize\bfseries Frame 2}
   & \multicolumn{1}{c}{\scriptsize\bfseries Frame 3}
   & \multicolumn{1}{c@{}}{\scriptsize\bfseries Frame 4}\\[1pt]

  GT 
  & \mfcell{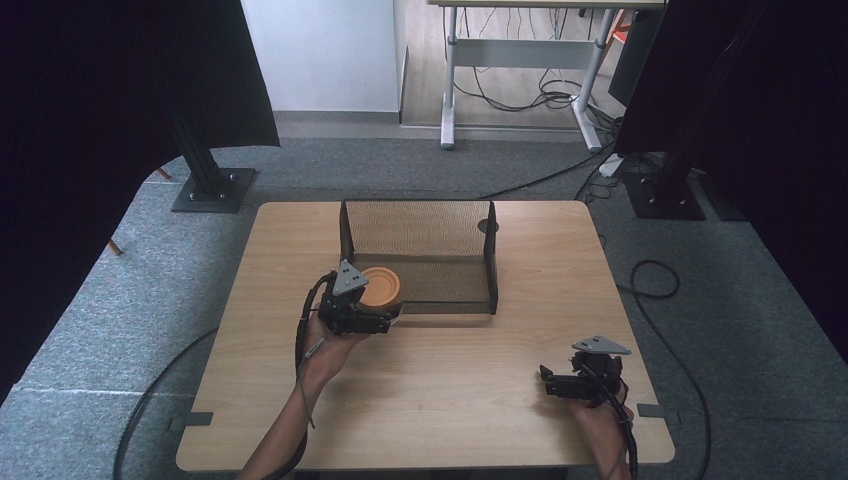}
  & \mfcell{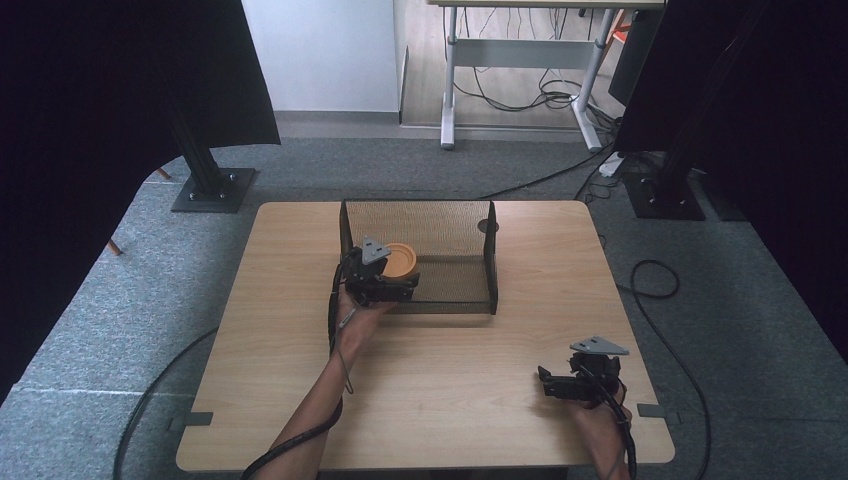}
  & \mfcell{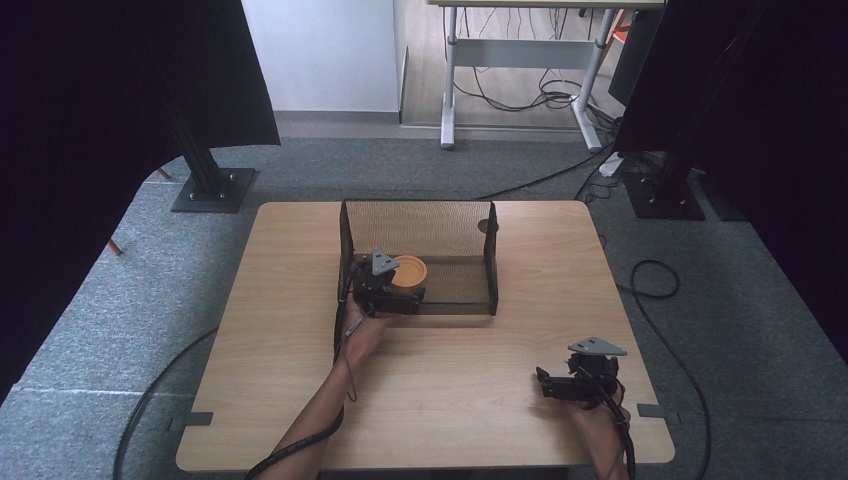}
  & \mfcell{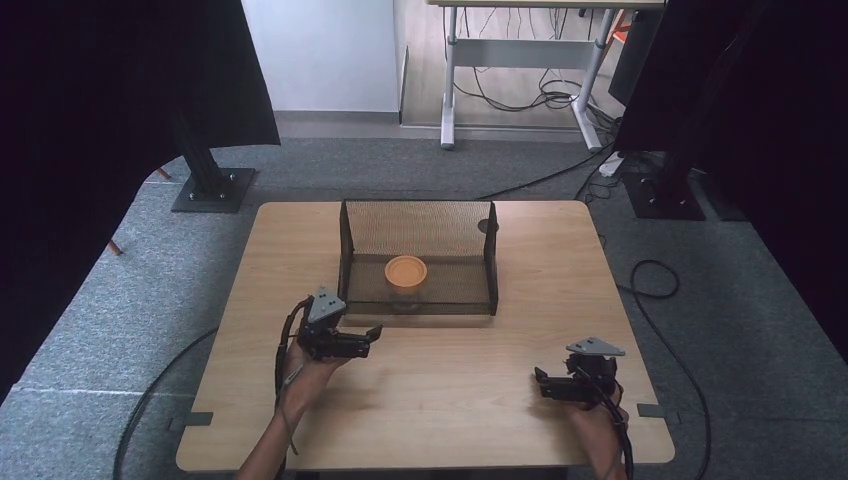}\\[1pt]

  RxBrain (ours)
  & \mfcell{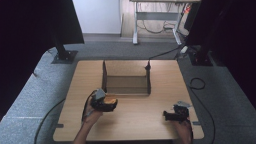}
  & \mfcell{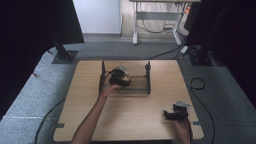}
  & \mfcell{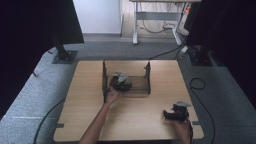}
  & \mfcell{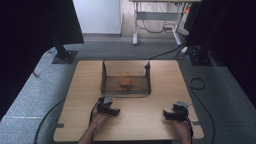}\\[1pt]

  Wan2.2-TI2V
  & \mfcell{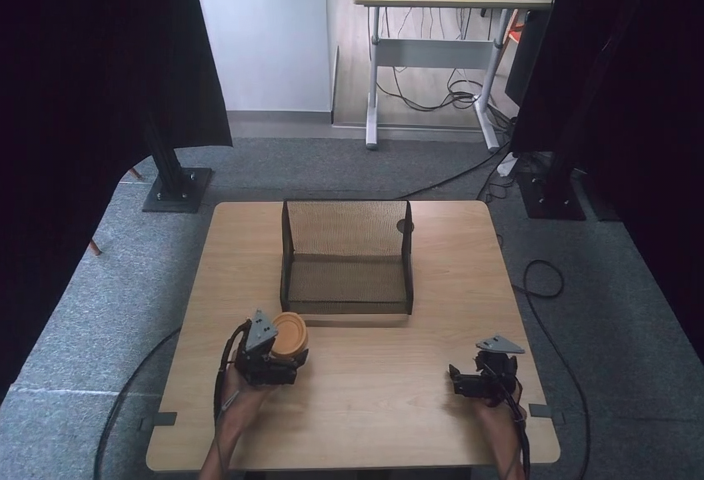}
  & \mfcell{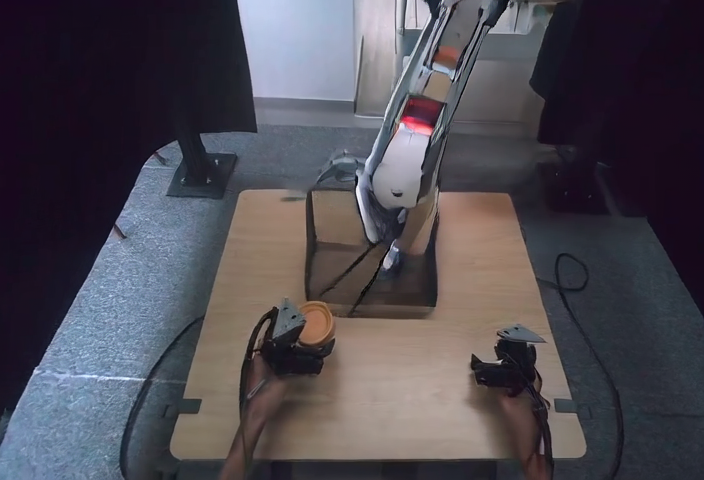}
  & \mfcell{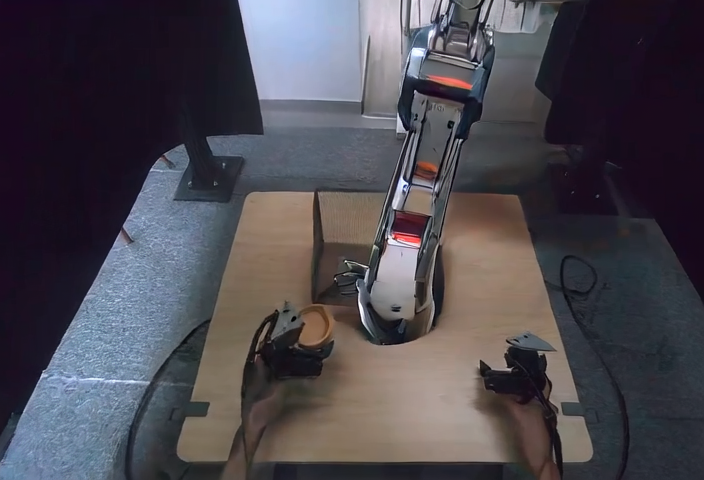}
  & \mfcell{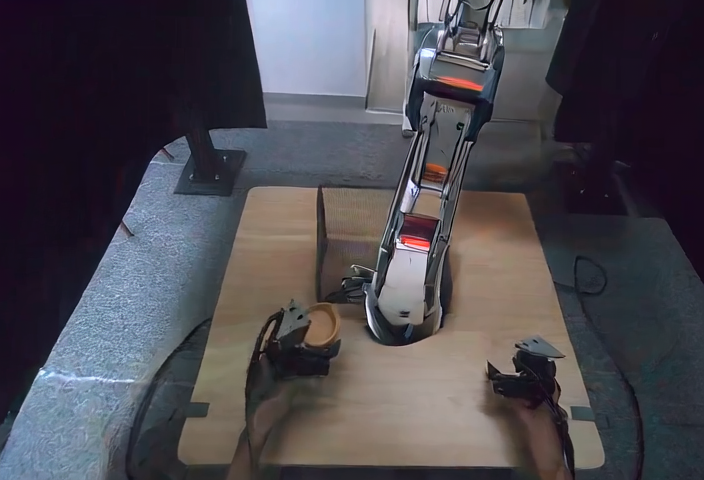}\\[1pt]

  Cosmos3
  & \mfcell{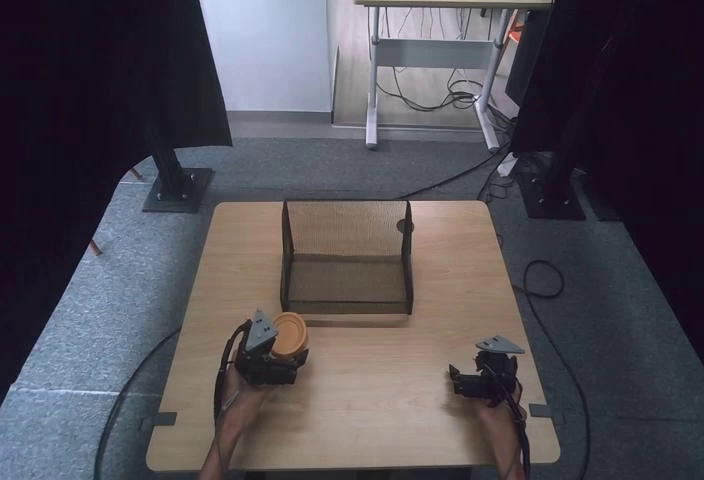}
  & \mfcell{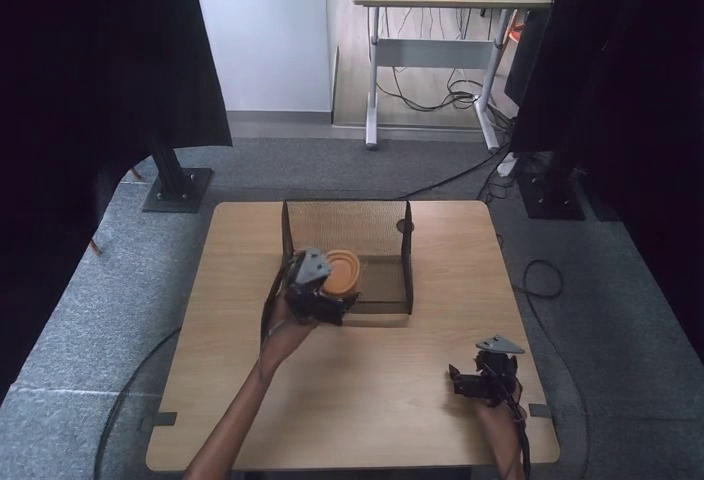}
  & \mfcell{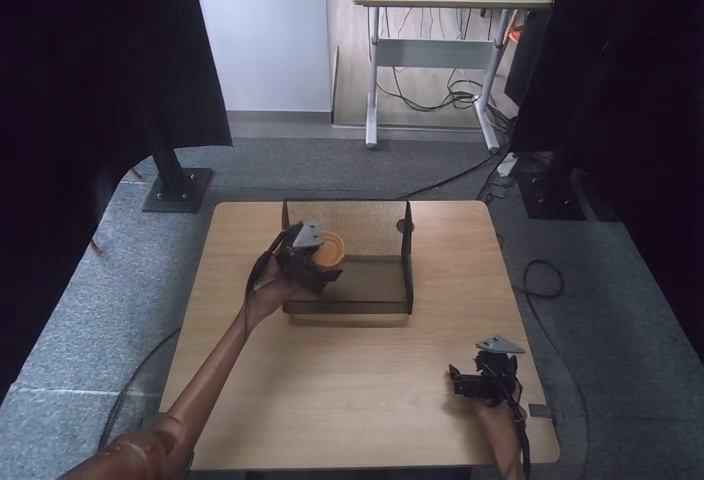}
  & \mfcell{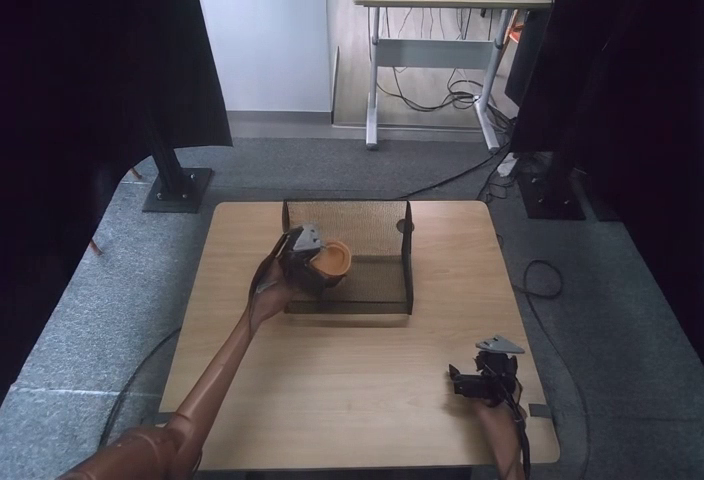}\\
  \end{tabular}}

  \caption{\textbf{RxBrain-Bench-WorldPred (\emph{UMI}).}
  Four-frame future prediction for handheld manipulation. Rows compare the
  ground truth (\emph{GT}), \model{} (\emph{HY-Unified}), the general-purpose
  I2V model \emph{Wan2.2-TI2V}, and the embodied world model \emph{Cosmos3}.
  \model{} preserves the scene layout and reproduces the intended grasping
  behavior.}

  \label{fig:qual-videogen-umi}
\end{figure}

\begin{table}[htbp]
    \centering
    \caption{\textbf{RxBrain-Bench-JointPlan: method comparison.}
    Held-out$S_{\mathrm{plan}}$ is the weighted MLLM-judge score; the five
    columns after it are its judge criteria, and DINO is the sixth rubric term (Image Similarity,
    folded into $S_{\mathrm{plan}}$). Baselines: \emph{Qwen-Agent} = Qwen3-VL-2B (reasoner) +
    Qwen-Image-Edit (generator); \emph{Cosmos3-Nano} = a single omni model orchestrated as an
    agent (chat reasoner + I2V goal-image generator); \emph{BAGEL-7B-MoT} = a unified interleaved
    image--text model~\citep{bagel} run as a planning agent. $n$ differs slightly across methods due to
    judge content-filtering; all share the same held-out balanced subset (seed 0).}
    \label{tab:planning-bench}
    \setlength{\tabcolsep}{5pt}\renewcommand{\arraystretch}{1.1}
    \begin{adjustbox}{max width=\linewidth}\small
    \begin{tabular}{l c ccccc c c}
      \toprule
      Method & $S_{\mathrm{plan}}$ & Obs & Plan & GoalImg & Consist & Chain & DINO & $n$ \\
      \midrule
      \textbf{\model{}}       & \textbf{0.68} & \textbf{0.83} & \textbf{0.78} & \textbf{0.52} & \textbf{0.67} & \textbf{0.69} & 0.72 & 540 \\
      \midrule
      Cosmos3-Nano (agent)              & 0.521 & 0.67 & 0.62 & 0.26 & 0.47 & 0.63 & \textbf{0.75} & 540 \\
      BAGEL-7B-MoT                      & 0.503 & 0.58 & 0.55 & 0.25 & 0.60 & 0.56 & 0.67 & 540 \\
      Qwen-Agent                        & 0.431 & 0.57 & 0.53 & 0.20 & 0.44 & 0.53 & 0.52 & 540 \\
      \bottomrule
    \end{tabular}
    \end{adjustbox}
  \end{table}
  
On RxBrain-Bench-JointPlan, \model{} achieves
$S_{\mathrm{plan}}=0.68$. The per-criterion breakdown shows that language
reasoning is the model's strongest capability, with observation understanding
($0.83$) and subtask planning ($0.78$) substantially outperforming goal-image
correctness ($0.52$), indicating that visual imagination remains the primary
bottleneck. Performance is consistently strong across most subsets
($0.61$--$0.75$), while the fine-grained tabletop benchmark \emph{arctic}
remains the most challenging ($0.48$). Under free-running rollouts, performance
gradually decreases with planning horizon, from $0.69$ at two steps to $0.55$
at eight steps, reflecting accumulated autoregressive errors. Compared with the
baselines (Table~\ref{tab:planning-bench}), \model{} substantially outperforms
the Cosmos3-Nano agent ($0.521$), the unified BAGEL-7B-MoT ($0.503$), and
Qwen-Agent ($0.431$), leading across
every judge criterion of $S_{\mathrm{plan}}$, with the largest gains on subtask
planning and chain completion.

\begin{figure}[H]
  \centering
  \setlength{\ilcw}{82pt}
  \noindent\makebox[\textwidth]{%
  \footnotesize
  \setlength{\tabcolsep}{1.5pt}\renewcommand{\arraystretch}{1.0}
  \begin{tabular}{@{}>{\bfseries\scriptsize}p{34pt}@{\hspace{2pt}}KKKKK@{}}
   & \multicolumn{1}{@{}c@{}}{\scriptsize\bfseries Step 1} & \multicolumn{1}{c}{\scriptsize\bfseries Step 2} & \multicolumn{1}{c}{\scriptsize\bfseries Step 3} & \multicolumn{1}{c}{\scriptsize\bfseries Step 4} & \multicolumn{1}{c@{}}{\scriptsize\bfseries Step 5}\\[1pt]
  GT goal &
   \ilcell{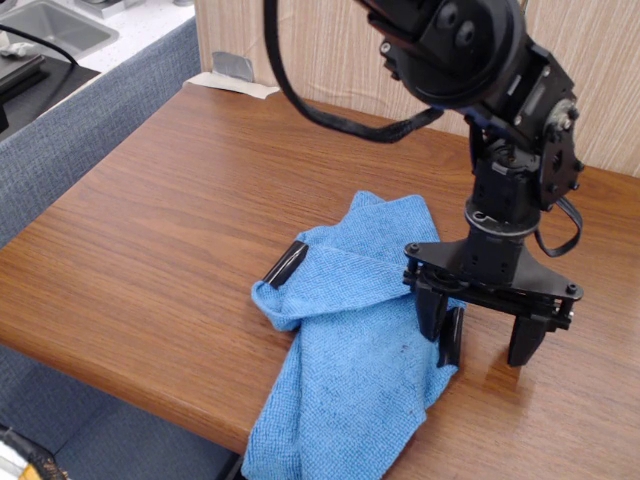}{The robotic gripper moves down and forward from above the table toward the right edge of the blue cloth, aligning its open jaws near a corner.} &
   \ilcell{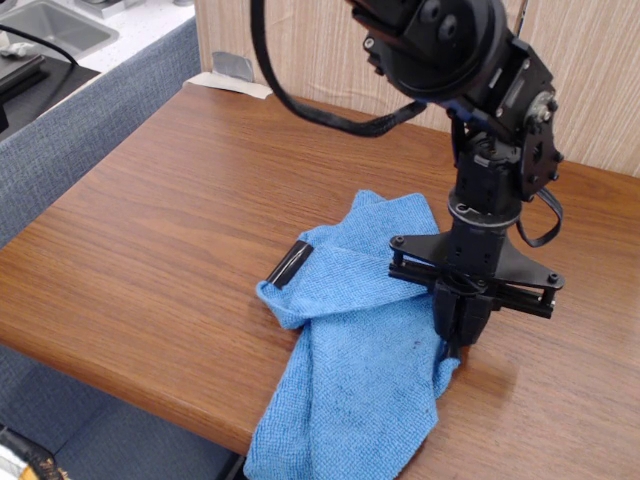}{The gripper closes its two jaws around the right-side edge of the blue cloth, pinching the fabric between the fingers.} &
   \ilcell{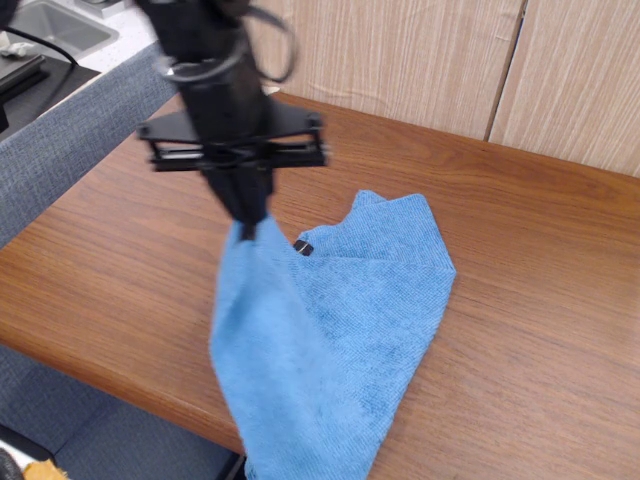}{With the fabric pinched, the gripper raises the held edge upward so part of the blue cloth lifts off the table and forms a hanging fold.} &
   \ilcell{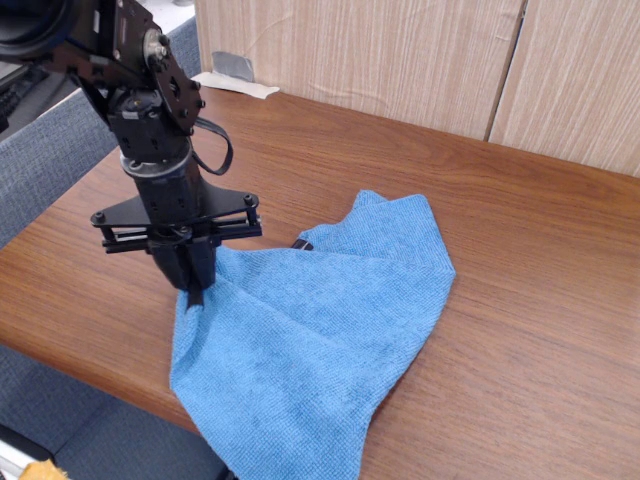}{The gripper carries the lifted section leftward while holding it, pulling the cloth across the tabletop and changing its spread.} &
   \ilcell{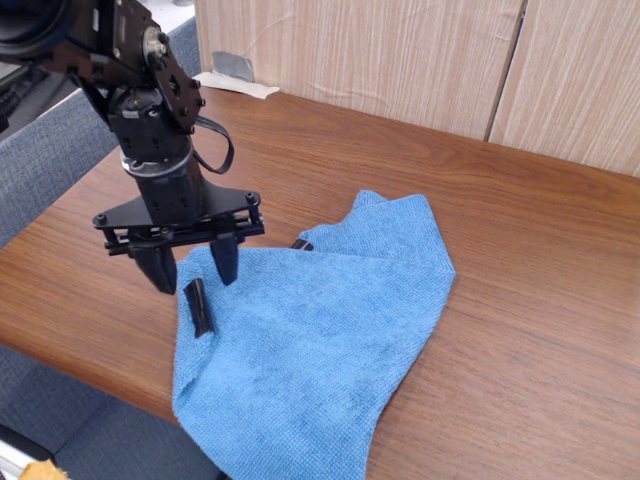}{The gripper opens its jaws and lets the lifted edge fall back onto the table, leaving the blue cloth in a new draped position.}\\[2pt]
  RxBrain (ours) &
   \ilcell{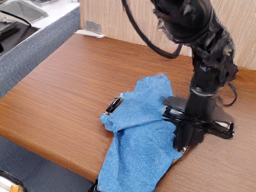}{The gripper hovers above the blue cloth and moves from the upper center area toward the right edge of the cloth on the table, aligning its jaws.} &
   \ilcell{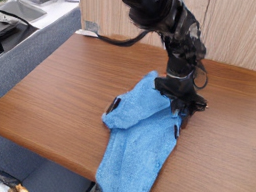}{The gripper lowers onto the right edge of the blue cloth and closes its jaws to pinch the fabric near the corner, securing the cloth.} &
   \ilcell{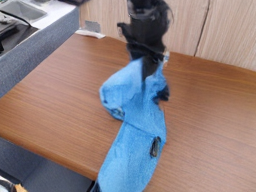}{With the cloth pinched, the gripper raises upward, pulling the right side of the blue cloth off the table so the fabric stands up and folds.} &
   \ilcell{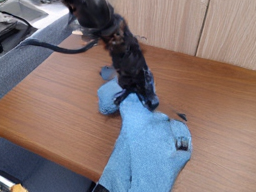}{The gripper carries the lifted cloth leftward and slightly inward over the table, shifting its position and changing its folded shape.} &
   \ilcell{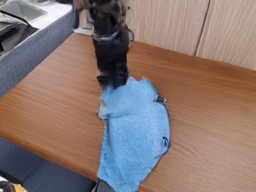}{The gripper opens its jaws and lets the cloth fall back onto the table, leaving the blue fabric in a new spread-out position.}\\[2pt]
  Qwen-agent &
   \ilcell{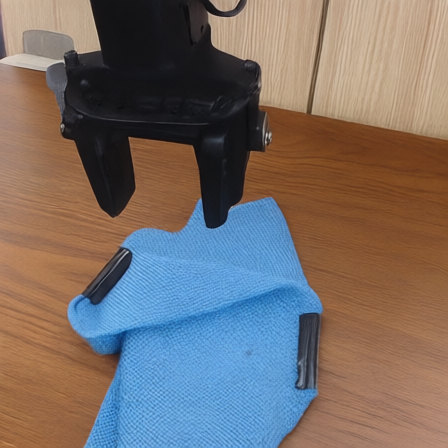}{Move to the right edge of the blue cloth.} &
   \ilcell{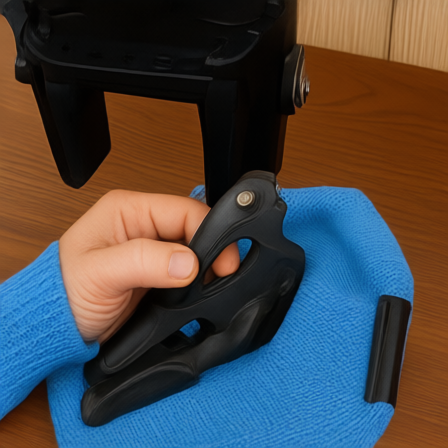}{Grasp the blue cloth with the gripper.} &
   \ilcell{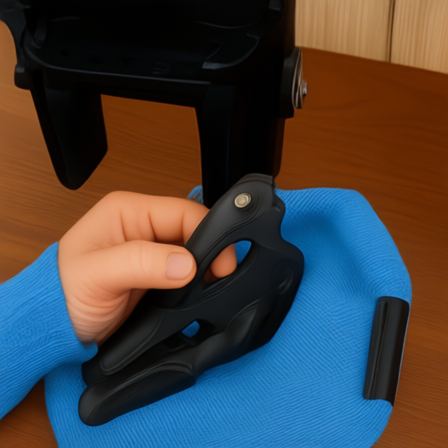}{Pull the blue cloth to the left.} &
   \ilcell{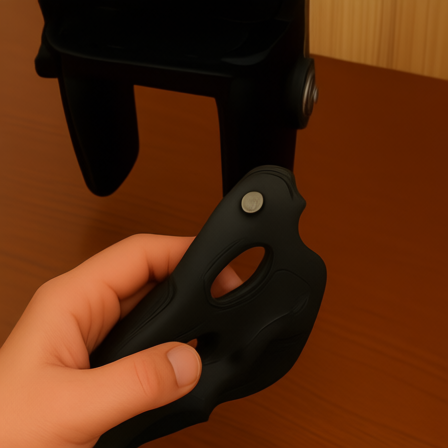}{Release the blue cloth.} &
   \ilcell{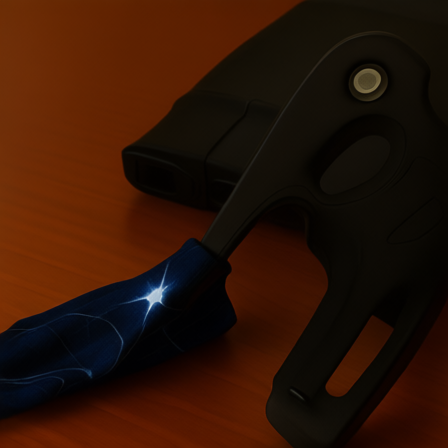}{Move the blue cloth to the left.}\\[2pt]
  Cosmos3 &
   \ilcell{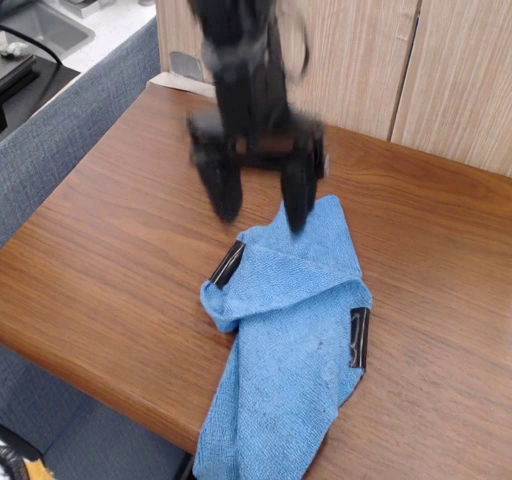}{Move the robot gripper to the right edge of the blue cloth.} &
   \ilcell{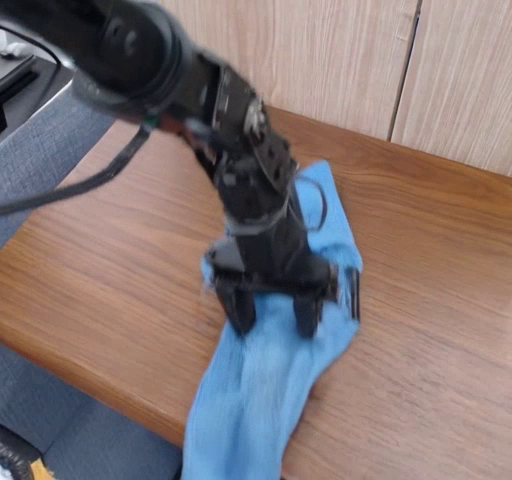}{Grasp the blue cloth at its right edge.} &
   \ilcell{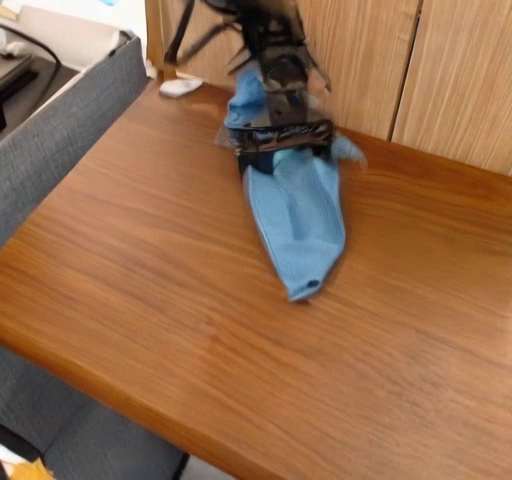}{Pull the blue cloth leftward while maintaining a firm grip.} &
   \ilcell{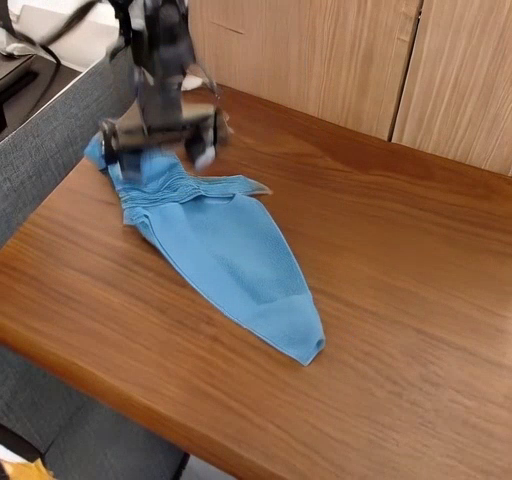}{Release the blue cloth after moving it to the desired position.} &
   \ilcell{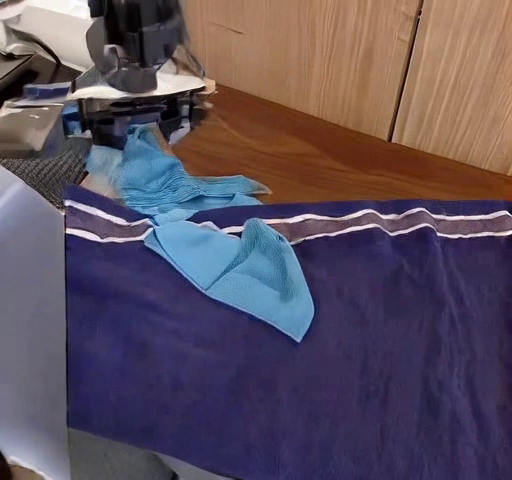}{Retract the robot gripper away from the blue cloth.}\\
  \end{tabular}}
  \caption{\textbf{RxBrain-Bench-JointPlan (\emph{BridgeV2}),
  task ``move to the right edge of the blue cloth, grasp it, pull it leftward, and
  release it.''} rows are models, columns are
  planning steps, and each cell pairs the imagined goal frame with the model's
  generated sub-step text. \model{} reproduces the grasp--pull--release sequence and
  the cloth's changing shape in step with the ground-truth plan (\emph{GT goal}).}
  \label{fig:qual-planning-bridge}
\end{figure}

Figures~\ref{fig:qual-planning-bridge}
show representative joint subgoal planning
rollouts on a held-out task: for the same initial observation and goal,
\model{} produces sub-step texts and predicted goal frames that stay aligned
with the ground-truth plan, whereas the modular Qwen-Agent tends to repeat a
generic instruction and drift visually, and the Cosmos3-Nano agent
hallucinates object appearances and scene changes over successive steps.

\FloatBarrier

%% file: content/action.tex
\section{Extending RxBrain to Action Generation}
Building on RxBrain, we extend the unified embodied model with an action-generation capability for robotic manipulation. Our goal is to investigate whether the world knowledge and state prediction priors acquired through multimodal pretraining can provide a strong foundation for action prediction without requiring large-scale action pretraining. To this end, this section presents the architecture of the proposed action model and evaluates its effectiveness on real-world robotic systems.
\label{sec:b2a}
\subsection{Action Model Architecture}
For action, we extend RxBrain with a modality-specialized branch containing dedicated attention projections $\{Q,K,V,O\}_a$, FFN, and layer norms. The branch is initialized by copying the corresponding parameters from the vision understanding branch. Meanwhile, action tokens remain in the shared global attention operation with observation, instruction, and state tokens, allowing action-specific parameterization while preserving cross-modal conditioning.
\begin{figure}[h]
    \centering
    \includegraphics[width=\linewidth]{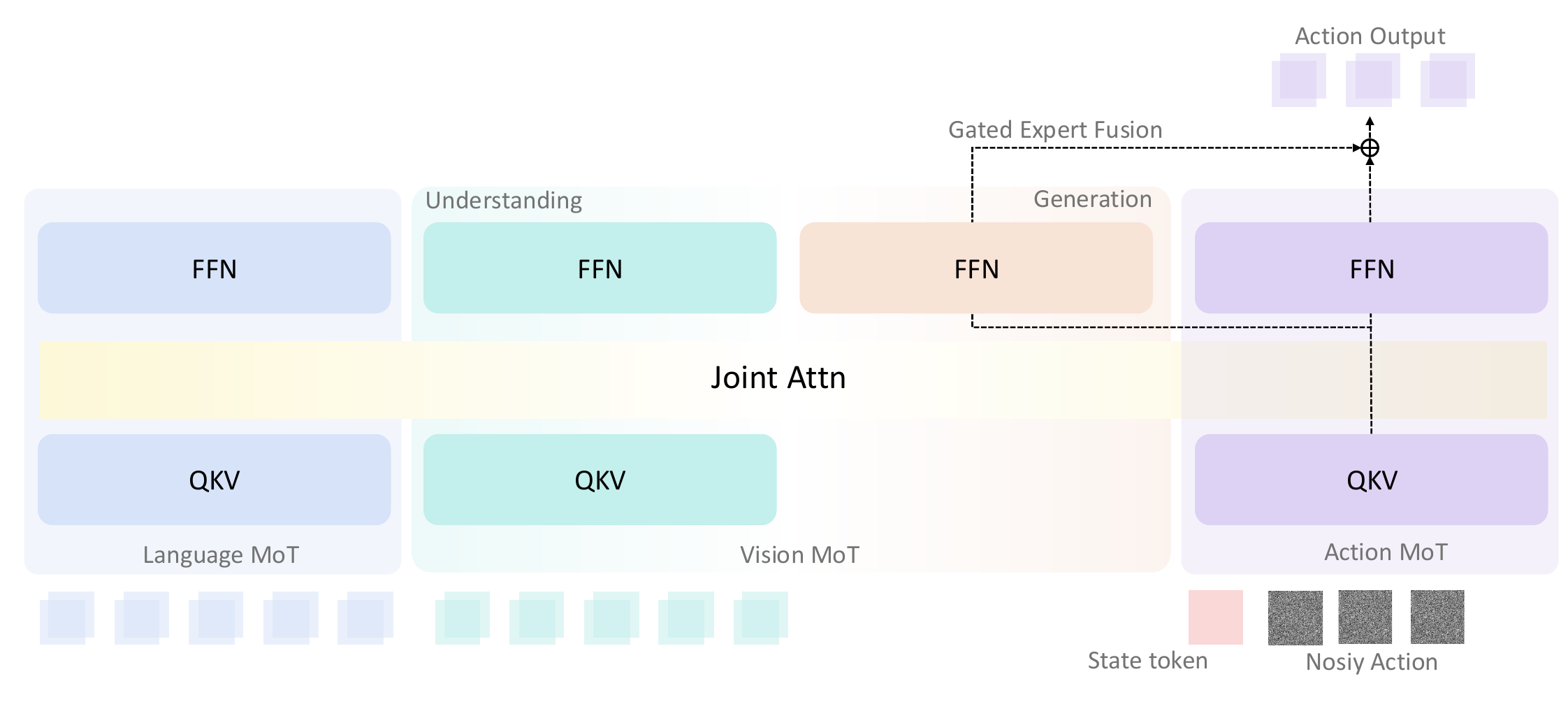}
    \caption{
        \textbf{Action model architecture.}
        A modality-specialized Action MoT branch uses dedicated attention and
        feed-forward parameters while sharing global self-attention with
        language, vision, and state tokens. A zero-initialized channel-wise
        gate transfers features from the pretrained generation
        expert to the action expert, whose outputs are decoded into action
        chunks.
    }
    \label{fig:action-model-architecture}
\end{figure}
\paragraph{Gated Expert Fusion.}
In RxBrain, the generation branch is pretrained for general image generation, world-state prediction
and joint subgoal planning, making its feed-forward expert
$\mathrm{FFN}_g$ a natural repository of predictive and planning
representations. Since action generation also depends on these capabilities,
the action branch should efficiently reuse the pretrained generation expert
rather than relearn the same knowledge from scratch.

Let $x_a$ denote the hidden states of the action tokens after joint
self-attention. We apply the action and generation experts to the same action-token representation, while retaining the branch-specific normalization of each expert. The fused feed-forward update is defined as

\begin{equation}
\mathrm{FFN}^{\mathrm{fused}}_a(x_a)
=
\mathrm{FFN}_a\!\bigl(\mathrm{LN}_a(x_a)\bigr)
+
g \odot
\mathrm{FFN}_g\!\bigl(\mathrm{LN}_g(x_a)\bigr),
\end{equation}

where $\mathrm{LN}_a$ and $\mathrm{LN}_g$ are the LayerNorm modules associated
with the action and generation branches, respectively, and $\odot$ denotes
element-wise multiplication. Thus, both experts process the same
post-attention action representation, but through their own learned
normalization parameters. This preserves the feature scaling expected by
each expert when the pretrained generation expert is reused for action
tokens.

The fused feed-forward update is incorporated through the standard residual
connection,

\begin{equation}
x_a^{\mathrm{out}}
=
x_a + \mathrm{FFN}^{\mathrm{fused}}_a(x_a).
\end{equation}

The per-channel gate $g$ is initialized to zero, so the fused module initially
reduces exactly to the standalone action expert. This provides a
function-preserving starting point without introducing an immediate
contribution from the generation branch. During action training, the gate
learns to selectively inject predictive and planning features from
$\mathrm{FFN}_g$, while preserving the modality-specific capacity of
$\mathrm{FFN}_a$. The learned magnitude of $g$ also provides a direct measure
of how strongly each channel relies on the pretrained generation expert.
\paragraph{Action representation and flow-matching head}
The proprioceptive state (bimanual end-effector pose $[\text{pos}_3, \text{quat}_4, \text{grip}_1]\times 2$, $16$-D) is mapped by a lightweight state encoder into a single proprioception token, and the next $H$ steps form an action chunk in which each step is one action token. Actions are generated with flow-matching formulation, applied in the normalized action space: given a clean action chunk $a$ and noise $\epsilon\sim\mathcal{N}(0,I)$, we form the interpolant $a_s=(1-s)\,a+s\,\epsilon$ and train the action head to regress the velocity field,

$$\mathcal{L}_{\text{action}} = \mathbb{E}_{s,a,\epsilon}\big[\,\lVert v_\phi(a_s, s, c) - (\epsilon - a)\rVert_2^2\,\big].$$

. Following \citep{pi0} and \citep{pi05}, the action chunk draws a single per-chunk flow time $s\sim\mathrm{Beta}(1.5,1)$ biased toward high noise,. At inference, the action chunk is denoised from pure noise by a few-step Euler ODE along the predicted velocity field and de-normalized into end-effector pose commands.

\subsection{Real-world Deployment}

\begin{figure}[h]
    \centering
    \includegraphics[
        width=\linewidth,
        trim={0 0.5cm 0 0.5cm},
        clip
    ]{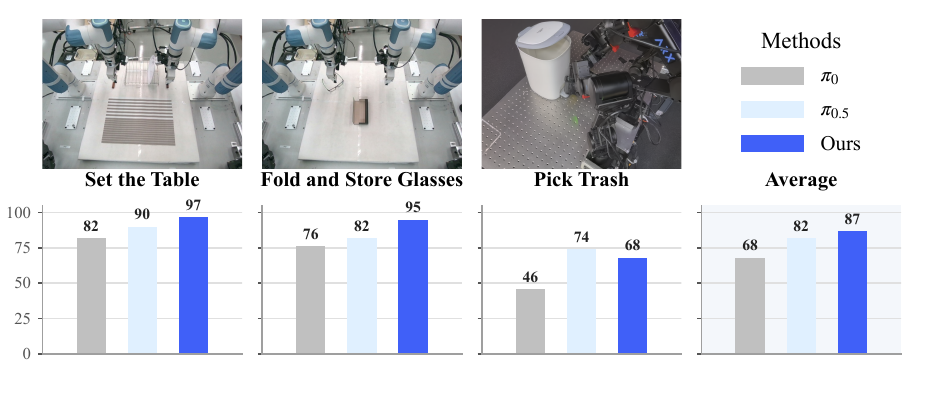}
    \caption{
        \textbf{Real-world robot evaluation on three manipulation tasks.}
        We compare our method with $\pi_0$ and $\pi_{0.5}$ on three manipulation tasks.
        Our method achieves the highest success rate on average.
    }
    \label{fig:real_robot_eval}
\end{figure}

\paragraph{Experimental Setup and Tasks.}
We further evaluate our method on two real-world robot embodiments:
the DOBOT X-Trainer and an ARX dual-arm platform.
The first two tasks are conducted on the DOBOT X-Trainer, while the
third task is evaluated on the ARX platform.
As illustrated in Fig.~\ref{fig:real_robot_eval}, we consider three
multi-stage household manipulation tasks.

In \textit{Set the Table}, the robot must retrieve a plate, a fork,
and a knife from a storage rack and arrange them into a valid place
setting on the table.
In \textit{Fold and Store Glasses}, the robot is required to fold
the temples of a pair of eyeglasses and place the folded eyeglasses
into a glasses case.
In \textit{Pick Trash}, the robot must first open the lid of a
trash bin and then pick up and deposit a piece of waste into the bin.

The baselines and our model are all trained on the same real-world robot data. In addition, we use the joint planning data construction pipeline illustrated in Section~\ref{sec:data} to annotate joint subgoal planning labels on real-world robot data, and these annotated data are incorporated into the training of our model.

\paragraph{Results.}
We evaluate each policy with 100 real-robot trials on each task and compare our method against two strong vision-language-action baselines, $\pi_0$ and $\pi_{0.5}$, using task success rate as the evaluation metric. As shown in Fig.~\ref{fig:real_robot_eval}, our method consistently outperforms both baselines across the three tasks. Specifically, our method achieves success rates of 97\% on \textit{Set the Table}, 95\% on \textit{Fold and Store Glasses} and 68\% on \textit{Pick Trash}, resulting in an average success rate of  87\%. In comparison, $\pi_0$ and $\pi_{0.5}$ achieve average success rates of  68\% and  82\%, respectively. These results demonstrate that our action generation design can effectively perform complex manipulation behaviors, even without additional action pretraining. This suggests that the proposed approach can learn a reliable action generation policy from the pre-trained model itself, enabling robust manipulation in real-world environments.

%% file: content/related.tex
\section{Related Work}
\label{sec:related}

\subsection{Vision-Language Models for Embodied Understanding}

Vision-language models (VLMs) bridge visual perception and language. Early contrastive methods such as CLIP~\citep{radford2021clip} align images and text in a shared embedding space, providing transferable representations for downstream recognition. Subsequent generative VLMs connect a pretrained vision encoder to a large language model: Flamingo~\citep{alayrac2022flamingo} introduces gated cross-attention for few-shot multimodal learning, BLIP-2~\citep{li2023blip2} bridges frozen encoders and LLMs with a lightweight querying transformer, and LLaVA~\citep{liu2023llava} shows that simple visual instruction tuning can enable strong open-ended visual dialogue. More recent model families such as Qwen-VL~\citep{bai2023qwenvl,wang2024qwen2vl,bai2025qwen3vl} and InternVL~\citep{chen2024internvl,zhu2025internvl3} scale data, resolution, and encoder capacity, achieving robust captioning, visual question answering, OCR, long-video understanding, and spatial grounding. Natively multimodal models such as GPT-4o~\citep{openai2024gpt4o} further extend this trend by coupling visual and linguistic processing more tightly within a single model. Together, these models provide increasingly reliable scene understanding, which has begun to empower embodied agents that must ground perception in the physical world.

Building on this understanding capability, a growing line of work develops VLMs tailored to embodied scenarios. PaLM-E~\citep{palme} incorporates continuous sensor inputs into a language model for embodied reasoning and planning, while RT-2~\citep{rt2} casts robot actions as language tokens and transfers web-scale vision-language knowledge to robotic control. Recent embodied VLMs further improve spatial grounding and task-level planning. RoboBrain~\citep{ji2025robobrain, robobrain2025v2} targets robotic manipulation by modeling planning, affordance perception, and trajectory prediction, and RynnBrain~\citep{dang2025rynnec, dang2026rynnbrain} emphasizes egocentric cognition, spatiotemporal grounding, and physical-space reasoning. MolmoAct~\citep{lee2025molmoact, fang2026molmoact2} introduces action reasoning models that transform observations and instructions into depth-aware perception tokens, editable spatial trajectory traces, and low-level robot actions, making the planning process more interpretable and steerable. Cosmos-Reason1~\citep{azzolini2025cosmosreason1} further focus on physical common sense and embodied reasoning, producing next-step decisions through multimodal reasoning. These models significantly strengthen embodied perception, spatial reasoning, and action planning. However, they mainly output language-level decisions, symbolic plans, trajectories, or actions, rather than explicitly generating future visual states.

\subsection{Generative World Models}

In parallel with VLMs, world models aim to predict how the environment evolves over time. Recent video generation models show that large-scale generative training can capture rich visual dynamics and synthesize temporally coherent future frames~\citep{wan2025,kondratyuk2023videopoet,blattmann2023stable, genie, genie2, lin2026genie3}. Beyond pixel-level generation, latent predictive models such as V-JEPA~\citep{bardes2024vjepa} and V-JEPA 2~\citep{assran2025vjepa2} learn to predict future or masked spatiotemporal representations, showing that video-based self-supervision can support understanding, prediction, and planning. These works suggest that video prediction and latent prediction provide useful forms of visual imagination.

Recent embodied world action models (WAMs) bring this predictive ability closer to robot control. Video Prediction Policy (VPP)~\citep{hu2024vpp} uses representations from pretrained video diffusion models to guide robot policy learning through predicted future visual features. Ctrl-World~\citep{guo2025ctrl} develops a controllable multi-view generative world model for policy-in-the-loop rollout, enabling fine-grained action-conditioned prediction and long-horizon robot policy evaluation in imagination. LingBot-VA~\citep{lingbot-va2026} jointly learns frame prediction and policy execution in a causal autoregressive diffusion framework, interleaving visual and action tokens for closed-loop robot control. DreamZero~\citep{ye2026dreamzero} further formulates world action models as zero-shot policies by jointly predicting future world states and actions with a pretrained video diffusion backbone. These models explicitly connect future prediction with action generation, making them closer to embodied imagination than standard VLMs. However, they mainly emphasize generative prediction and policy learning, while explicit scene understanding remains limited. They do not naturally support language-based inspection of imagined scenes or semantic verification of whether a predicted future state satisfies a task goal.

\subsection{Unified Multimodal Models for Understanding-Generation}

A more recent line of work seeks to unify visual understanding and generation within a single model, rather than treating them as separate systems. One family couples an autoregressive language backbone with a diffusion objective for images: Transfusion~\citep{zhou2025transfusion} trains a single transformer with a language-modeling loss on text tokens and a diffusion loss on image patches, so that one model both understands and generates. Chameleon~\citep{chameleon} instead tokenizes images and text into a shared discrete vocabulary and models them with a single early-fusion autoregressive objective. Building on such mixed-modal training, BAGEL~\citep{bagel} scales an interleaved image-text model and reports emerging capabilities such as image editing and free-form visual manipulation, while unified frameworks like Janus~\citep{janus} decouple the visual encoding pathways for understanding and generation to reduce interference between the two objectives. These models demonstrate that understanding and generation can share parameters and benefit from joint training.

More recently, this unification has been pushed toward embodied and physical-world settings. Cosmos3~\citep{agarwal2026cosmos} combines visual understanding, visual generation, world simulation, and action modeling within a unified Mixture-of-Transformers (MoT) architecture. This provides a strong foundation for embodied agents by enabling models to both interpret the current scene and imagine possible future states. 

Moving toward next-generation embodied cognition foundation models, we argue that models should not only possess strong embodied language reasoning and world state prediction capabilities, but also provide language-visual subgoal guidance, where high-level goals can be translated into intermediate visual objectives for planning and verification. Cosmos3 adopts specialized reasoning and generation pathways with different task modes to achieve multimodal world modeling. Inspired by this direction, we explore a more unified architecture that enables embodied reasoning, world state prediction, and joint language-visual subgoal planning within an interleaved reasoning process. This motivates the design of \model.

\subsection{Action Generation with Unified Models}
Embodied action generation has increasingly been studied using large pretrained models, including VLM-based~\citep{yuan2026qwen, wu2026pragmaticvlafoundationmodel, chen2025internvla, pi05, openvla} and World Model based~\citep{gao2026dreamdojo,ye2026world, lingbot-va2026, ye2026gigaworldpolicyefficientactioncenteredworldaction, kim2026cosmos}policies.
Recent works explore unified multimodal models for embodied action generation, where actions are modeled jointly with visual and language modalities. UP-VLA~\citep{zhang2025upvlaunifiedunderstandingprediction} incorporates visual understanding and future prediction objectives into VLA training, while MM-ACT~\citep{liang2025mm} introduces a unified discrete diffusion framework that jointly models images, language, and actions. Beyond multimodal action generation, RynnVLA-002~\citep{cen2025rynnvla} further integrate action modeling with world dynamics by jointly learning future states and executable behaviors. Motus~\citep{motus} further explores unified world modeling for embodied intelligence by integrating multiple paradigms, including vision-language-action modeling, world modeling, video generation, inverse dynamics, and video-action joint prediction, into a MoT framework. InternVLA-A1~\citep{internvla_a1} and BagelVLA~\citep{hu2026bagelvlaenhancinglonghorizonmanipulation} incorporate visual reasoning, generation, and action prediction, while Cosmos3~\citep{agarwal2026cosmos} unifies language, vision, audio, and physical actions within a unified physical AI foundation model.

%% file: content/conclusion.tex
\vspace{-1pt}
\section{Conclusion}
\label{sec:conclusion}

We introduced \model, an embodied cognition foundation model that represents embodied plans through joint language-visual reasoning and imagination. The core idea is that embodied planning should connect the logical structure of a task with the physical states to be achieved: language specifies planning steps, constraints, and decisions, while visual imagination grounds them in goal and intermediate world states. To support this capability, we developed a unified multimodal architecture, an automatic pipeline for constructing joint text-visual planning supervision from embodied videos, and Rxbrain-Bench for evaluating visual understanding, visual goal imagination, and joint embodied planning. Experiments suggest that RxBrain can maintain general multimodal capabilities while beginning to produce plans with complementary textual and visual components. Meanwhile, action generation based on RxBrain achieves strong performance on real-world robotic tasks. We hope this work provides an initial step toward embodied foundation models that move beyond isolated understanding or generation and enable joint subgoal planning through symbolic and visual representations.

\paragraph{Limitations.} While RxBrain demonstrates the potential of joint language-visual reasoning and imagination for embodied planning, it still has several limitations. First, the current model is relatively small in scale, which may limit its ability to handle highly complex reasoning, fine-grained visual imagination, and long-horizon planning. Scaling both the model and the training data is likely to further improve the quality and consistency of its joint planning outputs. Second, although RxBrain can generalize across diverse embodied scenarios, fully out-of-distribution environments, embodiments, or task domains still require additional fine-tuning to achieve better performance. Third, interleaved generation involves two distinct generative paradigms—autoregressive modeling for text and flow matching for visual outputs—which introduces discrepancies between training and inference across modalities. Although we mitigate this issue to some extent by shifting ground-truth VAE latents during training, we have not yet incorporated reinforcement learning or other sequence-level optimization methods to further align and improve the interleaved generation process. Addressing this limitation requires further improvements to both the model architecture and the underlying training infrastructure.

\paragraph{Future work.} Future work will focus on two directions. First, we will scale RxBrain to larger model sizes. We expect larger models to strengthen joint textual-visual reasoning and imagination, enabling more coherent planning primitives, more accurate world state prediction, and more reliable joint subgoal planning. Beyond improving the quality of coupled joint subgoal plan representations, we also aim to enhance the model's agentic autonomy, allowing it to better decompose open-ended tasks, monitor progress, revise plans, and interact with environments in a more self-directed manner. These directions will further move RxBrain toward a more capable embodied cognition foundation model.

%% file: content/appendix.tex
\label{sec:appendix}
\section{Contributors and Acknowledgements}
\label{app:contributors}

\paragraph{Project Sponsors:} Zhengyou Zhang$^{\dagger}$ and Han Hu

\paragraph{Project Leaders:} Yufei Huang and Yuchun Guo

\paragraph{Core Contributors:} Haotian Liang$^{*}$, Mingkang Chen$^{*}$, Xiaomeng Zhu, Xiangli Shi, Kaixuan Wang, Yunxuan Mao, Weijie Zhou, Ling Chen

\paragraph{Contributors:} Shirong Zeng, Yueyu Long, Yuchen Si, Yajuan Zhu, Xingyu Zhou, Minghui Wang, Wanjia He, Xin Yang, Lingzhu Xiang, Zhiqing Liu, Bohan Ma, Xiran Huang, Xuantang Xiong, Zisheng Lu, Tianshuo Yang, Zhiheng Liu

\vspace{0.5em}

We appreciate contributions from Ping Luo and Yao Mu for fruitful discussions.

\vspace{0.5em}
\noindent
$^{\dagger}$ Corresponding author. \quad $^{*}$ Equal contribution.

\clearpage
\section{Data Sources}
\label{app:dataset-descriptions}

Section~\ref{sec:data} summarizes the corpus by four source categories. This appendix supplements that summary with per-dataset detail omitted there for space. Each processed source split is described by its collection setup, why it was included, and the hours retained after the verification of \Secref{sec:data-quality}. The final scope contains $46$ splits and $50{,}177$ retained hours, distributed as $20$ Real-Robot, $1$ UMI, $3$ Simulation, and $22$ Egocentric Human splits.

\subsection{Real-Robot Data}

\textbf{RealOmni-Open.} RealOmni-Open contributes over $13$ thousand hours of dual-hand household manipulation collected with a handheld DAS gripper, with fisheye video, trajectories, and tactile signals \citep{realominiopen}. Its scale and tactile modality extend the corpus with contact-rich household manipulation priors that complement the largely RGB sources, and we use $9{,}005.9$ hours after verification.

\textbf{JAKA.} JAKA data is self-collected on single-arm JAKA platforms, recording RGB observations with synchronized robot states and actions for deployment-aligned training. We use $2{,}558.2$ hours after verification.

\textbf{Xtrainer.} Xtrainer data is self-collected on dual-arm Xtrainer platforms for bimanual tasks, recording RGB observations with synchronized robot states and actions for deployment-aligned training. We use $1{,}515.9$ hours after verification.

\textbf{RoboMIND.} RoboMIND provides $107$ thousand real-world trajectories over $479$ tasks and $96$ object classes on four embodiments including the Franka Panda, UR-5e, AgileX dual-arm, and a Tien Kung humanoid, with failure demonstrations alongside successful ones \citep{wu2025robomind}. Its standardized collection protocol across embodiments gives consistent cross-embodiment supervision, and we use $888.0$ hours after verification.

\textbf{Galaxea Open-World.} The Galaxea Open-World Dataset provides over $500$ hours of real-world mobile manipulation on one uniform embodiment, with subtask-level language annotations across residential, kitchen, retail, and office scenes \citep{galaxea2025}. Its diverse scene coverage with consistent embodiment and fine-grained language annotations makes it a strong source of long-horizon mobile manipulation supervision, and we use $536.0$ hours after verification.

\textbf{MolmoAct2 RT-1.} This processed source split uses RT-1 real-robot manipulation episodes collected on a mobile manipulator for single-arm tabletop tasks \citep{brohan2022rt1,lee2025molmoact}. We count it as an independent MolmoAct2 split rather than inside a bundled source paragraph, and we retain $365.7$ hours after verification.

\textbf{LET Base / LejuRobot.} The LET dataset adds full-size humanoid manipulation teleoperated on Kuavo platforms across industrial, home, and service scenes \citep{lejurobot2025let}. Its humanoid embodiment complements the arm-centric sources, and we use $318.0$ hours from the \texttt{let\_base\_dataset} production split after verification.

\textbf{RoboCOIN.} RoboCOIN is a multi-embodiment bimanual dataset of more than $180$ thousand teleoperated demonstrations across $421$ tasks and $15$ robot platforms, with hierarchical trajectory, segment, and frame annotations \citep{robocoin2025}. Its broad bimanual coverage across diverse platforms provides rich cross-embodiment supervision for dexterous two-arm manipulation, and we use $315.9$ hours after verification.

\textbf{AgiBot World.} AgiBot World contributes more than $1$ million trajectories across $217$ tasks in five domains, namely domestic, retail, industrial, restaurant, and office, collected on mobile bimanual humanoids with multi-view RGB, depth, and language annotations for the overall task and each sub-step \citep{bu2025agibot}. Its standardized pipeline with human-in-the-loop verification yields high-quality long-horizon data, and we use $278.7$ hours from the AgiBotWorld2026 release after verification.

\textbf{DROID.} DROID is an in-the-wild single-arm dataset of $76$ thousand trajectories and about $350$ published hours, collected on Franka Panda robots across $564$ scenes and $86$ tasks in $52$ buildings, with three synchronized RGB views, depth, calibration, and language instructions per episode \citep{khazatsky2024droid}. Its scene and task diversity makes it a strong source of real gripper-object manipulation under varied viewpoints, and we retain $255.6$ hours after verification.

\textbf{MolmoAct2 SO100/101.} This processed source split covers the SO100/101 robot trajectories released with MolmoAct2 \citep{lee2025molmoact}. It is counted independently from the other MolmoAct2 robot splits, and we retain $180.3$ hours after verification.

\textbf{MolmoAct2 BC-Z.} This processed source split uses language- and video-conditioned BC-Z manipulation trajectories collected by interactive imitation across more than $100$ tasks \citep{jang2022bcz,lee2025molmoact}. It is counted as its own MolmoAct2 split, and we retain $148.9$ hours after verification.

\textbf{BridgeData V2.} BridgeData V2 contributes $60{,}096$ teleoperated trajectories on a low-cost WidowX arm across $24$ environments and $13$ skills, conditioned on language instructions or goal images and recorded with fixed and wrist RGB-D views \citep{walke2023bridgedata}. We include it because its broad task and environment variation transfers across labs, and we use $109.1$ hours from $47{,}651$ annotated trajectories after verification.

\textbf{MolmoAct2 Bridge.} This processed source split uses BridgeData V2 trajectories through the MolmoAct2 release \citep{walke2023bridgedata,lee2025molmoact}. It is counted separately from the base BridgeData V2 entry above, and we retain $105.1$ hours after verification.

\textbf{MolmoAct2 DROID.} This processed source split uses DROID trajectories through the MolmoAct2 release \citep{khazatsky2024droid,lee2025molmoact}. It is counted once under Real-Robot Data, separately from the base DROID entry above, and we retain $50.8$ hours after verification.

\textbf{HumanoidEveryday.} HumanoidEveryday contributes humanoid manipulation trajectories with RGB, depth, LiDAR, tactile streams, and language annotations \citep{zhao2025humanoideveryday}. Its open-world humanoid coverage adds an embodiment and sensing configuration not represented by the arm-only sources, and we use $31.8$ hours after verification.

\textbf{CyberOrigin.} CyberOrigin is a self-collected real-robot production split used for deployment-aligned manipulation supervision. The current project bibliography has no verified external reference for this source, so we make no broader collection-scale claim and report only the $30.3$ hours retained after verification.

\textbf{RH20T robot.} RH20T contains contact-rich robot sequences spanning many tasks and skill categories on multiple embodiments, with synchronized vision, force and torque, audio, and proprioception \citep{fang2024rh20t}. We count its robot stream as an independent processed source split and retain $596.4$ hours after verification.

\textbf{MolmoAct2 Tabletop.} This processed UMI-style split covers tabletop manipulation trajectories from MolmoAct2 \citep{lee2025molmoact}. We count it separately from the robot MolmoAct2 splits and retain $0.5$ hours after verification.

\textbf{MolmoAct2 Household.} This processed UMI-style split covers household manipulation trajectories from MolmoAct2 \citep{lee2025molmoact}. We count it separately from the robot MolmoAct2 splits and retain $0.6$ hours after verification.

\subsection{UMI Data}

\textbf{UMI (self-collected).} UMI data is gathered with handheld universal manipulation grippers that capture gripper-object contact from a wrist-mounted view \citep{chi2024umi}. We retain $17{,}505.8$ hours of UMI clips after verification.

\subsection{Simulation Data}

\textbf{BEHAVIOR-1K.} BEHAVIOR-1K defines $1{,}000$ everyday household activities grounded in $50$ scenes with more than $5{,}000$ annotated objects, simulated in the OmniGibson engine with physics for rigid bodies, deformable bodies, and liquids \citep{li2023behavior1k}. We use it to add scripted and precisely labeled state transitions that balance the noisier in-the-wild sources, sampling $1{,}102.4$ hours into the corpus.

\textbf{InternData-A1.} InternData-A1 is a hybrid synthetic-real manipulation dataset spanning four embodiments \citep{interndata_a1}. Its pick-and-place, articulated, and long-horizon tasks add structured manipulation trajectories at scale, and we use $197.8$ hours after verification.

\textbf{SynData.} SynData provides multimodal manipulation clips with egocentric RGB-D, hand joint states, and fingertip keypoints \citep{psibotai2025syndata}. Following the production taxonomy used throughout this report, we count this processed split under Simulation Data and retain $17.2$ hours after verification.

\subsection{Egocentric Human Data}

\textbf{Ego4D.} Ego4D is a massive egocentric corpus of roughly $3{,}670$ published hours of unscripted daily-life video recorded by more than $900$ camera wearers across $74$ locations in $9$ countries \citep{grauman2022ego4d}. We draw on its head-mounted hand-object manipulation segments, which expose dense everyday object-state changes under natural first-person motion, and retain $3{,}670.0$ hours after verification.

\textbf{AI Data curated.} The \texttt{ai\_data\_by\_keyword\_2.9m} curated library contains web-sourced RGB clips that have passed an initial in-house screening. It broadens visual and scene coverage but carries no single external reference, so we treat it as weak contextual supervision and retain $3{,}584.5$ hours after verification.

\textbf{AI Data raw.} The \texttt{aidata\_raw} library is the separate unfiltered web-video pool from the same internal pipeline. It remains distinct from the curated split because the two rows undergo different source-level screening, and we retain $1{,}918.2$ hours after verification.

\textbf{Egocentric-10K.} Egocentric-10K is a large release of about $10{,}000$ published hours of first-person factory video from over $2{,}000$ workers, with unusually high rates of visible hands and active manipulation \citep{egocentric10k}. It adds industrial hand-object work that household sources rarely cover, and we use $1{,}300.0$ hours after verification.

\textbf{EgoVerse.} EgoVerse is imported as a dedicated egocentric production split and contributes long-form first-person activity video to the open-world human-data pool. The current project bibliography contains no verified EgoVerse citation, so we report only its role in the pipeline and the $1{,}132.8$ hours retained after verification.

\textbf{EgoDex.} EgoDex is a large-scale dexterous manipulation dataset of $829$ published hours of $30$\,Hz egocentric video with $338$ thousand demonstrations over $194$ tabletop tasks, collected with Apple Vision Pro and carrying paired 3D hand and finger tracking with upper-body pose \citep{hoque2025egodex}. We include it because its precise per-frame hand pose gives a strong bridge from human demonstration to gripper control, retaining $829.0$ hours after quality verification.

\textbf{Xperience.} Xperience is a head-mounted first-person video source retained as a distinct production split. Because the current project bibliography contains no verified citation for it, we avoid attributing unverified release-scale details and report the $800.0$ hours retained after verification.

\textbf{Something-Something V2.} Something-Something V2 provides $220{,}847$ short clips of templated object manipulations \citep{goyal2017something}. Its label design forces a focus on temporal dynamics rather than static appearance, which suits short state-transition supervision, and we retain $240.0$ hours after verification.

\textbf{Ego-Exo4D.} Ego-Exo4D provides about $1{,}286$ published hours of time-synchronized first- and third-person video of skilled activities across cooking, repair, sports, and music, recorded by $740$ participants across $13$ cities \citep{grauman2024egoexo4d}. Its paired ego and exo views give cross-view supervision for the same action, and we use $155.1$ hours after filtering.

\textbf{EPIC-KITCHENS-100.} EPIC-KITCHENS-100 records $100$ published hours of unscripted kitchen activity in $45$ home kitchens, with roughly $90$ thousand densely segmented action instances labeled with verbs and nouns \citep{damen2022rescaling}. Its dense cooking, cleaning, and storage interactions add a fine-grained action vocabulary, and we retain $91.2$ hours after verification.

\textbf{EgoMe.} EgoMe contributes $7{,}902$ paired exocentric and egocentric videos, about $82$ published hours, that record a person first observing and then imitating a demonstrated action \citep{qiu2025egome}. The observe-then-imitate structure links third-person intent to first-person execution, and we use $43.9$ hours after verification.

\textbf{CaptainCook4D.} CaptainCook4D adds egocentric cooking recordings with both correct executions and deliberately induced errors \citep{peddi2024captaincook4d}. The error cases are useful for modeling deviations from a plan rather than only successful trajectories, and we use $39.4$ hours after verification.

\textbf{HD-EPIC.} HD-EPIC adds $41$ published hours of kitchen video in $9$ kitchens with highly detailed annotations grounded in 3D, including recipe steps, fine-grained actions, audio events, and object movements \citep{perrett2025hdepic}. Its dense labels give clean action and object references for state-transition supervision, and we retain $38.1$ hours after verification.

\textbf{HoloAssist.} HoloAssist records $166$ published hours of two-person assistive manipulation across $20$ object-centric tasks, captured in mixed reality with seven synchronized streams including RGB, depth, hand pose, and eye gaze \citep{wang2023holoassist}. Its mistakes and corrective interventions enrich the action distribution beyond clean demonstrations, and we use $32.2$ hours after verification.

\textbf{HOT3D.} HOT3D offers egocentric multi-view recordings captured with Project Aria and Quest 3 and supplies 3D hand and object tracking \citep{banerjee2025hot3d}. Its tracked hand-object interactions add contact-rich state changes, and we use $5.5$ hours after verification.

\textbf{TACO.} TACO is a bimanual tool-action-object dataset pairing third-person and egocentric views with 3D hand and object meshes and action labels \citep{liu2024taco}. Tool use broadens contact types and object functions beyond bare-hand grasping, and we use $3.0$ hours after verification.

\textbf{ARCTIC.} ARCTIC records dexterous bimanual manipulation of articulated objects from synchronized third-person and egocentric views, with 3D hand and object meshes and dense dynamic contact \citep{fan2023arctic}. It is a clean source for contact-rich state changes such as opening and closing, and we retain $2.3$ hours after verification.

\textbf{HOI4D.} HOI4D provides $2.4$ million RGB-D egocentric frames over $4{,}000$ sequences, with panoptic and motion segmentation, 3D hand pose, category-level object pose, and hand action \citep{liu2022hoi4d}. Its category-level coverage gives structured object-state supervision that raw video lacks, and we use $0.9$ hours after verification.

\textbf{HO-Cap.} HO-Cap captures uni- and bimanual hand-object interactions with synchronized RGB-D and per-frame 3D hand and object pose \citep{wang2025hocap}. Its pose-rich interactions complement the larger but weakly labeled video sources, and we use $0.6$ hours after verification.

\textbf{TASTE-Rob.} TASTE-Rob provides $100{,}856$ egocentric hand-object interaction videos, each aligned with a language instruction and recorded from a fixed viewpoint for interaction clarity \citep{zhao2025tasterob}. Its instruction-aligned clips suit clean language-to-interaction supervision for our planning formats, and we use $2.6$ hours after verification.

\textbf{RH20T human.} RH20T pairs its robot sequences with human demonstration video for the corresponding tasks \citep{fang2024rh20t}. We process that human-view stream separately from the robot stream and retain $98.4$ hours after verification.

\textbf{OakInk2.} OakInk2 organizes $627$ sequences of bimanual hand-object manipulation into three levels, namely affordance, primitive task, and complex task \citep{zhan2024oakink2}. Its multi-view images with body, hand, and object pose annotations supply hierarchical task structure for long-horizon supervision, and we use $74.6$ hours after verification.

\section{Additional Data Construction Details}
\label{app:data-construction-details}

\subsection{Temporal Segment Annotation Method Details}
\label{app:segment-method-details}
This appendix expands the three-step Temporal Segment Annotation pipeline of \Secref{sec:data-segmentation} with the full notation, derivations, and algorithmic detail that the main text compresses for space. The step order below follows the pipeline of \Figref{fig:data-pipeline}: a problem setup that fixes the segment representation, a content-adaptive keyframe sampler, a single global parse that proposes candidate steps, and a bidirectional boundary refinement that turns each coarse step into precise visual anchors. Throughout, a single multimodal large language model, written $\Phi$, makes every semantic decision, so the module is one annotator queried with different prompts rather than a stack of specialized detectors.

\paragraph{Problem setup and notation.}
A raw video is an ordered sequence of frames $V=(f_1,\dots,f_{N})$ recorded at $\nu$ frames per second, so frame $f_j$ carries the timestamp $\tau_j=j/\nu$ and the clip has duration $T=N/\nu$. A \newterm{keyframe} is a frame index retained for inspection, and a \newterm{candidate step} is an action phrase proposed for the whole clip. We define the annotation target from the bottom up, fixing the smallest unit first and then nesting it into the segment representation and the full per-video sequence.

\begin{itemize}[leftmargin=1.2em]
  \item \textbf{Atomic planning step (smallest unit).} Following the annotation prompt, an
  \newterm{atomic planning step} is one individual execution of a single manipulation whose effect
  is a clearly visible change of object state. Grounding the unit on a visible state change
  keeps each atomic planning step aligned with the state transition that \model{} learns to predict,
  so it serves as one supervision unit for the world model.
  \item \textbf{Segment (one localized atomic planning step).} A \newterm{segment} pins one atomic
  planning step to the two frames that expose its state change and attaches its supervision, giving
  the tuple $S_i=\bigl(I_i^{\mathrm{start}},\,I_i^{\mathrm{end}},\,a_i,\,d_i,\,m_i\bigr)$ of
  \Eqref{eq:segment}, whose components are fixed as follows.
  \begin{itemize}[leftmargin=1.2em]
    \item $I_i^{\mathrm{start}}$ and $I_i^{\mathrm{end}}$ are the start and end anchors, namely the two frames of $V$ that bound the visible state change.
    \item $a_i$ is the short step name of the atomic planning step.
    \item $d_i$ is the detailed textual description of that step.
    \item $m_i$ is metadata recording the dataset identifier, the camera viewpoint, the action index, and the temporal location.
  \end{itemize}
  \item \textbf{Annotation of a video (ordered sequence).} The annotation of $V$ is the temporally ordered sequence of segments $\{S_i\}_{i=1}^{M}$. Repeated executions of the same manipulation appear as separate segments, and neighboring segments may overlap after independent boundary refinement while retaining their chronological indices.
\end{itemize}

Each $S_i$ therefore corresponds to exactly one atomic planning step. Quality Verification consumes $S_i$, may correct its detailed description $d_i$, and passes retained segments to Segment Structuring (\Secref{sec:data-organization}). The rest of this appendix explains how this sequence is produced from weak video alone, by first proposing candidate steps over the whole clip and then refining each one to its anchor frames.

\paragraph{Content-adaptive keyframe sampling.}
A long clip holds far more frames than the annotator can read at once, and uniform sampling wastes this budget on static intervals while still missing brief manipulations, so we place keyframes where the visual content actually changes. We first decode the clip at a reduced rate $\nu_s=\min(4,\nu)$, downscale each decoded frame to $192\times108$ and convert it to the LUV color space through an operator $\psi$, then measure the motion between consecutive decoded frames $f_{j_{k-1}},f_{j_k}$ as the mean absolute LUV difference
\begin{equation}
\delta_k=\frac{1}{|\Omega|}\sum_{u\in\Omega}\bigl|\psi(f_{j_k})[u]-\psi(f_{j_{k-1}})[u]\bigr|,
\label{eq:diff}
\end{equation}
where $\Omega$ indexes the downscaled pixel-channel grid. We smooth $\delta$ with a moving average of odd width near $0.5\nu$ to suppress per-frame flicker, and we mark a frame as an anchor when its smoothed score $\bar\delta$ is a local maximum that exceeds the adaptive threshold
\begin{equation}
\theta_\delta=\max\!\Bigl(Q_{p_{\mathrm{lo}}}(\bar\delta),\ Q_{0.5}(\bar\delta)+\beta\,\bigl[Q_{p_{\mathrm{hi}}}(\bar\delta)-Q_{0.5}(\bar\delta)\bigr]\Bigr),
\label{eq:thr}
\end{equation}
in which $Q_p$ denotes the $p$-quantile of the smoothed scores, $p_{\mathrm{lo}}$ a low-quantile floor, $p_{\mathrm{hi}}$ a high quantile, and $\beta$ a margin coefficient. This makes $\theta_\delta$ a per-clip robust statistic with the median $Q_{0.5}$ as a baseline, the gap between $Q_{p_{\mathrm{hi}}}$ and the median as an adaptive dispersion margin scaled by $\beta$, and the floor $Q_{p_{\mathrm{lo}}}$ guarding low-activity clips against noise-driven fragmentation. We then prune anchors greedily under a minimum spacing $\Delta_{\min}$ keeping higher $\bar\delta$ first, always insert the first and last frames, fill any temporal gap wider than $g_{\max}$ by interpolation so long static stretches stay represented, and cap the anchor set $\gK$ at $A_{\max}$ frames. For the global parse we uniformly subsample $\gK$ to $n$ keyframes $\gG\subseteq\gK$, bounding the model context while preserving chronological coverage of the whole clip. In practice we set $p_{\mathrm{lo}}=0.8$, $p_{\mathrm{hi}}=0.9$, $\beta=0.2$, $\Delta_{\min}=0.2$\,s, $g_{\max}=0.5$\,s, $A_{\max}=50$, and $n=30$, empirical constants applied uniformly to every source rather than values from an optimality criterion.

\paragraph{MLLM global action parsing.}
With the keyframe set $\gG$ fixed, the annotator receives the $n$ global keyframes in chronological order, numbered $0$ to $n-1$, and returns in one pass a one-sentence summary, a verb-object goal, the camera viewpoint, a binary annotatability flag, and an ordered list of candidate steps. Each candidate step $c_i=(a_i,d_i,p_i,q_i)$ contains a short step name $a_i$, a detailed description $d_i$, and start and end indices $p_i$ and $q_i$ in $\gG$, with $0\le p_i\le q_i\le n-1$. The prompt holds each candidate to the atomic-planning-step definition above and enforces $q_i<p_{i+1}$, so the coarse index ranges are ordered and non-overlapping. This constraint applies only to the coarse candidates and does not prohibit overlap between the final independently refined segments. When the annotatability flag is false, for instance when the manipulated object is too small or the scene is too cluttered or dark to track a state change, the model returns an empty step list and the clip yields no segments. Otherwise each $c_i$ is lifted to a coarse segment $\tilde S_i$ by mapping its keyframe indices back to the frame identifiers and timestamps carried by $\gG$.

\paragraph{Local candidate-frame boundary refinement.}
A coarse segment $\tilde S_i$ inherits the timestamps of sparse keyframes, so its boundaries are biased away from the true transition, and we make this bias precise for the action onset. Let $j^\star$ be the true onset, namely the first frame at which the manipulated object has visibly changed. Because the global parse may place a boundary only on a sampled keyframe in $\gG$, and because the annotator can report a change only once the new state is visible, the coarse onset is the first sampled keyframe at or after $j^\star$,
\begin{equation}
\hat\jmath_0=\min\{\,j\in\gG\ :\ j\ge j^\star\,\}\ \ge\ j^\star ,
\label{eq:lag}
\end{equation}
so $\hat\jmath_0$ is an upward-biased estimate of $j^\star$ whose expected error scales with the keyframe spacing rather than with the frame period. Conditioning on history alone cannot remove this bias, since a labeler that reads only frames up to a candidate boundary keeps the previously observed state until the change is unambiguous, which can only push the estimate later. The remedy is to bracket each boundary with a dense window of frames that straddle it and to relocalize it among these candidates while the opposite anchor is shown in the same prompt for reference, which replaces the keyframe spacing in \Eqref{eq:lag} by the much finer candidate spacing and pulls the estimate back toward $j^\star$. The offset is treated symmetrically, so refinement recovers a tight anchor pair around the genuine state change instead of the inflated coarse window.

\begin{algorithm}[t]
\caption{Bidirectional boundary refinement for one segment $\tilde S_i$}
\label{alg:refine}
\begin{algorithmic}[1]
\Require coarse bounds $\tau^s$, $\tau^e$; short step name $a_i$; detailed description $d_i$; video $V$ ($T$\,s, $\nu$\,fps, $N$ frames); candidate count $n_c$; annotator $\Phi$; next-segment start $\tau^{s}_{i+1}$ ($=T$ if none)
\Ensure anchor pair $(I_i^{\mathrm{start}}, I_i^{\mathrm{end}})$ with $I_i^{\mathrm{start}} < I_i^{\mathrm{end}}$
  \State $\Delta \gets \max(0.3,\ \tau^e - \tau^s)$
  \State sample $n_c$ frames $E$ from $[\tau^s{+}0.3\Delta,\ \min(\tau^e{+}0.5\Delta,\, T)]$;\enspace $I_i^{\mathrm{end}} \gets \Phi_{\mathrm{end}}(\tau^s,\, E,\, a_i,\, d_i)$ \Comment{look-ahead past coarse end}
  \State sample $n_c$ frames $B$ from $[\max(0,\,\tau^s{-}0.5\Delta),\ \tau^s{+}0.7\Delta]$;\enspace $I_i^{\mathrm{start}} \gets \Phi_{\mathrm{start}}(I_i^{\mathrm{end}},\, \mathrm{reverse}(B),\, a_i,\, d_i)$ \Comment{look-back before coarse start}
  \If{$I_i^{\mathrm{start}} \ge I_i^{\mathrm{end}}$}
    \State swap $I_i^{\mathrm{start}}$ and $I_i^{\mathrm{end}}$ \Comment{enforce temporal order}
  \EndIf
  \If{$I_i^{\mathrm{start}} = I_i^{\mathrm{end}}$} \Comment{degenerate: extend end to avoid zero-length segment}
    \State $I_i^{\mathrm{end}} \gets \min(\lfloor\tau^{s}_{i+1}\cdot\nu\rfloor {-} 1,\; N{-}1)$ if later segment exists, else $\min(I_i^{\mathrm{start}}{+}\mathrm{round}(\nu),\; N{-}1)$
  \EndIf
  \State \Return $(I_i^{\mathrm{start}},\ I_i^{\mathrm{end}})$
\end{algorithmic}
\end{algorithm}

\Algref{alg:refine} states the procedure for one segment, with the same annotator $\Phi$ invoked under an end-selection prompt $\Phi_{\mathrm{end}}$ and a start-selection prompt $\Phi_{\mathrm{start}}$. Both prompts receive the short step name $a_i$ and the detailed description $d_i$. We refine the end before the start, because the action is easiest to recognize once its completed state is fixed, and the recovered end then anchors the search for the start over the $n_c=8$ candidates of each window straddling the coarse boundary. The two selections are sequential within one segment, while different coarse segments may be refined in parallel.

Two checks keep each emitted segment well formed. The order check in \Algref{alg:refine} places $I_i^{\mathrm{start}}$ before $I_i^{\mathrm{end}}$, while the degenerate-expansion branch prevents the anchor pair from collapsing onto one frame. These checks act within each segment and do not impose cross-segment non-overlap. The output retains the chronological candidate order, but adjacent segments may overlap after independent local refinement. Because each step retains exactly one start anchor and one end anchor, the module emits one state-transition pair per step rather than a dense per-frame label map. The two anchors are rendered from the source video at native resolution so downstream training does not inherit the reduced resolution used during annotation.

\begin{table}[t]
    \centering
    \scriptsize
    \caption{\textbf{Review items in Quality Verification.} Each row pairs a check item with
  the failure or reject reason it captures, grouped into semantic quality, visual grounding,
  and consistency between the text and visual transition.}
    \label{tab:data-review}
    \definecolor{tablegray}{gray}{0.92}
    \definecolor{tableblue}{RGB}{0,0,200}
    \renewcommand{\arraystretch}{0.8}
    \begin{tabular}{@{}>{\raggedright\arraybackslash\color{black}\scriptsize}p{0.32\linewidth}@{\hspace{4pt}}>{\raggedright\arraybackslash\columncolor{tablegray}\color{tableblue}\itshape\scriptsize}p{0.59\linewidth}@{}}
    \toprule
    \textbf{Check item} &
    \multicolumn{1}{>{\raggedright\arraybackslash}p{0.59\linewidth}@{}}{\textbf{Failure or reject reason}} \\
    \midrule

    \textit{Semantic quality} &
    \multicolumn{1}{>{\raggedright\arraybackslash}p{0.58\linewidth}@{}}{} \\
    \midrule
    Single atomic planning step with clean boundaries & Clip mixes steps, has unclear boundaries, or runs too long to bound the transition \\
    Manipulated target clearly visible & Target is too small, occluded, poorly lit, or lost in a cluttered background \\
    Frame free of overlays and edits & Subtitle, watermark, logo, or an editing cut contaminates a frame \\
    Detailed description matches the anchors & $d_i$ disagrees with the frames or is too weak to verify against them \\
    Genuine action with a unique end state & No physical action occurs, or the end state is ambiguous \\
    Both anchors readable & A start or end anchor is missing or cannot be read \\

    \midrule
    \textit{Visual grounding} &
    \multicolumn{1}{>{\raggedright\arraybackslash}p{0.58\linewidth}@{}}{} \\
    \midrule
    Hand presence in each anchor & Hand or gripper visibility that the consistency check compares against the text \\
    Robot arm presence & Robot arm or gripper visibility recorded for downstream filtering \\
    Object-exit status & Unintended loss of the target rather than an intended exit at completion \\

    \midrule
    \textit{Consistency} &
    \multicolumn{1}{>{\raggedright\arraybackslash}p{0.58\linewidth}@{}}{} \\
    \midrule
    Both frames valid and same scene & A frame is blank or single-color, or the two frames share no common scene \\
    Stable camera viewpoint & Camera viewing direction shifts between the two anchors \\
    Upright camera orientation & Camera orientation is inverted or gravity-defying \\
    Description consistent with the scene & $d_i$ contradicts the object identity, count, position, direction, or state change \\
    Hand changes reflected in text & Hand or gripper visibility changes but $d_i$ omits it \\
    \bottomrule
  \end{tabular}
\end{table}

\subsection{Quality Verification Review Items}
\label{app:data-review-items}

Table~\ref{tab:data-review} expands the Quality Verification summary in \Secref{sec:data-quality} into the review items used to inspect each pre-verification segment $S_i$. Semantic quality checks whether one observable state transition is localized and described correctly by $d_i$. Visual grounding records whether hands or grippers, robot arms, and manipulated objects are available for filtering, while consistency checks frame validity, viewpoint stability, and agreement between $d_i$ and the visual transition.

Quality Verification consumes $S_i$ and retains segments with reliable anchors and valid state changes. When the visual transition is valid but $d_i$ is inaccurate or incomplete, the verifier corrects $d_i$ instead of discarding the segment. Segment Structuring consumes only the retained segments after this verification step. Pairs that are invalid, ambiguous, corrupted, or inconsistent are rejected.

\section{Per-Scene Taxonomy Statistics}
\label{app:taxonomy-sunburst}

The following figures show the per-scene segment distribution for each of the fourteen scene categories in the pretraining corpus.
Each chart displays the action-subtype breakdown within that scene, with wedge area proportional to the number of verified trainable segments.

\begin{multicols}{3}
    \noindent\includegraphics[width=\linewidth, trim=0.2cm 0.2cm 0.2cm 0.2cm, clip]{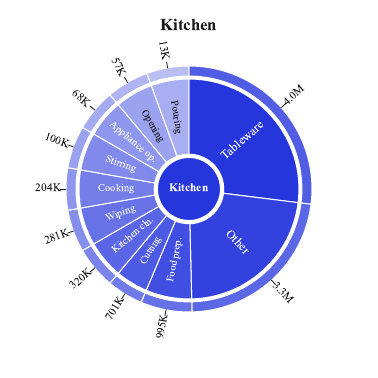}\vspace{2pt}
    \noindent\includegraphics[width=\linewidth, trim=0.2cm 0.2cm 0.2cm 0.2cm, clip]{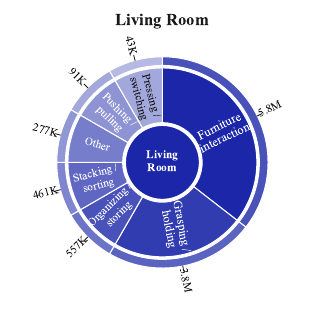}\vspace{2pt}
    \noindent\includegraphics[width=\linewidth, trim=0.2cm 0.2cm 0.2cm 0.2cm, clip]{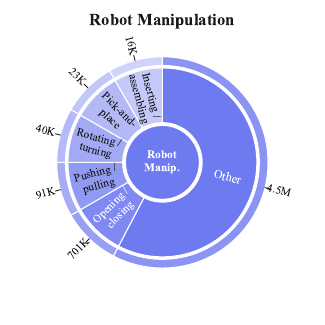}\vspace{2pt}
    \noindent\includegraphics[width=\linewidth, trim=0.2cm 0.2cm 0.2cm 0.2cm, clip]{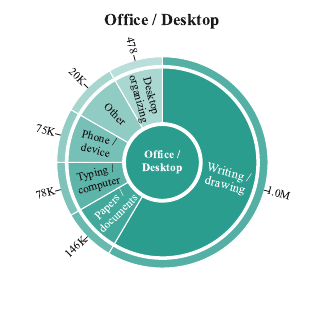}\vspace{2pt}
    \noindent\includegraphics[width=\linewidth, trim=0.2cm 0.2cm 0.2cm 0.2cm, clip]{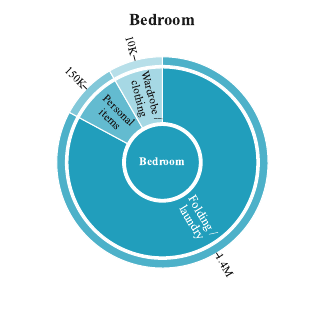}\vspace{2pt}
    \noindent\includegraphics[width=\linewidth, trim=0.2cm 0.2cm 0.2cm 0.2cm, clip]{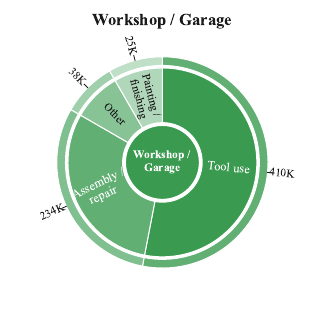}\vspace{2pt}
    \noindent\includegraphics[width=\linewidth, trim=0.2cm 0.2cm 0.2cm 0.2cm, clip]{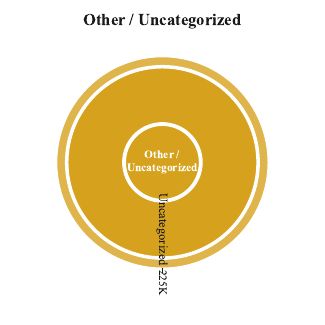}\vspace{2pt}
    \noindent\includegraphics[width=\linewidth, trim=0.2cm 0.2cm 0.2cm 0.2cm, clip]{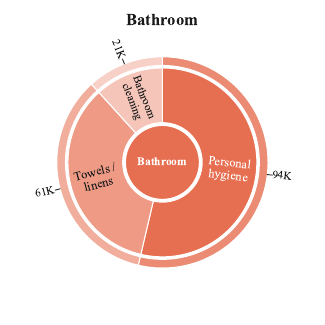}\vspace{2pt}
    \noindent\includegraphics[width=\linewidth, trim=0.2cm 0.2cm 0.2cm 0.2cm, clip]{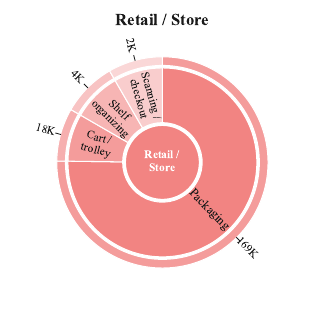}\vspace{2pt}
    \noindent\includegraphics[width=\linewidth, trim=0.2cm 0.2cm 0.2cm 0.2cm, clip]{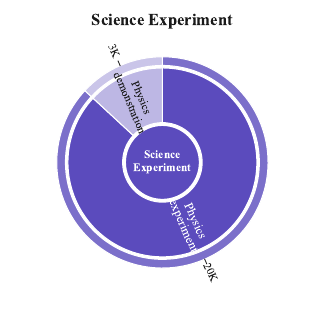}\vspace{2pt}
    \noindent\includegraphics[width=\linewidth, trim=0.2cm 0.2cm 0.2cm 0.2cm, clip]{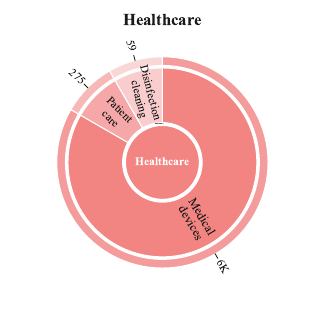}\vspace{2pt}
    \noindent\includegraphics[width=\linewidth, trim=0.2cm 0.2cm 0.2cm 0.2cm, clip]{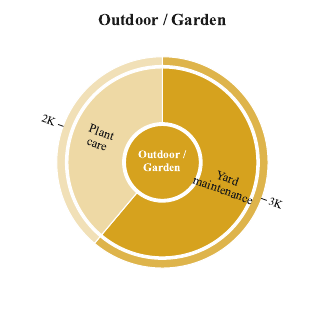}\vspace{2pt}
    \noindent\includegraphics[width=\linewidth, trim=0.2cm 0.2cm 0.2cm 0.2cm, clip]{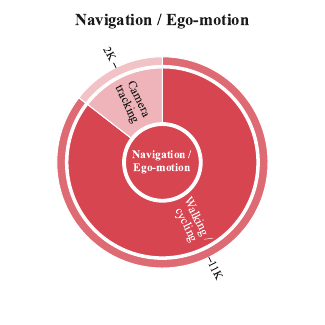}\vspace{2pt}
    \noindent\includegraphics[width=\linewidth, trim=0.2cm 0.2cm 0.2cm 0.2cm, clip]{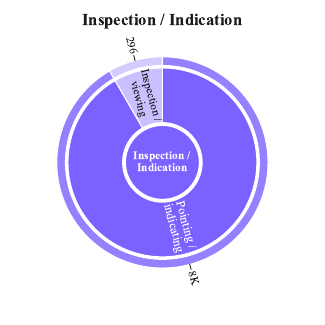}\vspace{2pt}
\end{multicols}
\captionof{figure}{\textbf{Per-scene taxonomy sunburst charts (14 scene categories).} Each
panel is one self-contained chart for a single scene category, flowing left-to-right and
top-to-bottom across three columns (and across pages as needed). The colored center disc names
the scene category; the ring around
it is divided into wedges, one per action subtype observed in that scene, with wedge angle
proportional to the subtype's share of verified trainable segments (very small subtypes are
padded to a minimum wedge width so their labels stay legible); the thin outer ring repeats the
same wedges and annotates each with its exact segment count. Reading any single chart therefore
gives both the relative composition and the absolute segment count of every action subtype
within that scene category.}
\label{fig:taxonomy-sunburst-all}
\section{Training Configuration and Data Mixture}
\label{app:training-config}

Table~\ref{tab:train-hparams} summarizes the optimization and sequence-packing
configuration for the two training stages: large-scale pretraining (Stage~1) and
supervised fine-tuning (Stage~2). Stage~1 initializes all
modules of HY-Embodied-0.5 as trainable and emphasizes the flow-matching
generation loss ($\lambda_{\mathrm{CE}}{=}0.25$, $\lambda_{\mathrm{FM}}{=}1.0$);
Stage~2 resumes from the Stage~1 final checkpoint and balances the two losses
($\lambda_{\mathrm{CE}}{=}\lambda_{\mathrm{FM}}{=}1.0$), with a lower peak learning
rate, a longer packed sequence, and a smaller per-device batch. Both stages use
AdamW (weight decay $0.01$), a cosine schedule, bf16 precision with ZeRO-2, a
$16{\times}$-downsampling VAE at $\le 256$\,px generation, flow-matching timestep
shift $1.0$, and $0.1$ classifier-free-guidance text-condition dropout.

  \begin{table}[H]
    \centering
    \small
    \caption{\textbf{Stage~1 data mixture.} Two buckets mixed at integer ratio
    $6{:}4$ ($\Sigma{=}10$); the per-step share is each bucket's ratio divided by $10$.}
    \label{tab:data-mix-s1}
    \renewcommand{\arraystretch}{1.15}
    \begin{tabular}{@{}llrrr@{}}
      \toprule
      Bucket & Content & Scale & Ratio & Per-step share \\
      \midrule
      b0 & Text-to-image (LAION $+$ COYO)          & $202.5$M & $6$ & $60.0\%$ \\
      b1 & Understanding (LLaVA-OV-1.5 $+$ MAmmoTH) & $113.4$M & $4$ & $40.0\%$ \\
      \bottomrule
    \end{tabular}
  \end{table}  
\paragraph{Data mixture (Stage~1).} The Stage~1 corpus is drawn from two buckets mixed
  at the integer ratio $6{:}4$ ($\Sigma{=}10$), which sets each bucket's expected share of
  optimization steps (Table~\ref{tab:data-mix-s1}). In contrast to Stage~2, the Stage~1
  mixture is deliberately generation-heavy so as to bootstrap the flow-matching path:
  web-scale text-to-image pairs (b0; $202.5$M pairs from LAION and COYO) are the larger
  source at $60.0\%$, while a single large-scale vision--language understanding bucket
  (b1; $113.4$M examples---the LLaVA-OneVision-1.5 mid-training corpus together with
  LLaVA-OneVision-1.5-Instruct and the MAmmoTH SI/OV collections) supplies the remaining
  $40.0\%$ to co-train the understanding path from the outset. The text-to-image bucket is
  sampled length-source.
\begin{table}[H]
  \centering
  \small
  \caption{\textbf{Training hyperparameters for the two stages.} Stage~1 is
  large-scale pretraining from HY-Embodied-0.5; Stage~2 is supervised fine-tuning resumed from the Stage~1 checkpoint. Global batch $=$ per-device batch
  $\times$ gradient accumulation $\times$ GPUs.}
  \label{tab:train-hparams}
  \renewcommand{\arraystretch}{1.15}
  \begin{tabular}{@{}lll@{}}
    \toprule
    Hyperparameter & Stage~1 (Pretraining) & Stage~2 (SFT) \\
    \midrule
    Initialization & HY-Embodied-0.5 (all modules trainable) & Stage-1 final checkpoint \\
    Loss weights ($\lambda_{\mathrm{CE}},\lambda_{\mathrm{FM}}$) & $0.25\,/\,1.0$ & $1.0\,/\,1.0$ \\
    Optimizer & AdamW (weight decay $0.01$) & AdamW (weight decay $0.01$) \\
    Peak learning rate & $2\times10^{-4}$ & $6.5\times10^{-5}$ \\
    LR schedule & cosine, $800$ warmup & cosine, $1050$ warmup \\
    Per-device batch & $20$ & $10$ \\
    Gradient accumulation & $1$ & $1$ \\
    GPUs (nodes $\times$ 8) & $368$ ($46\times8$) & $312$ ($39\times8$) \\
    Global batch & $7{,}360$ & $3{,}120$ \\
    Precision / parallelism & bf16 / ZeRO-2 & bf16 / ZeRO-2 \\
    Epochs & $1$ & $1$ \\
    Max sequence length & $8{,}192$ & $9{,}216$ \\
    Understanding pixel budget & $1{,}048{,}576$ & $524{,}288$ \\
    Generation resolution & $\le 256$\,px (VAE downsample $16$) & $\le 256$\,px \\
    Timestep shift (flow) & $1.0$ & $1.0$ \\
    CFG text-cond.\ dropout & $0.1$ & $0.1$ \\
    \bottomrule
  \end{tabular}
\end{table}

\paragraph{Data mixture (Stage~2).} The Stage~2 corpus is drawn from six buckets
mixed at the integer ratio $100{:}12{:}150{:}70{:}90$ ($\Sigma{=}422$), which
sets each bucket's expected share of optimization steps (Table~\ref{tab:data-mix}).
The mixture is deliberately understanding-heavy: general and embodied understanding
(b2) is the single largest source at $34.6\%$, and the two embodied generative
buckets---multi-frame world-model prediction (b3) and interleaved planning
(b4)---together account for $37.0\%$, while text-to-image data (b0, b1) preserves
general image-generation quality. Understanding buckets are sampled length-source.
\begin{table}[H]
  \centering
  \small
  \caption{\textbf{Stage~2 data mixture.} Six buckets mixed at integer ratio
  $100{:}12{:}150{:}70{:}90$ ($\Sigma{=}422$); the per-step share is each
  bucket's ratio divided by $422$.}
  \label{tab:data-mix}
  \renewcommand{\arraystretch}{1.15}
  \begin{tabular}{@{}llrrr@{}}
    \toprule
    Bucket & Content & Scale & Ratio & Per-step share \\
    \midrule
    b0 & T2I base (BLIP3o cascade-filtered) & $12.0$M & $100$ & $23.1\%$ \\
    b1 & T2I premium (BLIP3o-60k $+$ Self-Build T2I dataset-690k) & $0.75$M & $12$ & $2.8\%$ \\
    b2 & Understanding / MMU (spatial CoT $+$ knowledge $+$ pointing) & $20.8$M & $150$ & $34.6\%$ \\
    b3 & Embodied multi-frame world model & $8.5$M & $70$ & $16.2\%$ \\
    b4 & Embodied interleaved planning & $10.3$M & $90$ & $20.8\%$ \\
    \bottomrule
  \end{tabular}
\end{table}

\section{RxBrain-Bench: Generation-Side Evaluation Protocol}
\label{app:genbench-protocol}

This appendix details the evaluation methodology for the two \emph{generative}
tracks of RxBrain-Bench---interleaved planning (IL) and multi-frame video
generation (MF)---including the task definitions, the VLM-as-judge protocol and
rubrics, the local perceptual pass, and how each baseline is composed into a
comparable generator.

\paragraph{Task definitions and scale.}
\emph{IL (interleaved planning)} takes an observation ($1$--$3$ current-scene
frames) and a task instruction, and produces an interleaved plan
$[(\text{subtask text}_1, \text{goal image}_1), \dots]$: at every step the model
emits both the next subtask in language and the goal image depicting the state
after that subtask. The held-out set contains $4{,}756$ trajectories across $11$
datasets (\texttt{jaka}, \texttt{egodex}, \texttt{xtrainer}, \texttt{tasterob},
\texttt{taco}, \texttt{umi}, \texttt{egodex\_full}, \texttt{arctic},
\texttt{bridgev2}, \texttt{let}, \texttt{let\_base}), with planning depth
$n{=}1$--$5$ (counts $n_1{=}1536$, $n_2{=}749$, $n_3{=}366$, $n_4{=}1366$,
$n_5{=}739$). For cross-model comparison we draw a
\texttt{balanced\_subset(seed{=}0)} of $540$ records stratified by
dataset~$\times$~depth, so all agents are judged on the identical records.
\emph{MF (multi-frame generation)} takes an observation ($1$--$4$ frames) and an
instruction and produces a short opening text plus $N$ continuous future frames
depicting the frame-by-frame unfolding of a \emph{single} action (a motion clip,
not a multi-step plan). The held-out set contains $1{,}116$ continuous
trajectories, each with exactly $4$ ground-truth keyframes ($3$ intermediate $+$
$1$ end).

\paragraph{Judge protocol.}
Both tracks are scored by a GPT-5.5 VLM-as-judge (Azure Responses API), with one
multi-image call per record that is forced to emit a JSON schema of per-criterion
scores in $[0,1]$; we average $\text{votes}{=}2$ independent calls. All images are
resized to a longest side of $512$\,px before upload to bound token cost. The IL
judge (Table~\ref{tab:il-rubric}) receives the task, the observation frames, and
per step the GT subtask text, the model subtask text, the GT goal image, and the
model-generated image; the MF judge (Table~\ref{tab:mf-rubric}) receives the task,
the observation, the model-generated frame sequence, and the GT future frames.

\begin{table}[H]
  \centering
  \small
  \caption{\textbf{IL judge rubric} (weighted sum $S_{\mathrm{plan}}$). The first
  five criteria are scored by GPT-5.5; image similarity is supplied by the local
  perceptual pass (DINO cosine). Per-step criteria are averaged over the steps of a
  record.}
  \label{tab:il-rubric}
  \renewcommand{\arraystretch}{1.15}
  \begin{tabular}{@{}rll>{\raggedright\arraybackslash}p{0.42\linewidth}@{}}
    \toprule
    Weight & Criterion & Granularity & What it measures \\
    \midrule
    $10\%$ & observation\_understanding & record & whether the plan reads the current obs state correctly \\
    $25\%$ & subtask\_planning & per-step & whether each subtask is a reasonable next action \\
    $25\%$ & goal\_image\_correctness & per-step & whether the generated image depicts the subtask goal state (vs.\ GT goal) \\
    $20\%$ & subtask\_goal\_consistency & per-step & whether the generated image matches the model's own subtask text \\
    $10\%$ & chain\_completion & record & whether executing the full chain would complete the overall task \\
    $10\%$ & image\_similarity & per-step & local DINO cosine  \\
    \bottomrule
  \end{tabular}
\end{table}

\begin{table}[H]
  \centering
  \small
  \caption{\textbf{MF judge rubric} (weighted sum $S_{\mathrm{gen}}$). MF is a
  single action, so all five criteria are record-level.}
  \label{tab:mf-rubric}
  \renewcommand{\arraystretch}{1.15}
  \begin{tabular}{@{}rl>{\raggedright\arraybackslash}p{0.62\linewidth}@{}}
    \toprule
    Weight & Criterion & What it measures \\
    \midrule
    $15\%$ & observation\_continuity & whether the first generated frame continues the obs naturally \\
    $30\%$ & action\_correctness & whether the frames execute the instructed action on the correct object \\
    $20\%$ & goal\_completion & whether the final frame reaches the task's implied goal state \\
    $20\%$ & temporal\_plausibility & whether inter-frame motion is smooth and physically plausible \\
    $15\%$ & gt\_trajectory\_match & whether the generated sequence matches the GT future at the action level \\
    \bottomrule
  \end{tabular}
\end{table}

\paragraph{Perceptual pass.}
For every (GT, generated) frame pair we compute LPIPS, CLIP cosine, and DINO cosine
locally on CPU; the DINO cosine is used as the IL \emph{image\_similarity} term.
Because $S_{\mathrm{plan}}$ is a fixed-weight sum that simply drops a missing term
without renormalizing, a record whose perceptual pass has not run forfeits the full
$10\%$ image-similarity weight rather than redistributing it; the perceptual pass
must therefore precede judging for every record. To hide latency, generation
(GPU-bound) is overlapped with the perceptual pass (CPU) and the judge (API): a
rolling driver launches perceptual$\rightarrow$judge on each batch of $\sim70$
completed records, de-duplicating by a unique per-record key so no record is judged
twice.

\paragraph{IL baselines: composing agents into interleaved planners.}
The judge is architecture-agnostic: any pipeline that emits a (text, goal image)
pair per step can be scored by the same rubric. Each baseline is a
\textsc{reasoner} (scene $\rightarrow$ stepwise text) plus a \textsc{generator}
(subtask $+$ previous frame $\rightarrow$ goal image) run under \emph{free-running}
rollout---the step-$k$ goal image is conditioned on the step-$(k{-}1)$
\emph{generated} image (step $0$ on the last obs frame), which is exactly the
error-cascading regime the judge scores.
\begin{itemize}[leftmargin=1.4em,itemsep=2pt,topsep=2pt]
  \item \textbf{Qwen-Agent} ($S_{\mathrm{plan}}{=}0.431$): a heterogeneous pair of
  specialist models. Reasoner Qwen3-VL-2B-Instruct (greedy, prompted to emit
  exactly $n$ steps); generator Qwen-Image-Edit as standard img2img editing
  (\texttt{image}${=}$previous frame, \texttt{prompt}${=}$subtask,
  $\text{cfg}{=}4.0$, $40$ steps). The generator has no end-to-end goal modeling,
  giving the lowest goal-image and image-similarity scores.
  \item \textbf{Cosmos3-Nano} ($S_{\mathrm{plan}}{=}0.522$): a single omni model
  run in two passes. The reasoner is a plain VLM chat call; the generator cannot
  produce an obs-conditioned single image (its pure-image mode is text-to-image and
  discards the conditioning frame), so we instead generate a short image-to-video
  clip (first frame $=$ previous image, $17$ frames, $30$ steps, $\text{cfg}{=}5.0$)
  and take its last frame as the goal image. Because the chat and video services are
  mutually exclusive, planning and generation run as two separate service passes.
  \item \textbf{BAGEL-7B-MoT} ($S_{\mathrm{plan}}{=}0.503$): a single unified model
  in one local pass---the simplest of the three. The reasoner is the
  understanding path of \texttt{interleave\_inference}; the generator is BAGEL's
  native image editing, where the previous frame's VAE latent enters the generation
  context and the classifier-free-guidance image scale ($\text{cfg\_img}{=}2.0$,
  with $\text{cfg\_text}{=}4.0$, timestep shift $3.0$, $30$ steps) pulls sampling
  toward the obs-conditioned prediction---no image-to-video detour is needed.
\end{itemize}
Overall Cosmos3-Nano ($0.522$) $>$ BAGEL ($0.503$) $>$ Qwen-Agent ($0.431$): both
unified omni models beat the heterogeneous pair, and the stronger reasoner
(Cosmos3-Nano) converts its advantage into a higher planning score, whereas BAGEL's
native editing yields the best image--text consistency of the three.

\paragraph{MF baselines: continuous video generation.}
MF needs no reasoner---only obs $+$ instruction $\rightarrow$ video. Both baselines
are image-to-video models that natively accept a single first frame, so both use the
obs \emph{last} frame (the GT $t_0$ anchor) as the first frame with the instruction
in the prompt, generate a continuous clip, drop the first frame, and evenly
subsample $4$ frames aligned to the GT's $4$ keyframes.
\begin{itemize}[leftmargin=1.4em,itemsep=2pt,topsep=2pt]
  \item \textbf{Wan2.2-TI2V-5B} ($S_{\mathrm{gen}}{=}0.397$): a general-purpose I2V
  diffusion model ($49$ frames). It has the strongest observation continuity
  ($0.73$) but weak action correctness ($0.39$)---the characteristic gap of a
  general video model on fine-grained embodied manipulation (visual fluency
  $\ne$ correct action).
  \item \textbf{Cosmos3-Nano} ($S_{\mathrm{gen}}{=}0.575$): an embodied omni world
  model served as an I2V endpoint (first frame $+$ prompt). Its advantage is
  concentrated in action semantics---action correctness $0.60$ vs.\ Wan's $0.39$.
\end{itemize}
Cosmos3-Nano ($0.575$) leads Wan2.2-TI2V ($0.397$) across the board, with the gap
driven mainly by action correctness and goal completion.

Section~\ref{sec:data} summarizes the corpus by four source categories. This appendix supplements that summary with per-dataset detail omitted there for space. Each processed source split is described by its collection setup, why it was included, and the hours retained after the verification of \Secref{sec:data-quality}. The final scope contains $46$ splits and $50{,}177$ retained hours, distributed as $20$ Real-Robot, $1$ UMI, $3$ Simulation, and $22$ Egocentric Human splits.

\section{Inference Acceleration for Real-Robot Deployment}
\label{app:inference-accel}

In real-robot experiments the policy is served through the asynchronous
inference stack of HyVLA-0.5~\citep{zhang2026hyembodied}. Because the unoptimized
model has high per-step latency, asynchronous serving injects observation lag
that surfaces as jitter and degrades fine manipulation, motivating the
infrastructure-level acceleration described here. Every optimization is
constrained to be \emph{numerically lossless} at the single-trajectory level:
for each candidate we run an open-loop ground-truth rollout over a full
trajectory, overlay the predicted end-effector trajectory against ground truth
per dimension, and quantify position, orientation, and gripper error. An
optimization is accepted only when its fitted curves and errors match the
baseline, after which we compare its latency.

The policy generates an action chunk by iterating a flow-matching Euler ODE.
Each forward pass processes a fixed prefix of $\approx 989$ tokens (text,
vision, and proprioceptive state) together with a $16$-token action chunk, and
one prediction requires $K$ forward passes ($K{=}2$ at deployment). Profiling attributes the $210$\,ms baseline not to matrix
multiplication but to Mixture-of-Transformers (MoT) modality dispatch: every
layer recomputes the text/vision/action token routing via
\texttt{nonzero(modality\_mask==k)}, so a single forward triggers
$\approx 1091$ \texttt{aten::nonzero} calls ($174$\,ms CPU), compounded by
\texttt{index\_put\_} ($48$\,ms), \texttt{index} ($27$\,ms), and \texttt{copy\_}
($21$\,ms). The boolean indexing and \texttt{.any()} guards force repeated
GPU$\rightarrow$CPU synchronization and fragment the kernel stream, serializing
the pipeline. Because the token modality layout is fixed per request, this
routing can be computed once and reused.

\begin{table}[h]
  \centering
  \small
  \caption{\textbf{Per-frame inference latency}. All configurations are numerically equivalent within the bf16
  rounding level (position $<0.65$\,mm, orientation $<0.73^\circ$).}
  \label{tab:infer-latency}
  \renewcommand{\arraystretch}{1.15}
  \begin{tabular}{@{}lrr@{}}
    \toprule
    Configuration & Latency (ms) & Speedup \\
    \midrule
    Baseline (per-layer boolean dispatch) & $210$ & $1.00\times$ \\
    \;\;+ prefix KV cache \& action-only decode & \multicolumn{2}{c}{\emph{(lossless)}} \\
    \;\;+ de-synchronized dispatch \& CUDA graph & $143$ & $1.47\times$ \\
    \bottomrule
  \end{tabular}
\end{table}

We therefore apply three lossless optimizations.
\textbf{(i) Prefix KV caching:} the prefix is invariant across the $K$ ODE
steps, so we run it once per observation, cache the per-layer keys and values,
and reuse them, avoiding recomputation over the $\approx 989$ prefix tokens.
\textbf{(ii) Action-only fast decoding:} given the cached prefix, each ODE step
forwards only the $16$ action tokens---attending as queries over the cached
prefix and evaluating only the action-expert FFN---reducing per-step cost from
$O(989{+}16)$ to $O(16)$.
\textbf{(iii) De-synchronized modality dispatch} (the largest single gain): we
cache the precomputed integer routing indices keyed by
\texttt{id(modality\_mask)} and reuse them across all $32$ layers, and replace
boolean indexing with \texttt{index\_select} (gather) and \texttt{index\_copy\_}
(scatter). The rewrite is numerically equivalent (ascending indices, identical
write-back positions) and removes the per-layer \texttt{nonzero} calls and their
synchronization. Finally, the fixed-shape decoding step is captured and replayed
with a CUDA graph to eliminate per-kernel launch overhead.

Table~\ref{tab:infer-latency} summarizes the result. The combined optimizations
reduce per-frame latency from $210$\,ms to $143$\,ms ($-32\%$, $1.47\times$)
while keeping the output numerically equivalent to the baseline (position
deviation $<0.65$\,mm, orientation $<0.73^\circ$, i.e.\ at the bf16 rounding
level), satisfying the latency budget for real-time closed-loop control. We
further verified that FP8 quantization and image-token reduction yield no net
gain at this model scale (hidden size $2048$), making $143$\,ms the lossless
latency floor.

%% file: biblio.bib
@article{chameleon,
  title         = {Chameleon: Mixed-Modal Early-Fusion Foundation Models},
  author        = {{Chameleon Team}},
  journal       = {arXiv preprint arXiv:2405.09818},
  year          = {2024}
}

@article{janus,
  title         = {Janus: Decoupling Visual Encoding for Unified Multimodal
                   Understanding and Generation},
  author        = {Wu, Chengyue and Chen, Xiaokang and Wu, Zhiyu and Ma, Yiyang
                   and Liu, Xingchao and Pan, Zizheng and Liu, Wen and Xie, Zhenda
                   and Yu, Xingkai and Ruan, Chong and Luo, Ping},
  journal       = {arXiv preprint arXiv:2410.13848},
  year          = {2024}
}

@article{bagel,
  title         = {Emerging Properties in Unified Multimodal Pretraining},
  author        = {Deng, Chaorui and Zhu, Deyao and Li, Kunchang and Gou, Chenhui
                   and Li, Feng and Wang, Zeyu and Zhong, Shu and Yu, Weihao and
                   Nie, Xiaonan and Song, Ziang and Shi, Guang and Fan, Haoqi},
  journal       = {arXiv preprint arXiv:2505.14683},
  year          = {2025}
}

@inproceedings{sd3,
  title         = {Scaling Rectified Flow Transformers for High-Resolution Image
                   Synthesis},
  author        = {Esser, Patrick and Kulal, Sumith and Blattmann, Andreas and
                   Entezari, Rahim and M{\"u}ller, Jonas and Saini, Harry and
                   Levi, Yam and Lorenz, Dominik and Sauer, Axel and Boesel, Frederic
                   and Podell, Dustin and Dockhorn, Tim and English, Zion and
                   Lacey, Kyle and Goodwin, Alex and Marek, Yannik and Rombach, Robin},
  booktitle     = {International Conference on Machine Learning (ICML)},
  year          = {2024}
}

@article{flux,
  title         = {{FLUX.1} Kontext: Flow Matching for In-Context Image Generation
                   and Editing in Latent Space},
  author        = {{Black Forest Labs}},
  journal       = {arXiv preprint arXiv:2506.15742},
  year          = {2025}
}

@inproceedings{ldm,
  title         = {High-Resolution Image Synthesis with Latent Diffusion Models},
  author        = {Rombach, Robin and Blattmann, Andreas and Lorenz, Dominik and
                   Esser, Patrick and Ommer, Bj{\"o}rn},
  booktitle     = {IEEE/CVF Conference on Computer Vision and Pattern Recognition
                   (CVPR)},
  year          = {2022}
}

@inproceedings{rt2,
  title         = {{RT-2}: Vision-Language-Action Models Transfer Web Knowledge to
                   Robotic Control},
  author        = {Brohan, Anthony and Brown, Noah and Carbajal, Justice and
                   Chebotar, Yevgen and Chen, Xi and others},
  booktitle     = {Conference on Robot Learning (CoRL)},
  year          = {2023}
}

@inproceedings{openvla,
  title         = {OpenVLA: An Open-Source Vision-Language-Action Model},
  author        = {Kim, Moo Jin and Pertsch, Karl and Karamcheti, Siddharth and
                   Xiao, Ted and Balakrishna, Ashwin and Nair, Suraj and
                   Rafailov, Rafael and Foster, Ethan and Lam, Grace and
                   Sanketi, Pannag and Vuong, Quan and Kollar, Thomas and
                   Burchfiel, Benjamin and Tedrake, Russ and Sadigh, Dorsa and
                   Levine, Sergey and Liang, Percy and Finn, Chelsea},
  booktitle     = {Conference on Robot Learning (CoRL)},
  year          = {2024}
}

@article{pi0,
  title         = {{$\pi_0$}: A Vision-Language-Action Flow Model for General Robot
                   Control},
  author        = {Black, Kevin and Brown, Noah and Driess, Danny and Esmail, Adnan
                   and Equi, Michael and Finn, Chelsea and Fusai, Niccolo and
                   Groom, Lachy and Hausman, Karol and Ichter, Brian and others},
  journal       = {arXiv preprint arXiv:2410.24164},
  year          = {2024}
}

@article{pi05,
  title         = {{$\pi_{0.5}$}: A Vision-Language-Action Model with Open-World
                   Generalization},
  author        = {{Physical Intelligence} and Black, Kevin and Brown, Noah and
                   Darpinian, James and Finn, Chelsea and Levine, Sergey and others},
  journal       = {arXiv preprint arXiv:2504.16054},
  year          = {2025}
}

@inproceedings{palme,
  title         = {{PaLM-E}: An Embodied Multimodal Language Model},
  author        = {Driess, Danny and Xia, Fei and Sajjadi, Mehdi S. M. and
                   Lynch, Corey and Chowdhery, Aakanksha and Ichter, Brian and
                   Wahid, Ayzaan and Tompson, Jonathan and Vuong, Quan and Yu, Tianhe
                   and others},
  booktitle     = {International Conference on Machine Learning (ICML)},
  year          = {2023}
}

@inproceedings{genie,
  title         = {Genie: Generative Interactive Environments},
  author        = {Bruce, Jake and Dennis, Michael and Edwards, Ashley and
                   Parker-Holder, Jack and Shi, Yuge and Hughes, Edward and
                   Lai, Matthew and Mavalankar, Aditi and Steigerwald, Richie and
                   Apps, Chris and others},
  booktitle     = {International Conference on Machine Learning (ICML)},
  year          = {2024}
}

@misc{genie2,
  title         = {Genie 2: A Large-Scale Foundation World Model},
  author        = {{Google DeepMind}},
  year          = {2024},
  howpublished  = {\url{https://deepmind.google/discover/blog/genie-2-a-large-scale-foundation-world-model/}},
  note          = {Blog post}
}

@inproceedings{grauman2022ego4d,
  title         = {{Ego4D}: Around the World in 3,000 Hours of Egocentric Video},
  author        = {Grauman, Kristen and Westbury, Andrew and Byrne, Eugene and
                   Chavis, Zachary and Furnari, Antonino and Girdhar, Rohit and
                   Hamburger, Jackson and Jiang, Hao and Liu, Miao and Liu, Xingyu
                   and others},
  booktitle     = {IEEE/CVF Conference on Computer Vision and Pattern Recognition
                   (CVPR)},
  year          = {2022}
}

@inproceedings{grauman2024egoexo4d,
  title         = {{Ego-Exo4D}: Understanding Skilled Human Activity from First-
                   and Third-Person Perspectives},
  author        = {Grauman, Kristen and Westbury, Andrew and Torresani, Lorenzo and
                   Kitani, Kris and Malik, Jitendra and others},
  booktitle     = {IEEE/CVF Conference on Computer Vision and Pattern Recognition
                   (CVPR)},
  year          = {2024}
}

@inproceedings{goyal2017something,
  title         = {The ``Something Something'' Video Database for Learning and
                   Evaluating Visual Common Sense},
  author        = {Goyal, Raghav and Ebrahimi Kahou, Samira and Michalski, Vincent
                   and Materzynska, Joanna and Westphal, Susanne and Kim, Heuna and
                   Haenel, Valentin and Fruend, Ingo and Yianilos, Peter and
                   Mueller-Freitag, Moritz and others},
  booktitle     = {IEEE International Conference on Computer Vision (ICCV)},
  year          = {2017}
}

@article{damen2022rescaling,
  title         = {Rescaling Egocentric Vision: Collection, Pipeline and Challenges
                   for {EPIC-KITCHENS-100}},
  author        = {Damen, Dima and Doughty, Hazel and Farinella, Giovanni Maria and
                   Furnari, Antonino and Ma, Jian and Kazakos, Evangelos and
                   Moltisanti, Davide and Munro, Jonathan and Perrett, Toby and
                   Price, Will and Wray, Michael},
  journal       = {International Journal of Computer Vision (IJCV)},
  volume        = {130},
  number        = {1},
  pages         = {33--55},
  year          = {2022}
}

@inproceedings{perrett2025hdepic,
  title         = {{HD-EPIC}: A Highly-Detailed Egocentric Video Dataset},
  author        = {Perrett, Toby and Darkhalil, Ahmad and Sinha, Saptarshi and
                   Emara, Omar and Pollard, Sam and Parida, Kranti Kumar and
                   Liu, Kaiting and Gatti, Prajwal and Bansal, Siddhant and
                   Flanagan, Kevin and Chalk, Jacob and Zhu, Zhifan and
                   Guerrier, Rhodri and Abdelazim, Fahd and Zhu, Bin and
                   Moltisanti, Davide and Wray, Michael and Doughty, Hazel and
                   Damen, Dima},
  booktitle     = {IEEE/CVF Conference on Computer Vision and Pattern Recognition
                   (CVPR)},
  year          = {2025}
}

@inproceedings{wang2023holoassist,
  title         = {{HoloAssist}: An Egocentric Human Interaction Dataset for
                   Interactive {AI} Assistants in the Real World},
  author        = {Wang, Xin and Kwon, Taein and Rad, Mahdi and Pan, Bowen and
                   Chakraborty, Ishani and Andrist, Sean and Bohus, Dan and
                   Feniello, Ashley and Tekin, Bugra and Frujeri, Felipe Vieira and
                   Joshi, Neel and Pollefeys, Marc},
  booktitle     = {IEEE/CVF International Conference on Computer Vision (ICCV)},
  year          = {2023}
}

@inproceedings{peddi2024captaincook4d,
  title         = {{CaptainCook4D}: A Dataset for Understanding Errors in
                   Procedural Activities},
  author        = {Peddi, Rohith and Arya, Shivvrat and Challa, Bharath and
                   Pallapothula, Likhitha and Vyas, Akshay and Gouripeddi, Bhavya
                   and Zhang, Qifan and Wang, Jikai and Komaragiri, Vasundhara and
                   Ragan, Eric and Ruozzi, Nicholas and Xiang, Yu and Gogate, Vibhav},
  booktitle     = {Advances in Neural Information Processing Systems (NeurIPS)
                   Datasets and Benchmarks Track},
  year          = {2024}
}

@inproceedings{liu2022hoi4d,
  title         = {{HOI4D}: A 4D Egocentric Dataset for Category-Level
                   Human-Object Interaction},
  author        = {Liu, Yunze and Liu, Yun and Jiang, Che and Lyu, Kangbo and
                   Wan, Weikang and Shen, Hao and Liang, Boqiang and Fu, Zhoujie and
                   Wang, He and Yi, Li},
  booktitle     = {IEEE/CVF Conference on Computer Vision and Pattern Recognition
                   (CVPR)},
  year          = {2022}
}

@inproceedings{fan2023arctic,
  title         = {{ARCTIC}: A Dataset for Dexterous Bimanual Hand-Object
                   Manipulation},
  author        = {Fan, Zicong and Taheri, Omid and Tzionas, Dimitrios and
                   Kocabas, Muhammed and Kaufmann, Manuel and Black, Michael J. and
                   Hilliges, Otmar},
  booktitle     = {IEEE/CVF Conference on Computer Vision and Pattern Recognition
                   (CVPR)},
  year          = {2023}
}

@inproceedings{zhan2024oakink2,
  title         = {{OAKINK2}: A Dataset of Bimanual Hands-Object Manipulation in
                   Complex Task Completion},
  author        = {Zhan, Xinyu and Yang, Lixin and Zhao, Yifei and Mao, Kangrui and
                   Xu, Hanlin and Lin, Zenan and Li, Kailin and Lu, Cewu},
  booktitle     = {IEEE/CVF Conference on Computer Vision and Pattern Recognition
                   (CVPR)},
  year          = {2024}
}

@inproceedings{liu2024taco,
  title         = {{TACO}: Benchmarking Generalizable Bimanual Tool-Action-Object
                   Understanding},
  author        = {Liu, Yun and Yang, Haolin and Si, Xu and Liu, Ling and
                   Li, Zipeng and Zhang, Yuxiang and Liu, Yebin and Yi, Li},
  booktitle     = {IEEE/CVF Conference on Computer Vision and Pattern Recognition
                   (CVPR)},
  year          = {2024}
}

@inproceedings{banerjee2025hot3d,
  title         = {{HOT3D}: Hand and Object Tracking in 3D from Egocentric
                   Multi-View Videos},
  author        = {Banerjee, Prithviraj and Shkodrani, Sindi and Moulon, Pierre and
                   Hampali, Shreyas and Han, Shangchen and Zhang, Fan and
                   Zhang, Linguang and Fountain, Jade and Miller, Edward and
                   Basol, Selen and Newcombe, Richard and Wang, Robert and
                   Engel, Jakob Julian and Hodan, Tomas},
  booktitle     = {IEEE/CVF Conference on Computer Vision and Pattern Recognition
                   (CVPR)},
  year          = {2025}
}

@inproceedings{wang2025hocap,
  title         = {{HO-Cap}: A Capture System and Dataset for 3D Reconstruction and
                   Pose Tracking of Hand-Object Interaction},
  author        = {Wang, Jikai and Zhang, Qifan and Chao, Yu-Wei and Wen, Bowen and
                   Guo, Xiaohu and Xiang, Yu},
  booktitle     = {Advances in Neural Information Processing Systems (NeurIPS)
                   Datasets and Benchmarks Track},
  year          = {2025}
}

@article{hoque2025egodex,
  title         = {{EgoDex}: Learning Dexterous Manipulation from Large-Scale
                   Egocentric Video},
  author        = {Hoque, Ryan and Huang, Peide and Yoon, David J. and
                   Sivapurapu, Mouli and Zhang, Jian},
  journal       = {arXiv preprint arXiv:2505.11709},
  year          = {2025}
}

@inproceedings{khazatsky2024droid,
  title         = {{DROID}: A Large-Scale In-The-Wild Robot Manipulation Dataset},
  author        = {Khazatsky, Alexander and Pertsch, Karl and Nair, Suraj and
                   Balakrishna, Ashwin and Dasari, Sudeep and Karamcheti, Siddharth
                   and Nasiriany, Soroush and Srirama, Mohan Kumar and others},
  booktitle     = {Robotics: Science and Systems (RSS)},
  year          = {2024}
}

@inproceedings{walke2023bridgedata,
  title         = {{BridgeData V2}: A Dataset for Robot Learning at Scale},
  author        = {Walke, Homer and Black, Kevin and Zhao, Tony Z. and Vuong, Quan
                   and Zheng, Chongyi and Hansen-Estruch, Philippe and
                   He, Andre Wang and Myers, Vivek and Kim, Moo Jin and Du, Max and
                   Lee, Abraham and Fang, Kuan and Finn, Chelsea and Levine, Sergey},
  booktitle     = {Conference on Robot Learning (CoRL)},
  year          = {2023}
}

@inproceedings{wu2025robomind,
  title         = {{RoboMIND}: Benchmark on Multi-embodiment Intelligence Normative
                   Data for Robot Manipulation},
  author        = {Wu, Kun and Hou, Chengkai and Liu, Jiaming and Che, Zhengping and
                   Ju, Xiaozhu and Yang, Zhuqin and Li, Meng and Zhao, Yinuo and
                   Xu, Zhiyuan and Yang, Guang and others},
  booktitle     = {Robotics: Science and Systems (RSS)},
  year          = {2025}
}

@article{bu2025agibot,
  title         = {{AgiBot World Colosseo}: A Large-scale Manipulation Platform for
                   Scalable and Intelligent Embodied Systems},
  author        = {Bu, Qingwen and Cai, Jisong and Chen, Li and Cui, Xiuqi and
                   Ding, Yan and Feng, Siyuan and Gao, Shenyuan and He, Xindong and
                   Huang, Xu and Jiang, Shu and others},
  journal       = {arXiv preprint arXiv:2503.06669},
  year          = {2025}
}

@inproceedings{fang2024rh20t,
  title         = {{RH20T}: A Comprehensive Robotic Dataset for Learning Diverse
                   Skills in One-Shot},
  author        = {Fang, Hao-Shu and Fang, Hongjie and Tang, Zhenyu and Liu, Jirong
                   and Wang, Chenxi and Wang, Junbo and Zhu, Haoyi and Lu, Cewu},
  booktitle     = {IEEE International Conference on Robotics and Automation (ICRA)},
  year          = {2024}
}

@inproceedings{brohan2022rt1,
  title         = {{RT-1}: Robotics Transformer for Real-World Control at Scale},
  author        = {Brohan, Anthony and Brown, Noah and Carbajal, Justice and
                   Chebotar, Yevgen and Dabis, Joseph and Finn, Chelsea and
                   Gopalakrishnan, Keerthana and Hausman, Karol and Herzog, Alex and
                   Hsu, Jasmine and others},
  booktitle     = {Robotics: Science and Systems (RSS)},
  year          = {2023}
}

@inproceedings{jang2022bcz,
  title         = {{BC-Z}: Zero-Shot Task Generalization with Robotic Imitation
                   Learning},
  author        = {Jang, Eric and Irpan, Alex and Khansari, Mohi and Kappler, Daniel
                   and Ebert, Frederik and Lynch, Corey and Levine, Sergey and
                   Finn, Chelsea},
  booktitle     = {Conference on Robot Learning (CoRL)},
  year          = {2022}
}

@article{lee2025molmoact,
  title         = {{MolmoAct}: Action Reasoning Models that can Reason in Space},
  author        = {Lee, Jason and Duan, Jiafei and Fang, Haoquan and Deng, Yuquan and
                   Liu, Shuo and Li, Boyang and Fang, Bohan and Zhang, Jieyu and
                   Wang, Yi Ru and Lee, Sangho and Han, Winson and Pumacay, Wilbert
                   and Wu, Angelica and Hendrix, Rose and Farley, Karen and
                   VanderBilt, Eli and Farhadi, Ali and Fox, Dieter and Krishna, Ranjay},
  journal       = {arXiv preprint arXiv:2508.07917},
  year          = {2025}
}

@inproceedings{li2023behavior1k,
  title         = {{BEHAVIOR-1K}: A Benchmark for Embodied {AI} with 1,000 Everyday
                   Activities and Realistic Simulation},
  author        = {Li, Chengshu and Zhang, Ruohan and Wong, Josiah and Gokmen, Cem
                   and Srivastava, Sanjana and Mart{\'i}n-Mart{\'i}n, Roberto and
                   Wang, Chen and Levine, Gabrael and Lingelbach, Michael and
                   Sun, Jiankai and others},
  booktitle     = {Conference on Robot Learning (CoRL)},
  year          = {2022}
}

@article{qiu2025egome,
  title         = {{EgoMe}: A New Dataset and Challenge for Following Me via
                   Egocentric View in Real World},
  author        = {Qiu, Heqian and Shi, Zhaofeng and Wang, Lanxiao and
                   Xiong, Huiyu and Li, Xiang and Li, Hongliang},
  journal       = {arXiv preprint arXiv:2501.19061},
  year          = {2025}
}

@misc{egocentric10k,
  title         = {{Egocentric-10K}},
  author        = {{Build AI}},
  year          = {2025},
  howpublished  = {\url{https://huggingface.co/datasets/builddotai/Egocentric-10K}},
  note          = {Hugging Face dataset, approximately 10{,}000 hours of egocentric
                   factory video}
}

@misc{psibotai2025syndata,
  title         = {{SynData}: A Large-Scale Real-World Multimodal Dataset for
                   Embodied Intelligence},
  author        = {{PsiBotAI}},
  year          = {2025},
  howpublished  = {\url{https://huggingface.co/datasets/PsiBotAI/SynData}},
  note          = {Hugging Face dataset, real-world human manipulation captured with
                   an exoskeleton-glove system}
}

@misc{interndata_a1,
  title         = {{InternData-A1}},
  author        = {{InternRobotics}},
  year          = {2025},
  howpublished  = {\url{https://huggingface.co/datasets/InternRobotics/InternData-A1}},
  note          = {Hugging Face dataset, hybrid synthetic-real manipulation across
                   four embodiments}
}

@inproceedings{chi2024umi,
  title         = {Universal Manipulation Interface: In-The-Wild Robot Teaching
                   Without In-The-Wild Robots},
  author        = {Chi, Cheng and Xu, Zhenjia and Pan, Chuer and Cousineau, Eric and
                   Burchfiel, Benjamin and Feng, Siyuan and Tedrake, Russ and
                   Song, Shuran},
  booktitle     = {Robotics: Science and Systems (RSS)},
  year          = {2024}
}

@inproceedings{zhao2025tasterob,
  title         = {{TASTE-Rob}: Advancing Video Generation of Task-Oriented
                   Hand-Object Interaction for Generalizable Robotic Manipulation},
  author        = {Zhao, Hongxiang and Liu, Xingchen and Xu, Mutian and Hao, Yiming
                   and Chen, Weikai and Han, Xiaoguang},
  booktitle     = {IEEE/CVF Conference on Computer Vision and Pattern Recognition
                   (CVPR)},
  year          = {2025}
}

@article{galaxea2025,
  title         = {Galaxea Open-World Dataset and {G0} Dual-System {VLA} Model},
  author        = {{Galaxea Team}},
  journal       = {arXiv preprint arXiv:2509.00576},
  year          = {2025}
}

@article{zhao2025humanoideveryday,
  title         = {Humanoid Everyday: A Comprehensive Robotic Dataset for Open-World
                   Humanoid Manipulation},
  author        = {Zhao, Zhenyu and Jing, Hongyi and Liu, Xiawei and Mao, Jiageng and
                   Jha, Abha and Yang, Hanwen and Xue, Rong and Zakharov, Sergey and
                   Guizilini, Vitor and Wang, Yue},
  journal       = {arXiv preprint arXiv:2510.08807},
  year          = {2025}
}

@article{robocoin2025,
  title         = {{RoboCOIN}: An Open-Sourced Bimanual Robotic Data Collection for
                   Integrated Manipulation},
  author        = {{RoboCOIN Team}},
  journal       = {arXiv preprint arXiv:2511.17441},
  year          = {2025},
  note          = {Open-sourced by FlagOpen; full author list to be confirmed}
}

@misc{lejurobot2025let,
  title         = {{LET}: Full-Size Humanoid Robot Real-World Dataset},
  author        = {{LejuRobotics}},
  year          = {2025},
  howpublished  = {\url{https://huggingface.co/datasets/LejuRobotics/LET-Base-Dataset}},
  note          = {Hugging Face dataset, Kuavo humanoid teleoperation}
}

@misc{realominiopen,
  title         = {{RealOmni} Open Dataset},
  author        = {{GenRobot}},
  year          = {2025},
  howpublished  = {\url{https://www.genrobot.ai/data/open-dataset}},
  note          = {Open embodied-AI dataset, dual-hand household manipulation
                   collected with the GenDAS handheld gripper}
}

@inproceedings{zhou2025transfusion,
  title={Transfusion: Predict the next token and diffuse images with one multi-modal model},
  author={Zhou, Chunting and Yu, Lili and Babu, Arun and Tirumala, Kushal and Yasunaga, Michihiro and Shamis, Leonid and Kahn, Jacob and Ma, Xuezhe and Zettlemoyer, Luke and Levy, Omer},
  booktitle={International Conference on Learning Representations},
  volume={2025},
  pages={6446--6469},
  year={2025}
}

@article{team2026hy,
  title={HY-Embodied-0.5: Embodied Foundation Models for Real-World Agents},
  author={Team, HY and Yu, Xumin and Liu, Zuyan and Wang, Ziyi and Zhang, He and Rao, Yongming and Liu, Fangfu and Zhang, Yani and Zhao, Ruowen and Wang, Oran and others},
  journal={arXiv preprint arXiv:2604.07430},
  year={2026}
}

@article{dang2026rynnbrain,
  title={Rynnbrain: Open embodied foundation models},
  author={Dang, Ronghao and Guo, Jiayan and Hou, Bohan and Leng, Sicong and Li, Kehan and Li, Xin and Liu, Jiangpin and Mao, Yunxuan and Wang, Zhikai and Yuan, Yuqian and others},
  journal={arXiv preprint arXiv:2602.14979},
  year={2026}
}

@article{dang2025rynnec,
  title={Rynnec: Bringing mllms into embodied world},
  author={Dang, Ronghao and Yuan, Yuqian and Mao, Yunxuan and Li, Kehan and Liu, Jiangpin and Wang, Zhikai and Li, Xin and Wang, Fan and Zhao, Deli},
  journal={arXiv preprint arXiv:2508.14160},
  year={2025}
}

@article{agarwal2026cosmos,
  title={Cosmos 3: Omnimodal world models for physical ai},
  author={Agarwal, Niket and Ali, Arslan and Allen, Jon and Antolini, Martin and Aubame, Adeline and Azzolini, Alisson and Bai, Junjie and Bala, Maciej and Balaji, Yogesh and Bapst, Josh and others},
  journal={arXiv preprint arXiv:2606.02800},
  year={2026}
}

@article{azzolini2025cosmosreason1,
  title={Cosmos-reason1: From physical common sense to embodied reasoning},
  author={Azzolini, Alisson and Bai, Junjie and Brandon, Hannah and Cao, Jiaxin and Chattopadhyay, Prithvijit and Chen, Huayu and Chu, Jinju and Cui, Yin and Diamond, Jenna and Ding, Yifan and others},
  journal={arXiv preprint arXiv:2503.15558},
  year={2025}
}

@inproceedings{radford2021clip,
  title={Learning Transferable Visual Models From Natural Language Supervision},
  author={Radford, Alec and Kim, Jong Wook and Hallacy, Chris and Ramesh, Aditya and Goh, Gabriel and Agarwal, Sandhini and Sastry, Girish and Askell, Amanda and Mishkin, Pamela and Clark, Jack and Krueger, Gretchen and Sutskever, Ilya},
  booktitle={International Conference on Machine Learning (ICML)},
  pages={8748--8763},
  year={2021},
  organization={PMLR}
}

@inproceedings{alayrac2022flamingo,
  title={Flamingo: A Visual Language Model for Few-Shot Learning},
  author={Alayrac, Jean-Baptiste and Donahue, Jeff and Luc, Pauline and Miech, Antoine and Barr, Iain and Hasson, Yana and Lenc, Karel and Mensch, Arthur and Millican, Katherine and Reynolds, Malcolm and Ring, Roman and Rutherford, Eliza and Cabi, Serkan and Han, Tengda and Gong, Zhitao and Samangooei, Sina and Monteiro, Marianne and Menick, Jacob and Borgeaud, Sebastian and Brock, Andrew and Nematzadeh, Aida and Sharifzadeh, Sahand and Binkowski, Mikolaj and Barreira, Ricardo and Vinyals, Oriol and Zisserman, Andrew and Simonyan, Karen},
  booktitle={Advances in Neural Information Processing Systems (NeurIPS)},
  volume={35},
  pages={23716--23736},
  year={2022}
}

@inproceedings{li2023blip2,
  title={{BLIP-2}: Bootstrapping Language-Image Pre-training with Frozen Image Encoders and Large Language Models},
  author={Li, Junnan and Li, Dongxu and Savarese, Silvio and Hoi, Steven},
  booktitle={International Conference on Machine Learning (ICML)},
  year={2023}
}

@inproceedings{liu2023llava,
  title={Visual Instruction Tuning},
  author={Liu, Haotian and Li, Chunyuan and Wu, Qingyang and Lee, Yong Jae},
  booktitle={Advances in Neural Information Processing Systems (NeurIPS)},
  year={2023}
}

@article{bai2023qwenvl,
  title={{Qwen-VL}: A Versatile Vision-Language Model for Understanding, Localization, Text Reading, and Beyond},
  author={Bai, Jinze and Bai, Shuai and Yang, Shusheng and Wang, Shijie and Tan, Sinan and Wang, Peng and Lin, Junyang and Zhou, Chang and Zhou, Jingren},
  journal={arXiv preprint arXiv:2308.12966},
  year={2023}
}

@article{wang2024qwen2vl,
  title={Qwen2-vl: Enhancing vision-language model's perception of the world at any resolution},
  author={Wang, Peng and Bai, Shuai and Tan, Sinan and Wang, Shijie and Fan, Zhihao and Bai, Jinze and Chen, Keqin and Liu, Xuejing and Wang, Jialin and Ge, Wenbin and others},
  journal={arXiv preprint arXiv:2409.12191},
  year={2024}
}

@article{bai2025qwen3vl,
  title={Qwen3-vl technical report},
  author={Bai, Shuai and Cai, Yuxuan and Chen, Ruizhe and Chen, Keqin and Chen, Xionghui and Cheng, Zesen and Deng, Lianghao and Ding, Wei and Gao, Chang and Ge, Chunjiang and others},
  journal={arXiv preprint arXiv:2511.21631},
  year={2025}
}

@inproceedings{chen2024internvl,
  title={Internvl: Scaling up vision foundation models and aligning for generic visual-linguistic tasks},
  author={Chen, Zhe and Wu, Jiannan and Wang, Wenhai and Su, Weijie and Chen, Guo and Xing, Sen and Zhong, Muyan and Zhang, Qinglong and Zhu, Xizhou and Lu, Lewei and others},
  booktitle={Proceedings of the IEEE/CVF conference on computer vision and pattern recognition},
  pages={24185--24198},
  year={2024}
}

@article{zhu2025internvl3,
  title={Internvl3: Exploring advanced training and test-time recipes for open-source multimodal models},
  author={Zhu, Jinguo and Wang, Weiyun and Chen, Zhe and Liu, Zhaoyang and Ye, Shenglong and Gu, Lixin and Tian, Hao and Duan, Yuchen and Su, Weijie and Shao, Jie and others},
  journal={arXiv preprint arXiv:2504.10479},
  year={2025}
}

@article{openai2024gpt4o,
  title={{GPT-4o} System Card},
  author={OpenAI},
  journal={arXiv preprint arXiv:2410.21276},
  year={2024}
}

@article{ji2025robobrain,
  title={{RoboBrain}: A Unified Brain Model for Robotic Manipulation from Abstract to Concrete},
  author={Ji, Yuheng and Tan, Huajie and Shi, Jiayu and Hao, Xiaoshuai and Zhang, Yuan and Zhang, Hengyuan and Wang, Pengwei and Zhao, Mengdi and Mu, Yao and An, Pengju and Huang, Xinda and Wang, Jiaming and Zhang, Shengjie and Wang, Zhongyuan and others},
  journal={arXiv preprint arXiv:2502.21257},
  year={2025}
}

@article{robobrain2025v2,
  title={{RoboBrain 2.0} Technical Report},
  author={BAAI RoboBrain Team},
  journal={arXiv preprint},
  year={2025}
}

@article{wan2025,
      title={Wan: Open and Advanced Large-Scale Video Generative Models}, 
      author={Team Wan and Ang Wang and Baole Ai and Bin Wen and Chaojie Mao and Chen-Wei Xie and Di Chen and Feiwu Yu and Haiming Zhao and Jianxiao Yang and Jianyuan Zeng and Jiayu Wang and Jingfeng Zhang and Jingren Zhou and Jinkai Wang and Jixuan Chen and Kai Zhu and Kang Zhao and Keyu Yan and Lianghua Huang and Mengyang Feng and Ningyi Zhang and Pandeng Li and Pingyu Wu and Ruihang Chu and Ruili Feng and Shiwei Zhang and Siyang Sun and Tao Fang and Tianxing Wang and Tianyi Gui and Tingyu Weng and Tong Shen and Wei Lin and Wei Wang and Wei Wang and Wenmeng Zhou and Wente Wang and Wenting Shen and Wenyuan Yu and Xianzhong Shi and Xiaoming Huang and Xin Xu and Yan Kou and Yangyu Lv and Yifei Li and Yijing Liu and Yiming Wang and Yingya Zhang and Yitong Huang and Yong Li and You Wu and Yu Liu and Yulin Pan and Yun Zheng and Yuntao Hong and Yupeng Shi and Yutong Feng and Zeyinzi Jiang and Zhen Han and Zhi-Fan Wu and Ziyu Liu},
      journal = {arXiv preprint arXiv:2503.20314},
      year={2025}
}

@article{blattmann2023stable,
  title={Stable video diffusion: Scaling latent video diffusion models to large datasets},
  author={Blattmann, Andreas and Dockhorn, Tim and Kulal, Sumith and Mendelevitch, Daniel and Kilian, Maciej and Lorenz, Dominik and Levi, Yam and English, Zion and Voleti, Vikram and Letts, Adam and others},
  journal={arXiv preprint arXiv:2311.15127},
  year={2023}
}

@article{kondratyuk2023videopoet,
  title={Videopoet: A large language model for zero-shot video generation},
  author={Kondratyuk, Dan and Yu, Lijun and Gu, Xiuye and Lezama, Jos{\'e} and Huang, Jonathan and Schindler, Grant and Hornung, Rachel and Birodkar, Vighnesh and Yan, Jimmy and Chiu, Ming-Chang and others},
  journal={arXiv preprint arXiv:2312.14125},
  year={2023}
}

@article{bardes2024vjepa,
  title={Revisiting Feature Prediction for Learning Visual Representations from Video},
  author={Bardes, Adrien and Garrido, Quentin and Ponce, Jean and Rabbat, Michael and LeCun, Yann and Assran, Mahmoud and Ballas, Nicolas},
  journal={arXiv:2404.08471},
  year={2024}
}

@article{assran2025vjepa2,
  title={V-JEPA~2: Self-Supervised Video Models Enable Understanding, Prediction and Planning},
  author={Assran, Mahmoud and Bardes, Adrien and Fan, David and Garrido, Quentin and Howes, Russell and
Komeili, Mojtaba and Muckley, Matthew and Rizvi, Ammar and Roberts, Claire and Sinha, Koustuv and Zholus, Artem and
Arnaud, Sergio and Gejji, Abha and Martin, Ada and Robert Hogan, Francois and Dugas, Daniel and
Bojanowski, Piotr and Khalidov, Vasil and Labatut, Patrick and Massa, Francisco and Szafraniec, Marc and
Krishnakumar, Kapil and Li, Yong and Ma, Xiaodong and Chandar, Sarath and Meier, Franziska and LeCun, Yann and
Rabbat, Michael and Ballas, Nicolas},
  journal={arXiv preprint arXiv:2506.09985},
  year={2025}
}

@article{hu2024vpp,
  title={Video prediction policy: A generalist robot policy with predictive visual representations},
  author={Hu, Yucheng and Guo, Yanjiang and Wang, Pengchao and Chen, Xiaoyu and Wang, Yen-Jen and Zhang, Jianke and Sreenath, Koushil and Lu, Chaochao and Chen, Jianyu},
  journal={arXiv preprint arXiv:2412.14803},
  year={2024}
}

@article{guo2025ctrl,
  title={Ctrl-world: A controllable generative world model for robot manipulation},
  author={Guo, Yanjiang and Shi, Lucy Xiaoyang and Chen, Jianyu and Finn, Chelsea},
  journal={arXiv preprint arXiv:2510.10125},
  year={2025}
}

@article{lingbot-va2026,
  title={Causal World Modeling for Robot Control},
  author={Li, Lin and Zhang, Qihang and Luo, Yiming and Yang, Shuai and Wang, Ruilin and Han, Fei and Yu, Mingrui and Gao, Zelin and Xue, Nan and Zhu, Xing and Shen, Yujun and Xu, Yinghao},
  journal={arXiv preprint arXiv:2601.21998},
  year={2026}
}

@misc{ye2026dreamzero,
      title={World Action Models are Zero-shot Policies}, 
      author={Seonghyeon Ye and Yunhao Ge and Kaiyuan Zheng and Shenyuan Gao and Sihyun Yu and George Kurian and Suneel Indupuru and You Liang Tan and Chuning Zhu and Jiannan Xiang and Ayaan Malik and Kyungmin Lee and William Liang and Nadun Ranawaka and Jiasheng Gu and Yinzhen Xu and Guanzhi Wang and Fengyuan Hu and Avnish Narayan and Johan Bjorck and Jing Wang and Gwanghyun Kim and Dantong Niu and Ruijie Zheng and Yuqi Xie and Jimmy Wu and Qi Wang and Ryan Julian and Danfei Xu and Yilun Du and Yevgen Chebotar and Scott Reed and Jan Kautz and Yuke Zhu and Linxi "Jim" Fan and Joel Jang},
      year={2026},
      eprint={2602.15922},
      archivePrefix={arXiv},
      primaryClass={cs.RO},
      url={https://arxiv.org/abs/2602.15922}, 
}

@article{lin2026genie3,
  title   = {Fast and Ultra-Capable Protein Design: Advancing the Frontier
             Through Atomistic SE(3)-Equivariance with Genie 3},
  author  = {Lin, Yeqing and Lee, Minji and Vermani, Aakarsh and Jiang, Ellena
             and {De Cooman}, Sebastiaan and Spetko, Matej and AlQuraishi, Mohammed},
  journal = {bioRxiv},
  year    = {2026},
  doi     = {10.64898/2026.05.01.722168},
  url     = {https://www.biorxiv.org/content/10.64898/2026.05.01.722168v1}
}

@article{fang2026molmoact2,
  title={Molmoact2: Action reasoning models for real-world deployment},
  author={Fang, Haoquan and Duan, Jiafei and Clay, Donovan and Wang, Sam and Liu, Shuo and Huang, Weikai and Fan, Xiang and Tsai, Wei-Chuan and Chen, Shirui and Wang, Yi Ru and others},
  journal={arXiv preprint arXiv:2605.02881},
  year={2026}
}

@article{team2026lingbotworld,
  title={Advancing Open-source World Models},
  author={Team, Robbyant and Gao, Zelin and Wang, Qiuyu and Zeng, Yanhong and Zhu, Jiapeng and Cheng, Ka Leong and Li, Yixuan and Wang, Hanlin and Xu, Yinghao and Ma, Shuailei and others},
  journal={arXiv preprint arXiv:2601.20540},
  year={2026}
}

@misc{zhou2025paibench,
      title={PAI-Bench: A Comprehensive Benchmark For Physical AI}, 
      author={Fengzhe Zhou and Jiannan Huang and Jialuo Li and Deva Ramanan and Humphrey Shi},
      year={2025},
      eprint={2512.01989},
      archivePrefix={arXiv},
      primaryClass={cs.CV},
      url={https://arxiv.org/abs/2512.01989}, 
}

@article{yang2024vsi,
    title={{Thinking in Space: How Multimodal Large Language Models See, Remember and Recall Spaces}},
    author={Yang, Jihan and Yang, Shusheng and Gupta, Anjali and Han, Rilyn and Fei-Fei, Li and Xie, Saining},
    year={2024},
    journal={arXiv preprint arXiv:2412.14171},
}

@article{deng2026rbench,
  title={Rethinking Video Generation Model for the Embodied World},
  author={Deng, Yufan and Pan, Zilin and Zhang, Hongyu and Li, Xiaojie and Hu, Ruoqing and Ding, Yufei and Zou, Yiming and Zeng, Yan and Zhou, Daquan},
  journal={arXiv preprint arXiv:2601.15282},
  year={2026}}

@article{zhang2026hyembodied,
  title={Hy-Embodied-0.5-VLA: From Vision-Language-Action Models to a Real-World Robot Learning Stack},
  author={Zhang, He and Xiang, Lingzhu and Lin, Haitao and Huang, Zeyu and Wang, Minghui and Zhong, Dingyan and Dong, Yubo and Wu, Yihao and Rao, Yongming and Zhang, Dongsheng and He, Wanjia and Chen, Ling and Huang, Kai and Chen, Jiahao and Su, Sichang and Yu, Xumin and Wang, Ziyi and Zhu, Chengwei and Teng, Xiao and Guo, Yuchun and Zhang, Yufeng and Liu, Yuandong and Wang, Rui and Lu, Zisheng and Hu, Han and Zhang, Zhengyou},
  journal={arXiv preprint arXiv:2606.14409},
  year={2026}
}

@article{liang2024mixture,
  title={Mixture-of-transformers: A sparse and scalable architecture for multi-modal foundation models},
  author={Liang, Weixin and Yu, Lili and Luo, Liang and Iyer, Srinivasan and Dong, Ning and Zhou, Chunting and Ghosh, Gargi and Lewis, Mike and Yih, Wen-tau and Zettlemoyer, Luke and others},
  journal={arXiv preprint arXiv:2411.04996},
  year={2024}
}

@misc{motus,
            title={Motus: A Unified Latent Action World Model}, 
            author={Hongzhe Bi and Hengkai Tan and Shenghao Xie and Zeyuan Wang and Shuhe Huang and Haitian Liu and Ruowen Zhao and Yao Feng and Chendong Xiang and Yinze Rong and Hongyan Zhao and Hanyu Liu and Zhizhong Su and Lei Ma and Hang Su and Jun Zhu},
            year={2025},
            eprint={2512.13030},
            archivePrefix={arXiv},
            primaryClass={cs.CV},
            url={https://arxiv.org/abs/2512.13030}, 
      }

@misc{zhang2025upvlaunifiedunderstandingprediction,
      title={UP-VLA: A Unified Understanding and Prediction Model for Embodied Agent}, 
      author={Jianke Zhang and Yanjiang Guo and Yucheng Hu and Xiaoyu Chen and Xiang Zhu and Jianyu Chen},
      year={2025},
      eprint={2501.18867},
      archivePrefix={arXiv},
      primaryClass={cs.CV},
      url={https://arxiv.org/abs/2501.18867}, 
}

@article{liang2025mm,
  title={MM-ACT: Learn from Multimodal Parallel Generation to Act},
  author={Liang, Haotian and Chen, Xinyi and Wang, Bin and Chen, Mingkang and Liu, Yitian and Zhang, Yuhao and Chen, Zanxin and Yang, Tianshuo and Chen, Yilun and Pang, Jiangmiao and others},
  journal={arXiv preprint arXiv:2512.00975},
  year={2025}
}

@article{cen2025rynnvla,
  title={RynnVLA-002: A Unified Vision-Language-Action and World Model},
  author={Cen, Jun and Huang, Siteng and Yuan, Yuqian and Li, Kehan and Yuan, Hangjie and Yu, Chaohui and Jiang, Yuming and Guo, Jiayan and Li, Xin and Luo, Hao and Wang, Fan and Wang, Fan and Zhao, Deli},
  journal={arXiv preprint arXiv:2511.17502},
  year={2025}
}

@article{internvla_a1,
  title={InternVLA-A1: Unifying Understanding, Generation and Action for Robotic Manipulation},
  author={Cai, Junhao and Cai, Zetao and Cao, Jiafei and Chen, Yilun and He, Zeyu and Jiang, Lei and Li, Hang and Li, Hengjie and Li, Yang and Liu, Yufei and others},
  journal={arXiv preprint arXiv:2601.02456},
  year={2026}
}

@misc{hu2026bagelvlaenhancinglonghorizonmanipulation,
      title={BagelVLA: Enhancing Long-Horizon Manipulation via Interleaved Vision-Language-Action Generation}, 
      author={Yucheng Hu and Jianke Zhang and Yuanfei Luo and Yanjiang Guo and Xiaoyu Chen and Xinshu Sun and Kun Feng and Qingzhou Lu and Sheng Chen and Yangang Zhang and Wei Li and Jianyu Chen},
      year={2026},
      eprint={2602.09849},
      archivePrefix={arXiv},
      primaryClass={cs.RO},
      url={https://arxiv.org/abs/2602.09849}, 
}

@article{depth_anything_v2,
  title={Depth Anything V2},
  author={Yang, Lihe and Kang, Bingyi and Huang, Zilong and Zhao, Zhen and Xu, Xiaogang and Feng, Jiashi and Zhao, Hengshuang},
  journal={arXiv:2406.09414},
  year={2024}
}

@misc{ghosh2023genevalobjectfocusedframeworkevaluating,
      title={GenEval: An Object-Focused Framework for Evaluating Text-to-Image Alignment}, 
      author={Dhruba Ghosh and Hanna Hajishirzi and Ludwig Schmidt},
      year={2023},
      eprint={2310.11513},
      archivePrefix={arXiv},
      primaryClass={cs.CV},
      url={https://arxiv.org/abs/2310.11513}, 
}

@misc{du2024embspatialbenchbenchmarkingspatialunderstanding,
      title={EmbSpatial-Bench: Benchmarking Spatial Understanding for Embodied Tasks with Large Vision-Language Models}, 
      author={Mengfei Du and Binhao Wu and Zejun Li and Xuanjing Huang and Zhongyu Wei},
      year={2024},
      eprint={2406.05756},
      archivePrefix={arXiv},
      primaryClass={cs.AI},
      url={https://arxiv.org/abs/2406.05756}, 
}

@misc{tong2024cambrian1,
      title={Cambrian-1: A Fully Open, Vision-Centric Exploration of Multimodal LLMs},
      author={Shengbang Tong and Ellis Brown and Penghao Wu and Sanghyun Woo and Manoj Middepogu and Sai Charitha Akula and Jihan Yang and Shusheng Yang and Adithya Iyer and Xichen Pan and Austin Wang and Rob Fergus and Yann LeCun and Saining Xie},
      year={2024},
      eprint={2406.16860},
}

@article{ray2024sat,
  title={Sat: Spatial aptitude training for multimodal language models},
  author={Ray, Arijit and Duan, Jiafei and Tan, Reuben and Bashkirova, Dina and Hendrix, Rose and Ehsani, Kiana and Kembhavi, Aniruddha and Plummer, Bryan A and Krishna, Ranjay and Zeng, Kuo-Hao and others},
  journal={arXiv preprint arXiv:2412.07755},
  volume={3},
  year={2024}
}

@article{ma20243dsrbench,
  title={3DSRBench: A Comprehensive 3D Spatial Reasoning Benchmark},
  author={Ma, Wufei and Chen, Haoyu and Zhang, Guofeng and de Melo, Celso M and Yuille, Alan and Chen, Jieneng},
  journal={arXiv preprint arXiv:2412.07825},
  year={2024}
}

@article{team2025gemini,
  title={Gemini robotics: Bringing ai into the physical world},
  author={Team, Gemini Robotics and Abeyruwan, Saminda and Ainslie, Joshua and Alayrac, Jean-Baptiste and Arenas, Montserrat Gonzalez and Armstrong, Travis and Balakrishna, Ashwin and Baruch, Robert and Bauza, Maria and Blokzijl, Michiel and others},
  journal={arXiv preprint arXiv:2503.20020},
  year={2025}
}

@misc{tan2026robobrain25depthsight,
      title={RoboBrain 2.5: Depth in Sight, Time in Mind}, 
      author={Huajie Tan and Enshen Zhou and Zhiyu Li and Yijie Xu and Yuheng Ji and Xiansheng Chen and Cheng Chi and Pengwei Wang and Huizhu Jia and Yulong Ao and Mingyu Cao and Sixiang Chen and Zhe Li and Mengzhen Liu and Zixiao Wang and Shanyu Rong and Yaoxu Lyu and Zhongxia Zhao and Peterson Co and Yibo Li and Yi Han and Shaoxuan Xie and Guocai Yao and Songjing Wang and Leiduo Zhang and Xi Yang and Yance Jiao and Donghai Shi and Kunchang Xie and Shaokai Nie and Chunlei Men and Yonghua Lin and Zhongyuan Wang and Tiejun Huang and Shanghang Zhang},
      year={2026},
      eprint={2601.14352},
      archivePrefix={arXiv},
      primaryClass={cs.RO},
      url={https://arxiv.org/abs/2601.14352}, 
}

@article{zhou2025roborefer,
  title={RoboRefer: Towards Spatial Referring with Reasoning in Vision-Language Models for Robotics},
  author={Zhou, Enshen and An, Jingkun and Chi, Cheng and Han, Yi and Rong, Shanyu and Zhang, Chi and Wang, Pengwei and Wang, Zhongyuan and Huang, Tiejun and Sheng, Lu and others},
  journal={arXiv preprint arXiv:2506.04308},
  year={2025}
}

@article{yuan2024robopoint,
  title={RoboPoint: A Vision-Language Model for Spatial Affordance Prediction for Robotics},
  author={Yuan, Wentao and Duan, Jiafei and Blukis, Valts and Pumacay, Wilbert and Krishna, Ranjay and Murali, Adithyavairavan and Mousavian, Arsalan and Fox, Dieter},
  journal={arXiv preprint arXiv:2406.10721},
  year={2024}
}

@inproceedings{song2025robospatial,
  author    = {Song, Chan Hee and Blukis, Valts and Tremblay, Jonathan and Tyree, Stephen and Su, Yu and Birchfield, Stan},
  title     = {{RoboSpatial}: Teaching Spatial Understanding to {2D} and {3D} Vision-Language Models for Robotics},
  booktitle = {Proceedings of the IEEE/CVF Conference on Computer Vision and Pattern Recognition (CVPR)},
  year      = {2025},
  note      = {Oral Presentation},
}

@inproceedings{yin2025mindcube,
  title={Spatial mental modeling from limited views},
  author={Yin, Baiqiao and Wang, Qineng and Zhang, Pingyue and Zhang, Jianshu and Wang, Kangrui and Wang, Zihan and Zhang, Jieyu and Chandrasegaran, Keshigeyan and Liu, Han and Krishna, Ranjay and others},
  booktitle={Structural Priors for Vision Workshop at ICCV'25},
  year={2025}
}

@article{yang2025mmsi,
  title={MMSI-Bench: A Benchmark for Multi-Image Spatial Intelligence},
  author={Yang, Sihan and Xu, Runsen and Xie, Yiman and Yang, Sizhe and Li, Mo and Lin, Jingli and Zhu, Chenming and Chen, Xiaochen and Duan, Haodong and Yue, Xiangyu and Lin, Dahua and Wang, Tai and Pang, Jiangmiao},
  journal={arXiv preprint arXiv:2505.23764},
  year={2025}
}

@article{li2025viewspatial,
  title={Viewspatial-bench: Evaluating multi-perspective spatial localization in vision-language models},
  author={Li, Dingming and Li, Hongxing and Wang, Zixuan and Yan, Yuchen and Zhang, Hang and Chen, Siqi and Hou, Guiyang and Jiang, Shengpei and Zhang, Wenqi and Shen, Yongliang and others},
  journal={arXiv preprint arXiv:2505.21500},
  year={2025}
}

@inproceedings{wang2025site,
  title={Site: towards spatial intelligence thorough evaluation},
  author={Wang, Wenqi and Tan, Reuben and Zhu, Pengyue and Yang, Jianwei and Yang, Zhengyuan and Wang, Lijuan and Kolobov, Andrey and Gao, Jianfeng and Gong, Boqing},
  booktitle={Proceedings of the IEEE/CVF International Conference on Computer Vision},
  pages={9058--9069},
  year={2025}
}

@misc{intelligence2026pi07steerablegeneralistrobotic,
      title={${\pi}_{0.7}$: a Steerable Generalist Robotic Foundation Model with Emergent Capabilities}, 
      author={Physical Intelligence and Bo Ai and Ali Amin and Raichelle Aniceto and Ashwin Balakrishna and Greg Balke and Kevin Black and George Bokinsky and Shihao Cao and Thomas Charbonnier and Vedant Choudhary and Foster Collins and Ken Conley and Grace Connors and James Darpinian and Karan Dhabalia and Maitrayee Dhaka and Jared DiCarlo and Danny Driess and Michael Equi and Adnan Esmail and Yunhao Fang and Chelsea Finn and Catherine Glossop and Thomas Godden and Ivan Goryachev and Lachlan Groom and Haroun Habeeb and Hunter Hancock and Karol Hausman and Gashon Hussein and Victor Hwang and Brian Ichter and Connor Jacobsen and Szymon Jakubczak and Rowan Jen and Tim Jones and Gregg Kammerer and Ben Katz and Liyiming Ke and Mairbek Khadikov and Chandra Kuchi and Marinda Lamb and Devin LeBlanc and Brendon LeCount and Sergey Levine and Xinyu Li and Adrian Li-Bell and Vladislav Lialin and Zhonglin Liang and Wallace Lim and Yao Lu and Enyu Luo and Vishnu Mano and Nandan Marwaha and Aikys Mongush and Liam Murphy and Suraj Nair and Tyler Patterson and Karl Pertsch and Allen Z. Ren and Gavin Schelske and Charvi Sharma and Baifeng Shi and Lucy Xiaoyang Shi and Laura Smith and Jost Tobias Springenberg and Kyle Stachowicz and Will Stoeckle and Jiaming Tang and Jimmy Tanner and Shalom Tekeste and Marcel Torne and Kyle Vedder and Quan Vuong and Anna Walling and Haohuan Wang and Jason Wang and XuDong Wang and Chris Whalen and Samuel Whitmore and Blake Williams and Charles Xu and Sukwon Yoo and Lili Yu and Wuming Zhang and Zhuoyang Zhang and Ury Zhilinsky},
      year={2026},
      eprint={2604.15483},
      archivePrefix={arXiv},
      primaryClass={cs.LG},
      url={https://arxiv.org/abs/2604.15483}, 
}

@article{yuan2026qwen,
  title={Qwen-robotmanip technical report: Alignment unlocks scale for robotic manipulation foundation models},
  author={Yuan, Haoqi and Liang, Zhixuan and Chen, Anzhe and Wang, Ye and Li, Haoyang and Lin, Pei and Huang, Yiyang and Lei, Zixing and Zhang, Tong and Zhang, Jiazhao and others},
  journal={arXiv preprint arXiv:2606.17846},
  year={2026}
}

@misc{wu2026pragmaticvlafoundationmodel,
      title={A Pragmatic VLA Foundation Model}, 
      author={Wei Wu and Fan Lu and Yunnan Wang and Shuai Yang and Shi Liu and Fangjing Wang and Qian Zhu and He Sun and Yong Wang and Shuailei Ma and Yiyu Ren and Kejia Zhang and Hui Yu and Jingmei Zhao and Shuai Zhou and Zhenqi Qiu and Houlong Xiong and Ziyu Wang and Zechen Wang and Ran Cheng and Yong-Lu Li and Yongtao Huang and Xing Zhu and Yujun Shen and Kecheng Zheng},
      year={2026},
      eprint={2601.18692},
      archivePrefix={arXiv},
      primaryClass={cs.RO},
      url={https://arxiv.org/abs/2601.18692}, 
}

@article{chen2025internvla,
  title={Internvla-m1: A spatially guided vision-language-action framework for generalist robot policy},
  author={Chen, Xinyi and Chen, Yilun and Fu, Yanwei and Gao, Ning and Jia, Jiaya and Jin, Weiyang and Li, Hao and Mu, Yao and Pang, Jiangmiao and Qiao, Yu and others},
  journal={arXiv preprint arXiv:2510.13778},
  year={2025}
}

@article{ye2026world,
  title={World action models are zero-shot policies},
  author={Ye, Seonghyeon and Ge, Yunhao and Zheng, Kaiyuan and Gao, Shenyuan and Yu, Sihyun and Kurian, George and Indupuru, Suneel and Tan, You Liang and Zhu, Chuning and Xiang, Jiannan and others},
  journal={arXiv preprint arXiv:2602.15922},
  year={2026}
}

@article{gao2026dreamdojo,
  title={DreamDojo: A Generalist Robot World Model from Large-Scale Human Videos},
  author={Gao, Shenyuan and Liang, William and Zheng, Kaiyuan and Malik, Ayaan and Ye, Seonghyeon and Yu, Sihyun and Tseng, Wei-Cheng and Dong, Yuzhu and Mo, Kaichun and Lin, Chen-Hsuan and others},
  journal={arXiv preprint arXiv:2602.06949},
  year={2026}
}

@article{kim2026cosmos,
  title={Cosmos policy: Fine-tuning video models for visuomotor control and planning},
  author={Kim, Moo Jin and Gao, Yihuai and Lin, Tsung-Yi and Lin, Yen-Chen and Ge, Yunhao and Lam, Grace and Liang, Percy and Song, Shuran and Liu, Ming-Yu and Finn, Chelsea and others},
  journal={arXiv preprint arXiv:2601.16163},
  year={2026}
}

@misc{ye2026gigaworldpolicyefficientactioncenteredworldaction,
      title={GigaWorld-Policy: An Efficient Action-Centered World--Action Model}, 
      author={Angen Ye and Boyuan Wang and Chaojun Ni and Guan Huang and Guosheng Zhao and Hao Li and Hengtao Li and Jie Li and Jindi Lv and Jingyu Liu and Min Cao and Peng Li and Qiuping Deng and Wenjun Mei and Xiaofeng Wang and Xinze Chen and Xinyu Zhou and Yang Wang and Yifan Chang and Yifan Li and Yukun Zhou and Yun Ye and Zhichao Liu and Zheng Zhu},
      year={2026},
      eprint={2603.17240},
      archivePrefix={arXiv},
      primaryClass={cs.CV},
      url={https://arxiv.org/abs/2603.17240}, 
}

@article{singh2025openai,
  title={Openai gpt-5 system card},
  author={Singh, Aaditya and Fry, Adam and Perelman, Adam and Tart, Adam and Ganesh, Adi and El-Kishky, Ahmed and McLaughlin, Aidan and Low, Aiden and Ostrow, AJ and Ananthram, Akhila and others},
  journal={arXiv preprint arXiv:2601.03267},
  year={2025}
}
